\documentclass{article}

    \PassOptionsToPackage{numbers, compress}{natbib}

\usepackage{natbib}
\usepackage{multibib}

\newcites{app}{Additional References}
\bibliographystyleapp{plainnat}

\usepackage[final]{neurips_2025}

\usepackage[utf8]{inputenc} 
\usepackage[T1]{fontenc}    
\usepackage{url}            
\usepackage{booktabs}       
\usepackage{amsfonts}       
\usepackage{nicefrac}       
\usepackage{microtype}      
\usepackage[dvipsnames]{xcolor}         
\usepackage{enumitem}
\usepackage{wrapfig}
\usepackage{csquotes}

\usepackage{tabularx}
\usepackage[most]{tcolorbox}
\usepackage{array}  

\usepackage{times,latexsym}
\usepackage{xspace,mfirstuc,tabulary}

\usepackage{times}
\usepackage{latexsym}
\usepackage{inconsolata}
\usepackage{graphicx}
\usepackage{subcaption}
\usepackage{wrapfig}
\usepackage{multirow}
\usepackage{longtable}
\usepackage{pifont}
\usepackage{lipsum}
\usepackage{mdframed}
\usepackage{soul}  %
\usepackage{tikz}
\usepackage[edges]{forest}
\definecolor{hidden-draw}{RGB}{0,0,0}
\usepackage{subcaption}
\usepackage[export]{adjustbox}
\usepackage{ragged2e}

\usepackage{xspace}  

\usepackage{booktabs} 
\usepackage{subcaption} 
\usepackage{graphicx}
\usepackage{booktabs}
\usepackage{makecell}
\usepackage{amsmath}

\usepackage{hyperref}

\title{Stable Cinemetrics : Structured Taxonomy and Evaluation for Professional Video Generation}

\newsavebox\authorbox

\begin{lrbox}{\authorbox}
\centering
\begin{tabular}{c}
    \begin{tabular}{ccc}
        Agneet Chatterjee$^{1,2,*}$ & Rahim Entezari$^{1}$ & Maksym Zhuravinskyi$^1$ \\
        Max Lapin$^1$ & Reshinth Adithyan$^1$ & Amit Raj$^3$ \\
        Chitta Baral$^2$ & Yezhou Yang$^2$ & Varun Jampani$^1$ \\
    \end{tabular}
    \\[2em] 
    $^1$Stability AI \quad $^2$Arizona State University \quad $^3$Google DeepMind \\[1em]
\end{tabular}
\end{lrbox}
\author{\usebox{\authorbox}}
\begin{document}

\renewcommand\thefootnote{} %
\footnotetext{$^*$Work done during an internship at Stability AI.}
\renewcommand\thefootnote{\arabic{footnote}} %

\maketitle

\vspace{-3.50em}
\begin{center}
\textcolor{blue}{\url{https://stable-cinemetrics.github.io/}}
\end{center}

\begin{abstract}
Recent advances in video generation have enabled high-fidelity video synthesis from user provided prompts. However, existing models and benchmarks fail to capture the complexity and requirements of professional video generation. Towards that goal, we introduce \textbf{S}table \textbf{Cine}metrics, a structured evaluation framework that formalizes filmmaking controls into four disentangled, hierarchical taxonomies: \emph{Setup, Event, Lighting, and Camera}. Together, these taxonomies define 76 fine-grained control nodes grounded in industry practices. Using these taxonomies, we construct a benchmark of prompts aligned with professional use cases and develop an automated pipeline for prompt categorization and question generation, enabling independent evaluation of each control dimension. We conduct a large-scale human study spanning 10+ models and 20K videos, annotated by a pool of 80+ film professionals. Our analysis, both coarse and fine-grained reveal that even the strongest current models exhibit significant gaps, particularly in Events and Camera-related controls. To enable scalable evaluation, we train an automatic evaluator, a vision-language model aligned with expert annotations that outperforms existing zero-shot baselines. SCINE is the first approach to situate professional video generation within the landscape of video generative models, introducing taxonomies centered around cinematic controls and supporting them with structured evaluation pipelines and detailed analyses to guide future research.
\end{abstract}
\section{Introduction}

The field of video generative models has made significant progress in recent years \cite{maslej2025artificialintelligenceindexreport}, drawing substantial interest from both academia and industry. This can be evidenced by the growing number of benchmarks \cite{Huang_2024_CVPR, NEURIPS2024_c6483c8a, bansal2025videophy, Liu_2024_CVPR}, datasets \cite{nan2025openvidm, wang2024videofactory}, and both open- \cite{blattmann2023stablevideodiffusionscaling, wan2025wanopenadvancedlargescale, kong2025hunyuanvideosystematicframeworklarge, yang2025cogvideox} and closed- \cite{videoworldsimulators2024, veo2, ray2_lumalabs_2025} source models that have collectively driven the field forward. 
The foundational nature of these models makes them useful for several downstream tasks, including video editing \cite{jiang2025vaceallinonevideocreation}, 3D generation \cite{voleti2024sv3dnovelmultiviewsynthesis} and robotics \cite{pmlr-v235-zhou24f}. This widespread adoption of video generative models underpins the growing assertion that they represent a revolution for \textit{professional video generation}.

Generative vision offers tremendous potential for media creation, but a fundamental question remains: how can we shift generative video from casual, exploratory synthesis to a medium that supports professional-grade, controllable cinematic outputs? The important distinction between casual and professional generative video lies in the critical gap of cinematic control \cite{brown2016cinematography}: while today's models can generate videos of "\textit{an astronaut riding a horse}", professional creation necessitates granular control over cinematic elements such as the framing of the shot, position of the key light, and even whether the astronaut smiles before \textit{or} after the horse gallops away - a truly professional video generation system must put every one of those cinematic choices back in the creator’s hands. The need for this exact control over every cinematic element, from the timing of a smile to the quality of light, is the very reason filmmakers shoot multiple takes, selecting only the frames where everything comes together to tell the story most effectively \cite{dmytryk2018film}.

\begin{figure}[]
  \centering
  \includegraphics[width=\linewidth]{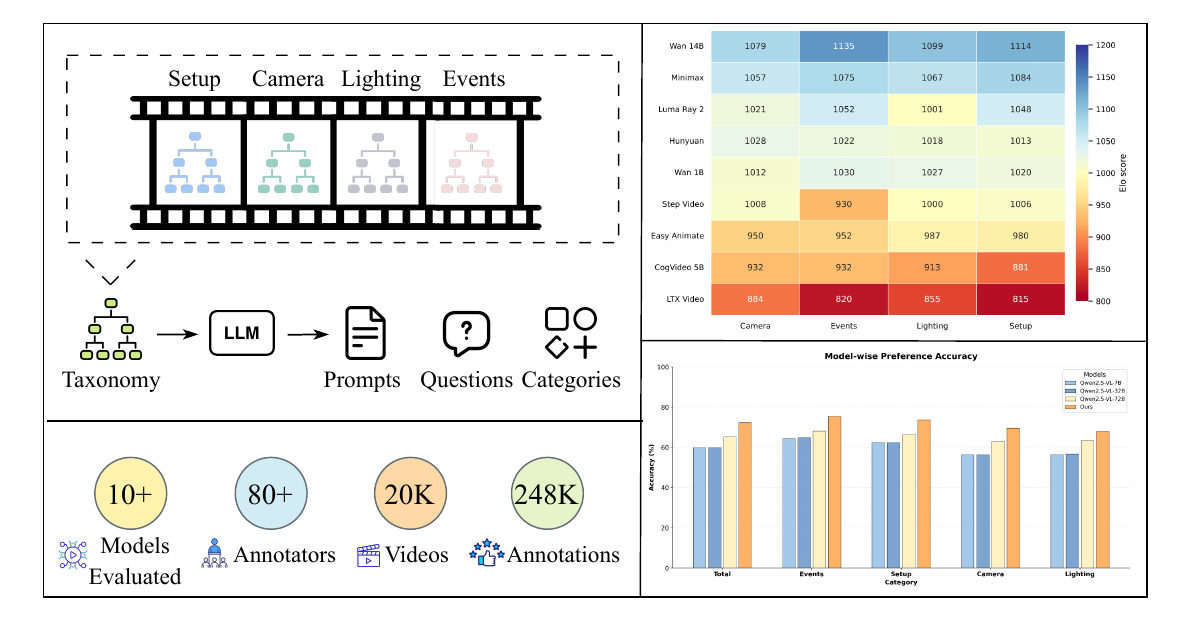}
  \caption {\textbf{Stable Cinemetrics} introduces structured taxonomies grounded in the controls required for professional video generation. These taxonomies form the foundation of our prompt based benchmark that mirrors real-world shot creation, progressing from scriptwriting to on-screen visuals. Every control element in a prompt is automatically categorized back to the taxonomy, enabling the generation of isolated evaluation questions for independent investigation into each element. This supports large scale human evaluation enabling both coarse and fine-grained insights into the capabilities of current models for professional video generation. To drive scalable annotations, we develop our own VLMs that outperform existing models in alignment with human judgements.}
  \label{fig:teaser}
  \vspace{-2.25em}
\end{figure}

In this work, we investigate the intersection of video generative models and the nuanced control mechanisms associated with professional video production. Despite rapid advancements in video generative modeling, the field still lacks both a clear definition of essential cinematic controls and standardized evaluation protocols for benchmarking progress at this crucial intersection. To bridge this gap, we present SCINE (\textbf{S}table \textbf{Cine}metrics) -  an evaluation suite specifically designed to characterize this intersection, enabling us to directly address the question: "\textit{Are Current Video Generative Models Ready for Professional Use?}"

The core focus of SCINE is to develop a taxonomy, at the intersection of generative vision and professional video generation, built on the principles of control and specificity \cite{block2020visual, katz1991film, rabiger2013directing}. These principles are important in film-making, as every decision carries cinematic meaning; a low angle conveys power while low-key lighting evokes drama. SCINE captures these principles by organizing control knobs into four cinematic pillars - Setup, Events, Lighting, and Camera - enabling the evaluation of video generative models along the same axes that a professional would. This disentanglement also allows evaluation around \textit{personalization}: which aligns with the collaborative nature of real productions, in comparison to the monolithic nature of current video generation models. Our taxonomy is hierarchical, branching from coarse cinematic concepts to leaf-level controls that naturally map onto computer vision concepts such as object semantics and scene geometry \cite{szeliski2022computer}. 

Leveraging our taxonomy, we generate two prompt types: story-driven and visual exposition, to mirror professional workflows. Story-driven prompts act as mini-screenplays \cite{field2005screenplay}, specifying characters, dialogue, actions, and emotions. We enrich these with visual exposition cues, by sampling control nodes from our taxonomy, emulating the transition from script to shot in filmmaking \cite{seger2004script}.  Sampling control nodes allows automated (a) prompt categorization: mapping each control element to the taxonomy, and (b) generation of targeted evaluation questions for each element, allowing disentagled evaluation of each cinematic control. The structured nature of our taxonomy supports scalable human annotation: we evaluate 10+ models across 20K generated videos with feedback from 80+ professionals. This enables analysis across control dimensions where we observe substantial variance in performance across taxonomy pillars, even for top-performing models such as WAN-14B and Minimax. Our taxonomy facilitates both coarse insights, showing that models struggle most with Events and Camera and fine-grained comparisons, such as better performance on shot size over camera framing, and on natural over artificial lighting. To support automatic evaluation of fine-grained cinematic controls in generated videos, we train a vision-language model (VLM) that aligns with the large-scale human annotations. Our model outperforms existing baselines, achieving an overall accuracy of 72.36\% with human annotators. An overview of SCINE, outlining its contributions is shown in Figure \ref{fig:teaser}.

\section{Related Work} 

\textbf{Video Generative Models and Evaluations.} Video generative models can broadly be classified into two categories based on their input conditioning: image-to-video (I2V) and text-to-video (T2V). While I2V approaches such as Stable Video Diffusion (SVD) \cite{blattmann2023stablevideodiffusionscaling} have been widely adopted by the community, the focus of our work is to evaluate T2V models for professional use. We choose text as an input modality, since it is an effective and free-form way of describing the controls defined in our taxonomy. The Sora Preview \cite{videoworldsimulators2024} served as a catalyst for a wave of T2V model releases across the closed \cite{veo2, polyak2025moviegencastmedia} and open-source communities. State of the art open-source models include Wan \cite{wan2025wanopenadvancedlargescale}, HunyuanVideo \cite{kong2025hunyuanvideosystematicframeworklarge} and StepVideo \cite{ma2025stepvideot2vtechnicalreportpractice}. 

Traditional reference-based metrics such as FVD \cite{unterthiner2019accurategenerativemodelsvideo} are widely used in video generation, however they have been found to not align with human judgment \cite{luo2025beyond} and are susceptible to content bias \cite{ge2024contentbiasfrechetvideo}. Several T2V evaluation benchmarks have also emerged, with VBench \cite{Huang_2024_CVPR, zheng2025vbench20advancingvideogeneration} gaining broad adoption. VBench evaluates T2V models by developing text prompts that evaluate generated videos across dimensions such as temporal flickering, aesthetic quality and motion smoothness, while employing automatic metrics like CLIP-based similarity \cite{radford2021learningtransferablevisualmodels}, LAION aesthetic predictor \cite{laion2022aesthetic} and, frame interpolation consistency \cite{li2023amtallpairsmultifieldtransforms}. While comprehensive, VBench is sub-optimal for professional video generation, lacking fine grained controls desired by professionals. Its broad evaluation scope e.g., sampling from 8 general categories like Animal and Vehicles, does not reflect domain specific requirements in a professional context. Additional T2V benchmarks include : VideoPhy \cite{bansal2025videophy} - which evaluates physical plausibility by measuring adherence of generated videos to real-world physics; DEVIL \cite{NEURIPS2024_c6483c8a}, which quantifies structural and perceptual dynamics of videos at a frame and segment level, and; T2V-CompBench \cite{sun2025t2vcompbenchcomprehensivebenchmarkcompositional}, which studies compositional consistency \cite{huang2025t2icompbenchenhancedcomprehensivebenchmark} in video generation. SCINE differs from existing benchmarks such as VBench or MovieGenBench \cite{polyak2025moviegencastmedia} by being the first to specifically evaluate video generative models for use in professional video creation. This distinction is brought about by our taxonomy, which also introduces flexibility over the entire evaluation pipeline: managing complexity of prompts, incorporating personalization, and automating both prompt categorization and question generation. Existing benchmarks lack cinematic depth; for example, a prompt such as "\textit{A man is walking}" from VBench-2 \cite{zheng2025vbench20advancingvideogeneration} omits key details such as character appearance, setting, or camera movement; all essential to setup a cinematic shot. Existing benchmarks are static, relying on fixed prompt sets that limit extensibility. In contrast, SCINE's taxonomy guided prompt generation enables future-proof evaluation, allowing prompt complexity to scale with model capabilities.

\textbf{Structured Video Generation and Shot-Level Control.} MovieNet \cite{huang2020movienet} is a large-scale, annotated corpus of 1,100 movies with extensive metadata including character bounding boxes/IDs, scene boundaries, action/place tags, cinematic‑style tags, and synopsis segments. Storyboard-inspired video generation has also been explored by the community: VDS \cite{rao2023dynamic} develops a pre‑visualisation system that converts story and camera textual inputs into full‑length 3‑D animatic storyboards; the candidate storyboards are further classified with a shot ranking discriminator trained to distinguish professional film clips from randomly sampled ones. VAST \cite{zhang2024vast10unifiedframework} reformulates T2V generation as a two-stage process: first generating detailed storyboards with pose and layout information from text, followed by video synthesis conditioned on both the storyboard and the text prompt. Preliminary work on multi-shot video generation includes VideoGen-of-Thought \cite{zheng2025videogenofthoughtstepbystepgeneratingmultishot}, which decomposes the task into four stages such as keyframe rendering and latent-space smoothing, to maintain identity and stylistic consistency across shots. MovieAgent \cite{wu2025automatedmoviegenerationmultiagent} approaches the problem via hierarchical planning, where a system of LLM agents employs Chain-of-Thought reasoning \cite{wei2023chainofthoughtpromptingelicitsreasoning} to parse scripts into scenes, shots, camera motions, dialogue, and audio, subsequently invoking diffusion-based models for multi-shot synthesis. However, the aforementioned works do not examine the fundamental structure of a shot, the nuanced controls and the interplay of cinematic elements that define it. Prior efforts often highlight isolated aspects, such as camera motion or the quality of light, while overlooking the interplay amongst each control dimension. In contrast, SCINE formalizes a structured taxonomy encompassing the full spectrum of shot-level attributes.
\begin{figure}[t]
  \centering
  \begin{minipage}[c]{0.48\linewidth}
    \centering
    \includegraphics[valign=t, width=\linewidth]{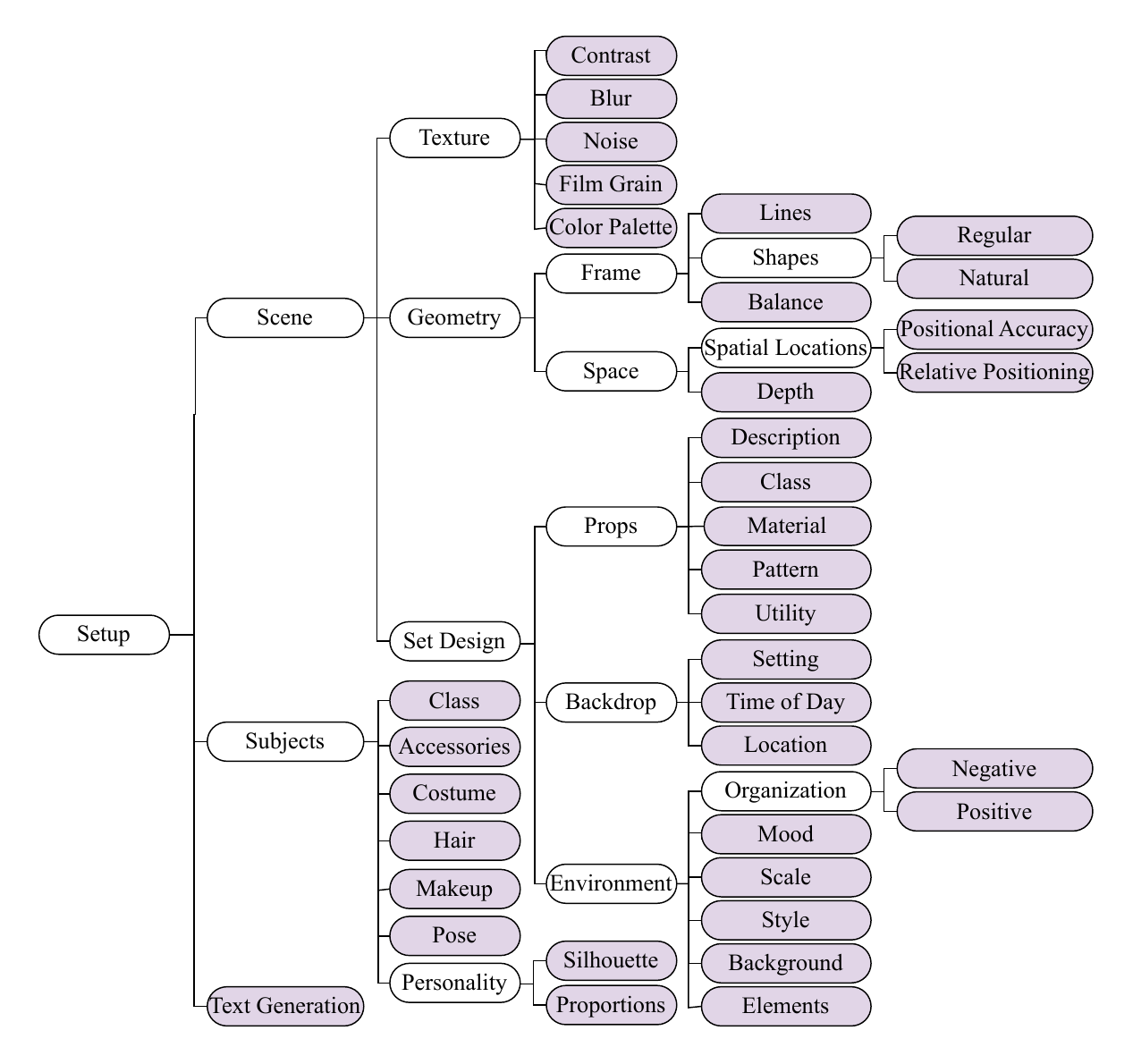}
    \subcaption{The \textbf{Setup} taxonomy outlines the visual components within the frame, including subjects, props, and environmental context.}\label{fig:1a}
  \end{minipage}
  \hfill
  \begin{minipage}[c]{0.48\linewidth}
    \centering
    \includegraphics[valign=t,width=\linewidth]{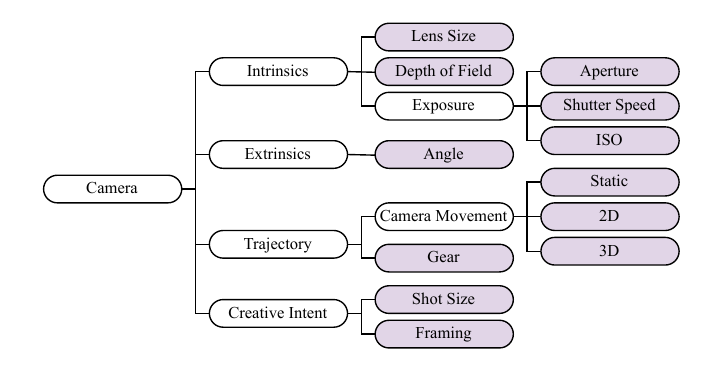}
    \subcaption{The \textbf{Camera} taxonomy defines all controls related to camera configuration during a shot setup.}\label{fig:1b}

    \vspace{0.2em} 

    \includegraphics[valign=t,width=\linewidth]{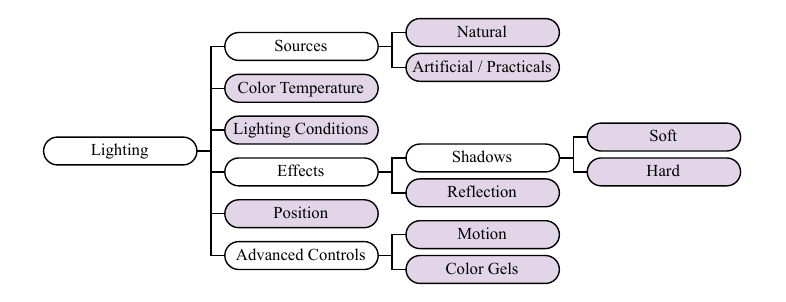}
    \subcaption{The \textbf{Lighting} taxonomy specifies the illumination of shot, through light sources, their properties, and their interaction with the scene.}\label{fig:1c}
  \end{minipage}

  \caption{\textbf{Setup}, \textbf{Camera} and \textbf{Lighting} Taxonomies that structure the visual elements of a shot.}
  \label{fig:taxonomy_triple}
\end{figure}

\section{Stable Cinemetrics}
\begin{wrapfigure}[14]{r}{0.5\columnwidth}
  \centering
  \includegraphics[width=\linewidth]{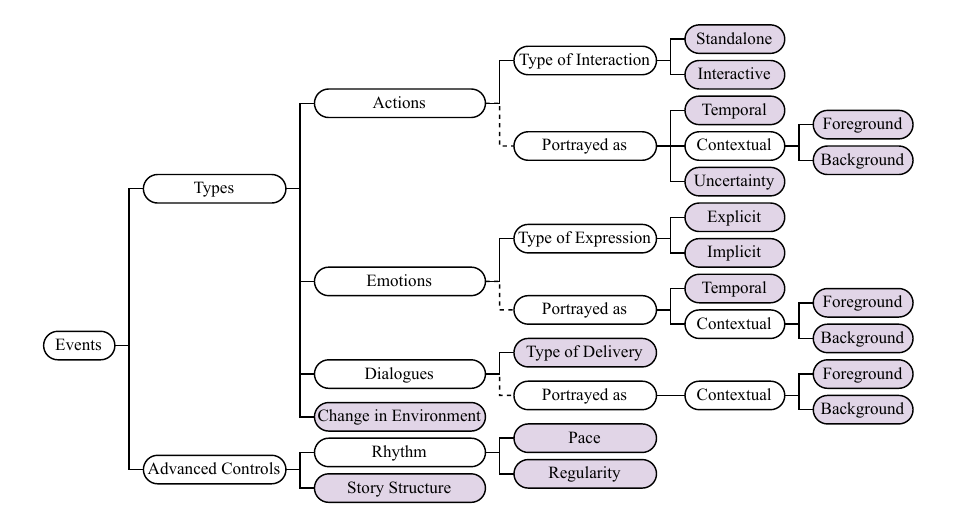}
  \caption{The \textbf{Events} taxonomy captures the narrative dimension of a shot which includes actions, emotions, and their fine grained portrayal as they evolve over time within a shot.}
  \label{fig:events}
\end{wrapfigure}

The subsequent sections detail our proposed taxonomy (Section \ref{sec:taxonomy_desc}), and its underlying design principles. Next, we develop our benchmark comprising of prompts designed for professional use (Section \ref{sec:prompts}). Section \ref{sec:question_gen} describes how  prompt categorization and question generation are performed, enabling large-scale evaluations of video generative models for professional use.

\subsection{Taxonomy Design} \label{sec:taxonomy_desc}

Our taxonomies are developed in iteration with industry professionals, including organizations established by the Big Five Studios \cite{studiobinder_major_studios}, independent cinematographers and screenwriters, and an Academy Award winning Visual Effects Artist. The central guiding question in our taxonomy development was: "\textit{What \textbf{controls} do \textbf{professionals} require when setting up a \textbf{shot}?}".

A \textbf{shot} is the atomic unit of filmmaking; an uninterrupted sequence without cuts in which cinematic meaning emerges from the coordination of multiple cinematic choices. The average shot length (ASL) in feature films is 5–10 seconds \cite{cutting2015shot}, which closely aligns with the temporal limits of current video generation models. This duration is not a limitation; it is a compact canvas where rich and \textit{enough} narrative and visual complexity can unfold. A single shot entails numerous controls to convey intent, emotion, and story. This motivates our decision to \textit{design the taxonomy at the shot level}, where fine-grained control is paramount.
\textbf{Control} is the key distinction between casual and professional video creation. In casual settings, users often accept model outputs with minimal intervention, delegating the \textit{key} creative decisions to the model. In contrast, professional use demands precise, deliberate control at every stage of the generative process. In fact, pixel-level control is common in film-making \cite{mediabee2024colorgrading, wood2007pixel},  underscoring the importance of fine-grained adjustments in achieving the desired visual effect.
\textbf{Professionals} such as cinematographers and directors are primarily responsible for defining a shot's creative intent. While filmmaking is collaborative, these roles offer a practical abstraction for modeling control. A key insight is that, despite overlap, they are sufficiently \textit{disentangled} to support distinct control dimensions: screenwriters rarely specify lighting or camera movement, and production designers are typically not associated with emotional tone or narrative pacing. These factors motivate the development of our 4 control pillars, each contributing to the composition of a shot. Our taxonomies are structured as hierarchical trees, where \texttt{leaf} nodes correspond to the most granular control parameters, each associated with a set of \underline{values}. We describe each pillar below:

\textbf{Setup}. Setup (Figure \ref{fig:1a}) encompasses all visible elements within the frame. We organize it into three top-level groups:
(1) \texttt{Scene} aggregates environmental controls, including \texttt{Texture}, \texttt{Geometry}, and \texttt{Set Design}. \texttt{Texture} covers aspects such as \texttt{contrast} and \texttt{color palette}, which govern the surface feel of the shot. \texttt{Geometry} captures dominant \texttt{shapes} and the \texttt{spatial arrangement} of elements within the scene. \texttt{Set Design} comprises \texttt{Props} and their \texttt{attributes}; the \texttt{Backdrop}, which establishes the macro context of the set; and \texttt{Environment}, which defines micro-level elements contributing to the "feel" of the shot.
(2) \texttt{Subjects} refer to the focal characters within a shot, defined by attributes such as \texttt{costumes} and \texttt{accessories}.
(3) \texttt{Text Generation} refers to on-screen typography such as titles or lettering, designed to appear as integrated graphical elements. Each node in Setup has cinematic meaning: a \underline{dawn} (\texttt{Time of Day}) setup combined with \underline{mist} (\texttt{Elements}) can suggest danger, while a \underline{clean} (\texttt{Organization}), \underline{symmetrical} hallway (\texttt{Balance}) conveys order. 

\textbf{Lighting}. \enquote{\textit{Lighting is the key to turning amateur footage into professional stories and presentation}} - Jay Holben \cite{holben2011shot}. Motivated by this principle, we define the following groups for Lighting (Figure~\ref{fig:1b}):
(1) \texttt{Source}, the origin of illumination within the shot;
(2) \texttt{Color Temperature}, which controls the warmth of the light;
(3) \texttt{Lighting Conditions}, preset configurations describing scene-wide illumination;
(4) \texttt{Effects}, visual outcomes resulting from light interacting with the scene;
(5) \texttt{Position}, the spatial relation of the light source to the subject; and
(6) \texttt{Advanced Controls} such as \texttt{flickering} modulation and the use of \texttt{color gels} to adjust lighting hue. Each control knob corresponds to distinct cinematic expressions: a shot with only a \underline{back light} (\texttt{Position}) evokes mystery, while \underline{hard} \texttt{shadows} are often used to amplify tension.
  
\textbf{Camera}. The camera taxonomy (Figure~\ref{fig:1c}) encompasses all camera-related control dimensions involved in a shot. We organize these into 4 high-level groups:
(1) \texttt{Intrinsics}: optical and exposure parameters governing the light captured by the camera;
(2) \texttt{Extrinsics}: position and orientation of the camera relative to the subject;
(3) \texttt{Trajectory}: motion of the camera and the supporting gear that enables it; and
(4) \texttt{Creative Intent}: compositional choices that shape the narrative or emotional tone of a shot.
Prior works have primarily focused on camera motion control \cite{he2025cameractrl, hou2025trainingfree}; however, we show that a much broader range of camera parameters can be independently manipulated while setting up a shot. Each parameter has tangible cinematic impact; for example, a \underline{shallow} \texttt{depth of field} can isolate the subject from the background to direct emotional focus while, an \underline{insert} \texttt{framing} spotlights narrative details with precision.

\textbf{Events}. Events (Figure \ref{fig:events}) encodes the narrative substance of a shot, namely the depicted \texttt{actions}, \texttt{emotions}, and \texttt{dialogues} - which are further decomposed into dependent nodes for fine-grained control. These dependent nodes represent attributes that cannot exist independently of their parent categories; for instance, these nodes can specify the type of interaction or the delivery mode of a dialogue.  Emotions may appear \texttt{explicitly} (\underline{visible tears}) or \texttt{implicitly} (\underline{a clenched jaw}), while actions can be \texttt{stand-alone} or \texttt{interactive}. The \texttt{Portrayed As} category captures aspects such as: \texttt{Temporal}, which refers to the unfolding pattern of the event (e.g., laughter erupting \texttt{simultaneously} vs. \texttt{sequentially}), and \texttt{Contextual}, which indicates whether the event occurs in the \texttt{foreground} or \texttt{background}. \texttt{Advanced Controls} refines pacing and the story structure of the shot, such as a \underline{turning point} or \underline{climax}. While recent works \cite{wang2024worldsimulatorgoodstory, feng2024tcbenchbenchmarkingtemporalcompositionality} evaluate T2V models on sequential event generation, we show that, from a professional standpoint, Events encompass a much broader and richer evaluative space.

\begin{table}[t]
\centering
\small
\caption{Structured prompt upsampling with SCINE taxonomies. We show how control nodes from our taxonomies enable the generation of visually expressive (\textit{SCINE Visuals}) prompts from narrative scripts (\textit{SCINE Scripts}). The table also demonstrates how a single script can yield multiple visual interpretations, enabled by our taxonomy guided prompt generation pipeline. This aligns with filmmaking principles, where a script can be visually realized in diverse ways depending on the creative choices made by the filmmakers.}

\resizebox{\linewidth}{!}{
\begin{tabular}{|p{2.7cm}|p{5.6cm}|p{5.6cm}|}
\hline
\multicolumn{3}{|c|}{{Baseline script:} \emph{A man serves dinner to his family.}}\\
\hline
{Taxonomy Branch} &
{Baseline choice $\rightarrow$ narrative impact} &
{Alternative choice $\rightarrow$ narrative impact}\\
\hline
 \texttt{Depth of Field} (Camera) &
{Shallow} $\rightarrow$ isolates food/serving hand, romantic warmth &
{Deep} $\rightarrow$ every family member equally sharp, ensemble clarity \\
\hline
 \texttt{Camera Movement} (Camera) &
{Static tripod + gentle dolly-in} $\rightarrow$ calm focus on gesture; subtle emphasis &
{Handheld tracking} $\rightarrow$ urgency, energetic family chaos\\
\hline
\texttt{Lighting Source} (Lighting) &
{Warm tungsten practicals} $\rightarrow$ cozy, inviting domestic glow &
{Cool morning daylight through windows} $\rightarrow$ brisk freshness and emotional distance\\
\hline
\texttt{Backdrop / Time of Day} (Setup) &
{Evening interior} $\rightarrow$ nostalgic comfort, winding-down mood &
{Bright morning interior} $\rightarrow$ optimism and upbeat tempo\\
\hline
\texttt{Props} (Setup) &
{Earth-tone wooden utensils} $\rightarrow$ homely &
{Silver cutlery} $\rightarrow$ formal, upscale\\
\hline
\multicolumn{3}{|p{14cm}|}{%
  \begin{minipage}{14cm}
{Upsampled prompt:}
\emph{A man serves dinner to his family with shallow depth of field on a static tripod with a gentle dolly-in, under warm tungsten interior lighting in the evening,  in a cozy earth-tone kitchen with wooden utensils.}
  \end{minipage}%
}
\\
\hline
\end{tabular}
}
\label{tab:taxonomy_prompt_mapping}
\end{table}

The taxonomies define a total of 76 leaf-level controls that can be independently adjusted when crafting a shot. We structure our taxonomies as hierarchical trees to enable disentanglement and multi-level abstraction of cinematic controls. Attributes within each branch are highly correlated, while branches remain independent, ensuring, for example, adjusting \texttt{Depth of Field} does not affect \texttt{Camera Movement}. The tree structure naturally supports multi-level abstraction, aligning with how filmmakers conceptualize scenes, starting from high-level intent and refining toward specific implementation. For example, a directive such as “set a tense alley at night” can be decomposed into an \underline{EXT} \texttt{Setting}, a \underline{cool} \texttt{Color Temperature} and a \underline{Deep} \texttt{Depth of Field}. This structure also allows easier scalability; adding a new detail like \underline{floating ember sparks}, fits cleanly under (\texttt{Environment → Elements}) without disrupting the rest of the taxonomy. An alternate structure such as a flat, linear list would not support such extensibility.

Developing the taxonomy is a non-trivial task. This required multiple iterations with experts since professionals interpret and prioritize controls differently. Furthermore, shot creation is a multi-stage process, starting from script-writing to setup design to camera blocking, making unification under a single structured framework challenging. While some taxonomies \cite{movielabs2024images} focus only on filmmaking aspects, they lack structure and are not aligned with generative modeling. Our goal instead was to impose structure, and develope a taxonomy that bridges professional filmmaking and generative video models. Details of each \texttt{control} node and their corresponding \underline{values} are provided in Appendix \ref{sec:taxonomy}.

\subsection{Designing Prompts for Professional Use} \label{sec:prompts}
\begin{wrapfigure}[20]{r}{0.50\textwidth}
  \centering
  \includegraphics[width=\linewidth]{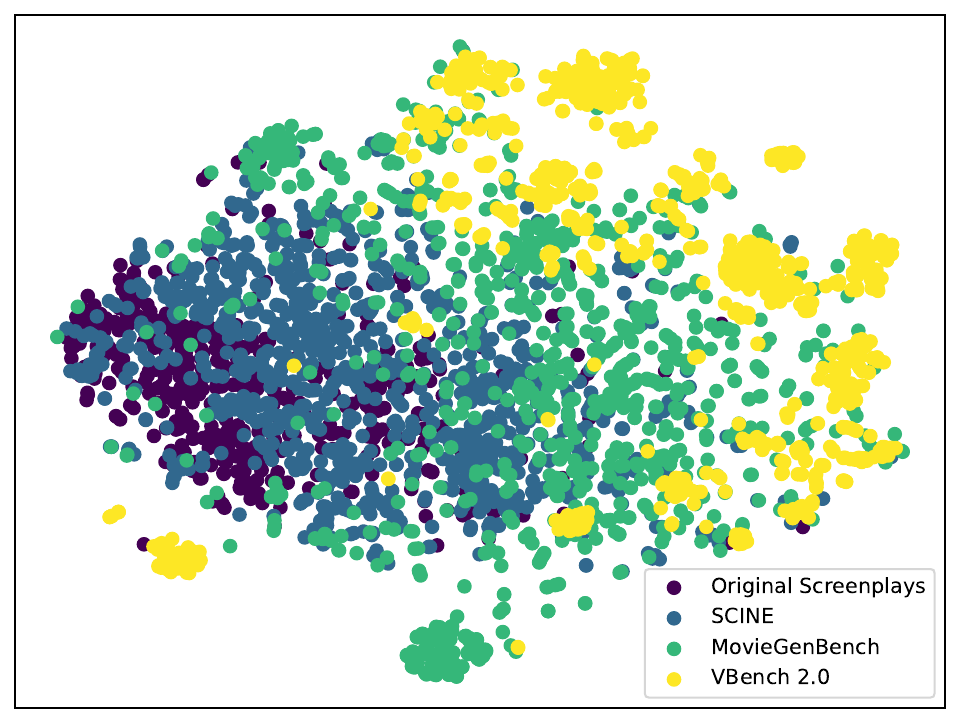}
  \caption{t-SNE visualization showing substantial overlap between ground truth screenplays and prompts in \textit{SCINE Scripts}, in comparison to existing prompt based benchmarks such as VBench-2.0}
  \label{fig:tsne}
\end{wrapfigure}
The taxonomies form the foundation for constructing prompts tailored to professional use. Our core approach in creating prompts involves sampling \underline{values} from the \texttt{control} nodes and creating prompts that reflect realistic cinematic intent. Mirroring the filmmaking process, we first generate narrative scripts and then inject visual elements into these scripts : 

\textbf{Scripts.} These prompts, referred to as \textit{SCINE-Scripts}, contain the narrative content of individual shots. We collaborate with a professional screenwriter to create seed prompts that meet strict constraints: a single shot, under 10 seconds, with no reliance on off-screen elements. These seed prompts, along with sampled nodes from the \underline{Events} taxonomy are provided as input to a LLM, for prompt generation. We use the Events taxonomy for these prompts because it directly encodes narrative beats i.e., what happens in a shot. It covers nodes such as physical dynamics (\texttt{actions}) and verbal interactions (\texttt{dialogues}), which are crucial to the story conveyed in a shot. To ensure prompt diversity, we vary parameters across multiple LLM invocations, sampling emotions from Plutchik’s model \cite{plutchik1991emotions}, alternating actions, dialogue structures, genres, and subject composition. We use LLMs, as prior work \cite{mirowski2022cowritingscreenplaystheatrescripts, tang2025understandingscreenwriterspracticesattitudes} have shown their effectiveness in screenplay generation. t-SNE visualization (Figure \ref{fig:tsne}) of \textit{SCINE-Scripts} embeddings \cite{wang2024multilinguale5textembeddings} shows substantial overlap with ground-truth screenplays \cite{saxena2024moviesumabstractivesummarizationdataset}, whereas prompts from VBench-2.0 \cite{zheng2025vbench20advancingvideogeneration} and MovieGenBench \cite{polyak2025moviegencastmedia} exhibit minimal overlap. 

\textbf{Visual Exposition.} We refer to these set of prompts as \textit{SCINE-Visuals}, which enrich \textit{SCINE-Scripts} with visual elements from the \underline{Camera}, \underline{Lighting}, and \underline{Setup} taxonomies. In contrast to \textit{Events}, these taxonomies offer fine-grained control over the visual style and composition of a shot. For each base prompt in \textit{SCINE-Scripts}, we sample values from one or more control nodes and inject them to expand the prompt with structured visual specifications. \textit{SCINE-Visuals} highlight a key advantage of our taxonomy: structured prompt upsampling. Unlike existing prompt upsampling techniques that delegate all creative decisions to the LLM, our method constrains generation within the taxonomy, enabling more controlled and interpretable prompt expansion. Table~\ref{tab:taxonomy_prompt_mapping} provides a breakdown of how \textit{SCINE-Scripts} is subsequently upsampled via the taxonomy to generate \textit{SCINE-Visuals}.

\subsection{Category and Question Generation} \label{sec:question_gen}
Next, we extract \textit{categories} and generate \textit{questions} for each prompt. Categories link prompts back to the taxonomy, allowing fine-grained evaluations across different levels of abstraction; a single prompt can map to multiple categories across different taxonomies. For each category, we generate targeted questions that are shown to human annotators during video evaluations. These questions are specific in nature and target only a single control node, enabling its isolated evaluation. Unlike high-level prompt adherence questions, which lack fine-grained attribution, our framework supports per-control annotations. A minimal example is shown below : 

\begin{minipage}{\linewidth}
\small     

\textbf{Prompt:}
A tight close-up focuses on a fireplace, its embers flickering brightly.

\vspace{0.35em} 
\begin{itemize}[leftmargin=*,nosep]
  \item \textbf{Category} : {Lighting → Advanced Controls → Motion} | 
        \textbf{Question} : Does the scene exhibit dynamic flickering effects in its lighting that align with the description?
  \item \textbf{Category} : {Camera → Creative Intent → Shot Size} |  
        \textbf{Question} : Does the video include a tight close‑up shot that captures the detailed framing?
\end{itemize}
\end{minipage}

\begin{figure}[t]
  \centering
  \includegraphics[width=0.75\linewidth]{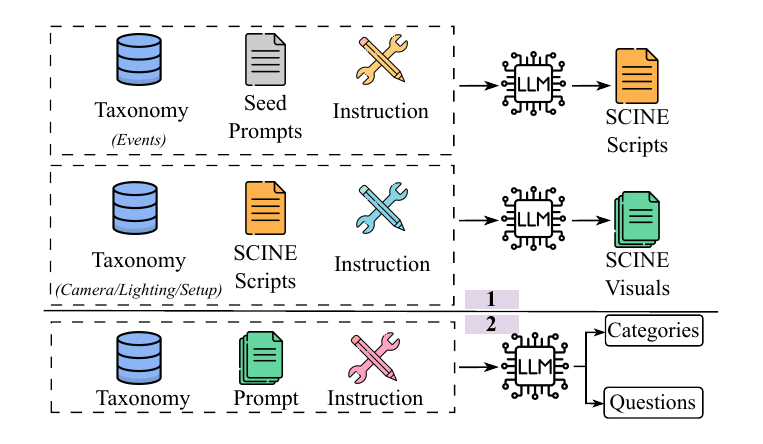}
  \caption {\textbf{1. Prompt Generation Pipeline} SCINE Scripts are created by passing seed prompts and sampled \underline{Events} taxonomy nodes to an LLM, forming the narrative component of our benchmark. SCINE Visuals are then generated through structured upsampling, where nodes from the \underline{Camera}, \underline{Lighting}, and \underline{Setup} taxonomies are sampled and injected into each SCINE Script to create prompts that capture visual exposition. 2. \textbf{Automatic Categorization and Question Generation} Given a SCINE prompt and taxonomy, we \textit{categorize} each taxonomy element present in the prompt and generate a corresponding \textit{question} to enable isolated evaluation of each control node.}
  \label{fig:eval_pipeline}
\end{figure}

The overall evaluation pipeline, depicting prompt generation, categorization and question generation, is presented in Figure \ref{fig:eval_pipeline}. Additional details are presented in Appendix \ref{sec:prompts_gen}.

\section{Are current Video Generative Models Ready for Professional Use?}\label{sec:t2v}
We now evaluate state of the art text-to-video (T2V) models against the professional standards defined in our taxonomy pillars. Our analysis reveals both strengths and persistent challenges of current models, offering an overview of how current models align with professional quality expectations.

\begin{figure}[]              
  \centering                   
  \begin{subfigure}[t]{0.48\linewidth}
    \centering
    \includegraphics[width=\linewidth]
    {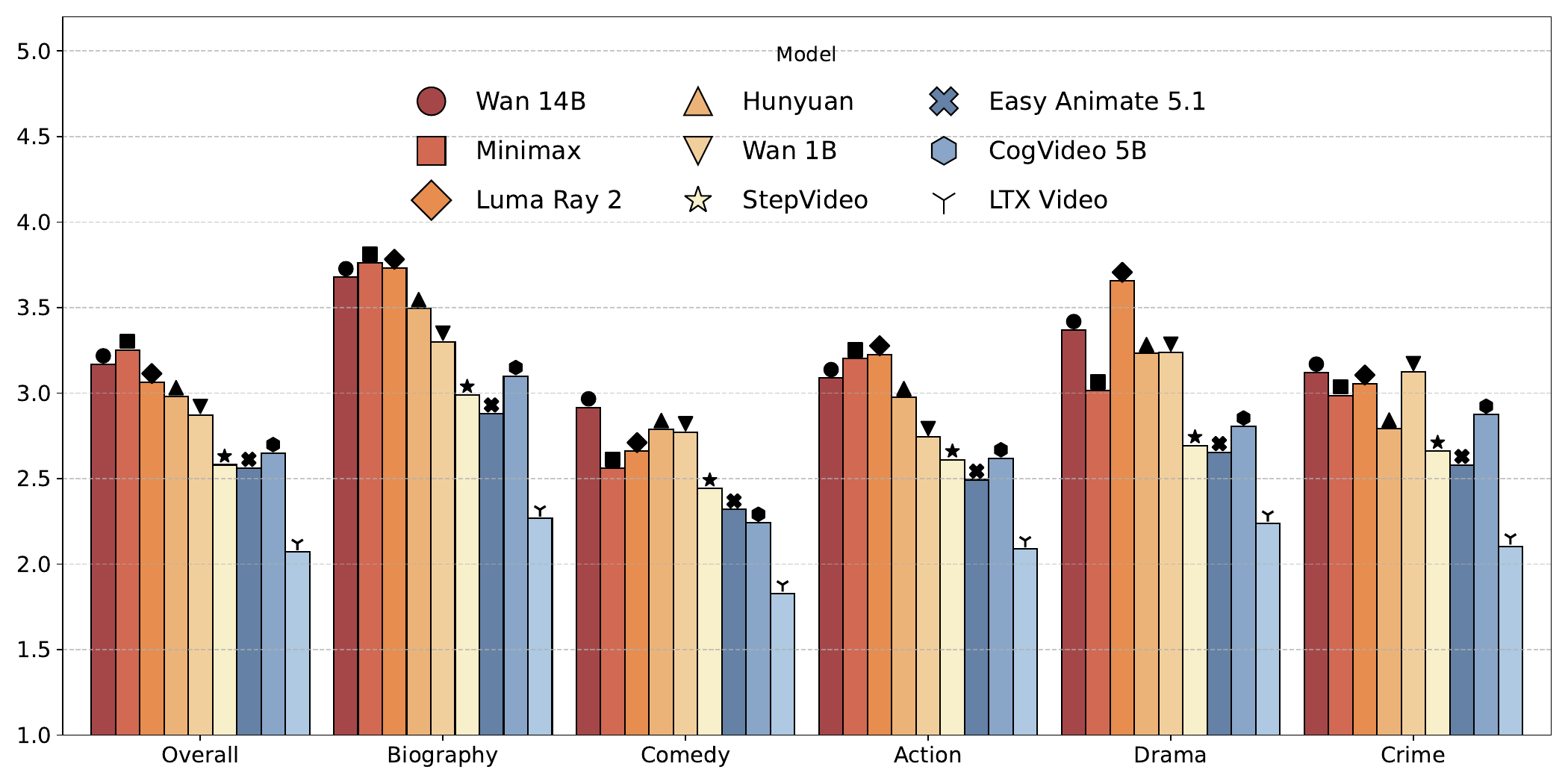}
    \caption{Overall results on \textit{SCINE Scripts}, across genres. Minimax and Wan 14B emerge as the strongest models. The strongest genre is Biography while all models consistently struggle at Comedy.}
    \label{fig:scene_scripts_overall}
  \end{subfigure}
  \hfill                         
  \begin{subfigure}[t]{0.48\linewidth} 
    \centering
    \includegraphics[width=\linewidth]
    {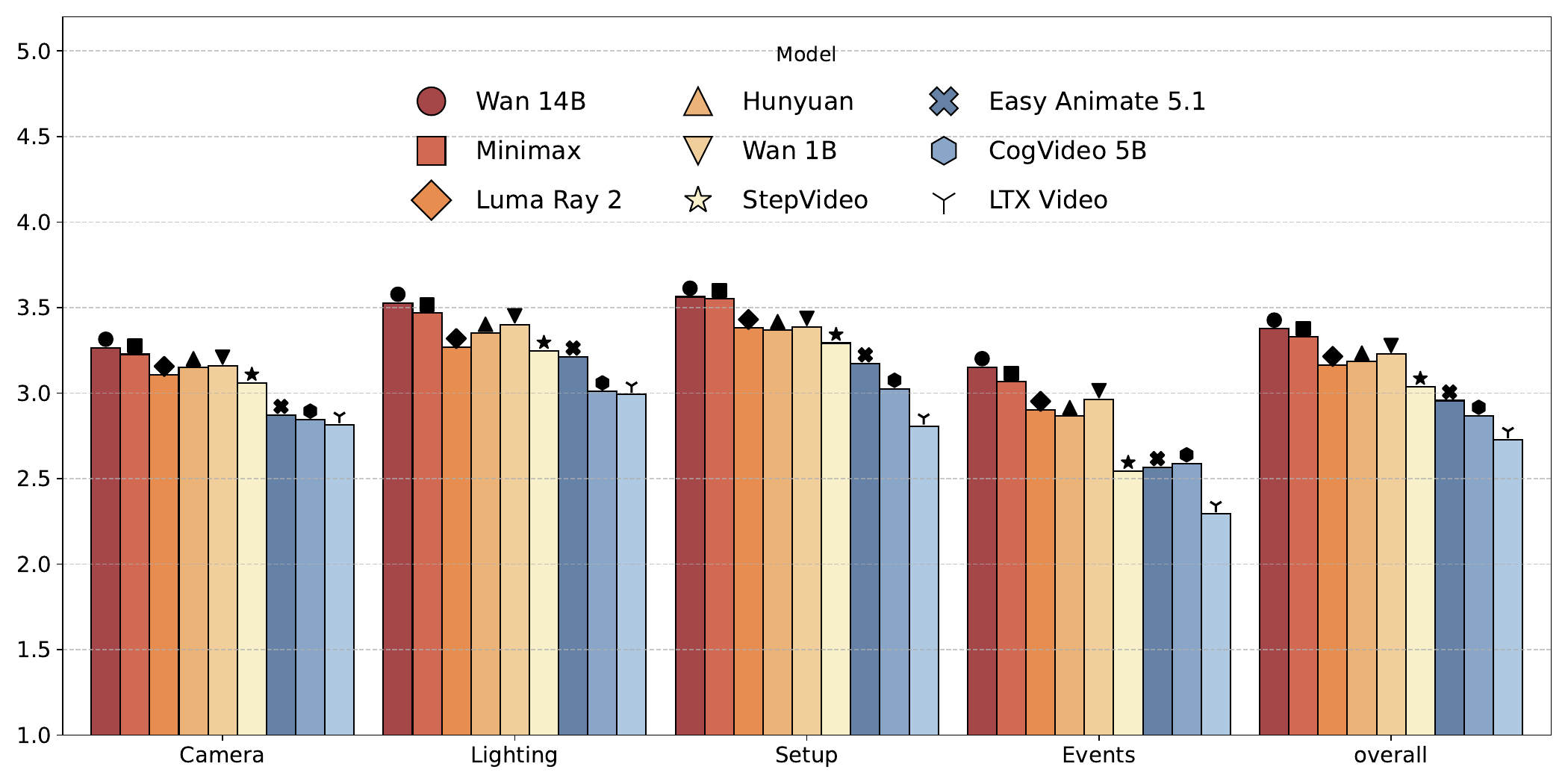}
    \caption{Overall results on \textit{SCINE Visuals} across four pillars of professional control. \textit{Events} emerges as the most challenging category across models, while \textit{Setup} yields the strongest performance.}
    \label{fig:scine_visuals_overall}
  \end{subfigure}
              
  \caption{Overall results on SCINE \textit{Scripts} and \textit{Visuals}. }
  \label{fig:overall_results}
\end{figure}

\subsection{Experimental Setup} 

\textbf{Prompts.}
The SCINE benchmark comprises two prompt categories, \textit{Scripts} and \textit{Visuals}, each aligned with distinct professional roles (Table \ref{tab:scine_groups}). The \textit{Visuals} prompts are created by systematically upsampling \textit{Scripts} using our taxonomies leading to a total of 2,089 prompts. We categorize prompts by difficulty: we create basic prompts by limiting the number of sampled control nodes, while in advanced prompts, we do not impose any restriction.

\textbf{Models.} We evaluate 13 state-of-the-art T2V models, both open-source (WAN 1B/14B \cite{wan2025wanopenadvancedlargescale}, Hunyuan-Video \cite{kong2025hunyuanvideosystematicframeworklarge}, Step Video \cite{ma2025stepvideot2vtechnicalreportpractice}, CogVideoX 5B \cite{yang2025cogvideox}, LTX-Video \cite{hacohen2024ltxvideorealtimevideolatent}, Pyramid Flow \cite{jin2025pyramidalflowmatchingefficient}, Easy Animate 5.1 \cite{xu2024easyanimatehighperformancelongvideo}, Mochi \cite{genmo2024mochi}), VChitect-2.0 \cite{fan2025vchitect20paralleltransformerscaling} and closed source (Minimax \cite{minimax2024video}, Luma Ray 2 \cite{lumalabs2025ray2}, Pika 2.2 \cite{pikalabs2025pika2.2}). Our goal is to assess each model’s suitability for role-specific professional tasks, like evaluating narrative fidelity from a screenwriter’s perspective.  Unless otherwise noted, we use default sampling parameters and maintain a consistent seed per prompt, across models for fair comparison. 

\begin{wraptable}[11]{r}[0pt]{0.55\textwidth}
\vspace{-\baselineskip}  
  \centering
  \caption{The \textbf{SCINE} benchmark includes prompts tailored to professional roles, where each prompt is paired with multiple, fine grained evaluation questions.}
  \resizebox{\linewidth}{!}{
  \begin{tabular}{@{} l  l  c  c @{}}
    \toprule
    \makecell[l]{Target Role} 
      & \makecell[l]{Target Taxonomies} 
        & \makecell[c]{\# of\\Prompts} 
          & \makecell[c]{Avg. Questions per\\Prompt} \\
    \midrule
    \multicolumn{4}{c}{\itshape SCINE‐Scripts} \\
    Screenwriters 
      & Events 
        & 1133 
          & $2.57\pm0.98$ \\
    \midrule
    \multicolumn{4}{c}{\itshape SCINE‐Visuals} \\
    Cinematographers 
      & \makecell[l]{Camera, Lighting} 
        & 355 
          &$5.42\pm4.47$ \\
    Production Designers 
      & Setup 
        & 298 
          &$4.00\pm3.35$ \\
          
    Directors 
      & \makecell[l]{All} 
        & 303 
          & $10.48\pm5.07$ \\
    \bottomrule
  \end{tabular}
  }
  \label{tab:scine_groups}
\end{wraptable}

\textbf{Human Annotation Setup.} To ensure high-quality evaluation, we work with a pool of 84 expert annotators with an average of 6.5 years of experience in film production, across roles such as cinematographers, film editors, screenwriters, visual communication designers, and directors. Annotators were shown a prompt along with two generated video samples. For each prompt, they were presented with taxonomy derived evaluation categories and corresponding questions. Each video was rated independently on a 1–5 scale, where 1 indicated complete misalignment with the category and 5 indicated a perfect match. Although the evaluation was non-comparative, our UX ablation studies showed that displaying two videos side by side improved annotator calibration, especially when selecting middle range scores. To promote consistency and reduce subjectivity, we developed a comprehensive annotation guide covering each control node in the taxonomy. We collect 3 votes for every video-question pair across  \textbf{13,457} unique questions, collecting a total of \textbf{248,536} pairwise annotations. We observe an intra-class correlation coefficient (ICC) \cite{c0b394f2-cc2b-323c-8a63-399792baabe1} of 80.4\% for the 1-5 ratings at the model-pair level, and 95.5\% when the models are considered individually. Further, we conduct Wilcoxon signed-rank tests \cite{wilcoxon1992individual} across 45 model pairs where we observe statistically significant preferences in 37 of them, which highlights that the annotators agree on model preferences. 

\subsection{Results and Analysis} 
\begin{wrapfigure}[16]{}{0.5\textwidth}
    \centering
    \includegraphics[width=0.5\textwidth]{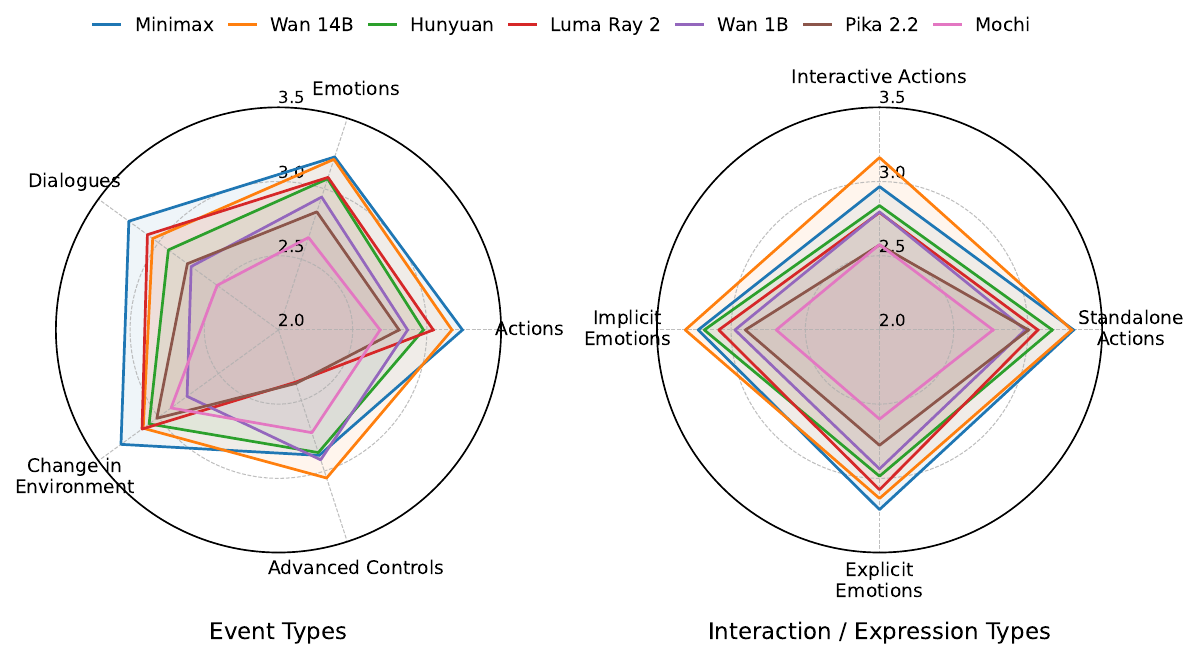}
 \caption {Fine-grained evaluation on \textbf{Events}. Models handle environmental changes well but struggle with dialogues and shot pacing. Standalone actions outperform interactive, and implicit emotions are easier than explicit.}
  \label{fig:events_zoomed_in}
\end{wrapfigure}

\begin{figure}[]
  \centering
  \includegraphics[width=\linewidth]{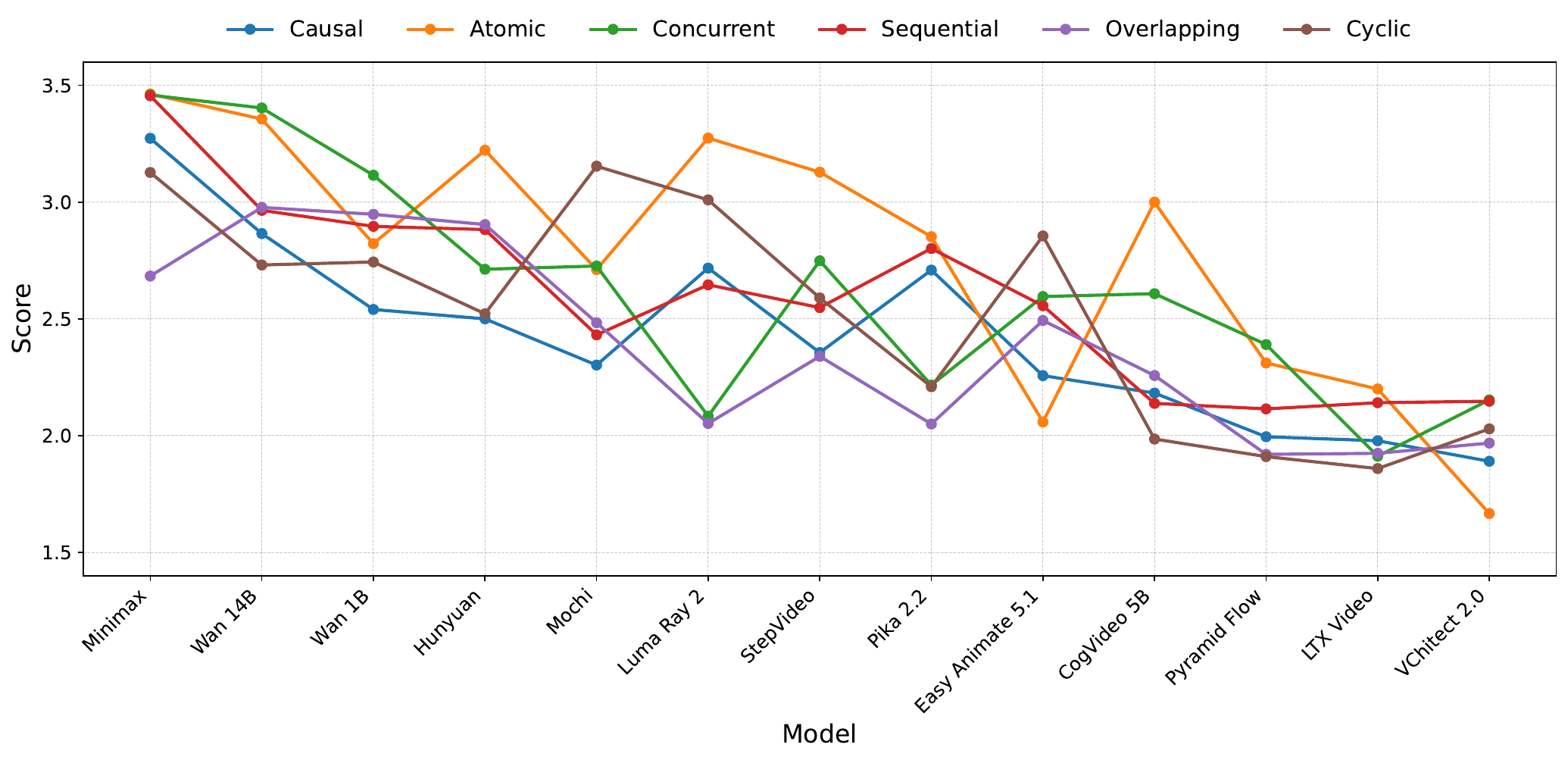}
  \caption {\textbf{Model performance on Events across temporal portryal of Actions}. Atomic actions are handled well, whereas models struggle with causal and overlapping Events.}
  \label{fig:models_actions}
\end{figure}

\paragraph{SCINE Scripts} We first evaluate models on narrative event generation, i.e. the \textit{story} dimension of a shot. Figure~\ref{fig:scene_scripts_overall} compares model performance across different genres, focusing on the accuracy and coherence of generated events. Minimax and WAN-14B emerge as the overall top performers, while LTX-Video consistently underperforms. We observe that models generally perform better on the Biography genre, whereas Comedy proves challenging.  In Figure ~\ref{fig:events_zoomed_in}, we zoom into sub-categories within \textit{Events~$\rightarrow$~Types}. Minimax leads in nearly all categories, showing the largest margins in \textit{Dialogues} and \textit{Change in Environment}, 
 but falls short in \textit{Advanced Controls}, where WAN-14B outperforms.  Further, models are better at stand-alone actions compared to interactive actions and portray implicit emotions better than explicit ones. Within Event Types, \textit{Actions} exhibit lowest variance across models. Figure \ref{fig:models_actions} shows Events performance across 13 models and 6 Temporal portrayal of Actions. Models handle atomic and concurrent actions well, but struggle with causal, overlapping, and cyclic events. Performance on Causal and Sequential events is highly correlated ($\rho=0.94$), as is performance on Concurrent and Overlapping events ($\rho=0.86$) across models. Despite variation across models, all show limitations in multiple \textit{Event} aspects, highlighting opportunities for improvement. 

\paragraph{SCINE Visuals}

Figure~\ref{fig:scine_visuals_overall} demonstrates overall results on SCINE Visuals. Consistent with the pair-wise rankings in Figure~\ref{fig:teaser}, WAN-14B and Minimax emerge as top performers across all pillars.
We find that current models struggle most with \textit{Events} and \textit{Camera}, while elements of \textit{Setup} and \textit{Lighting} are comparatively easier to capture. Only the top three models- WAN-14B, WAN-1B, and Minimax - reliably depict \textit{Events}, with a substantial performance gap from the rest. While \textit{Camera} scores are low across the board, the narrow spread suggests that all models face similar limitations. \textit{Lighting} shows the most consistent performance, with most models achieving relatively high scores whereas \textit{Setup} yields the highest absolute scores for the top-performing models. 

 \begin{wrapfigure}[13]{r}{0.5\textwidth}
 \vspace{-\baselineskip}  
  \centering
  \includegraphics[width=0.5\textwidth]{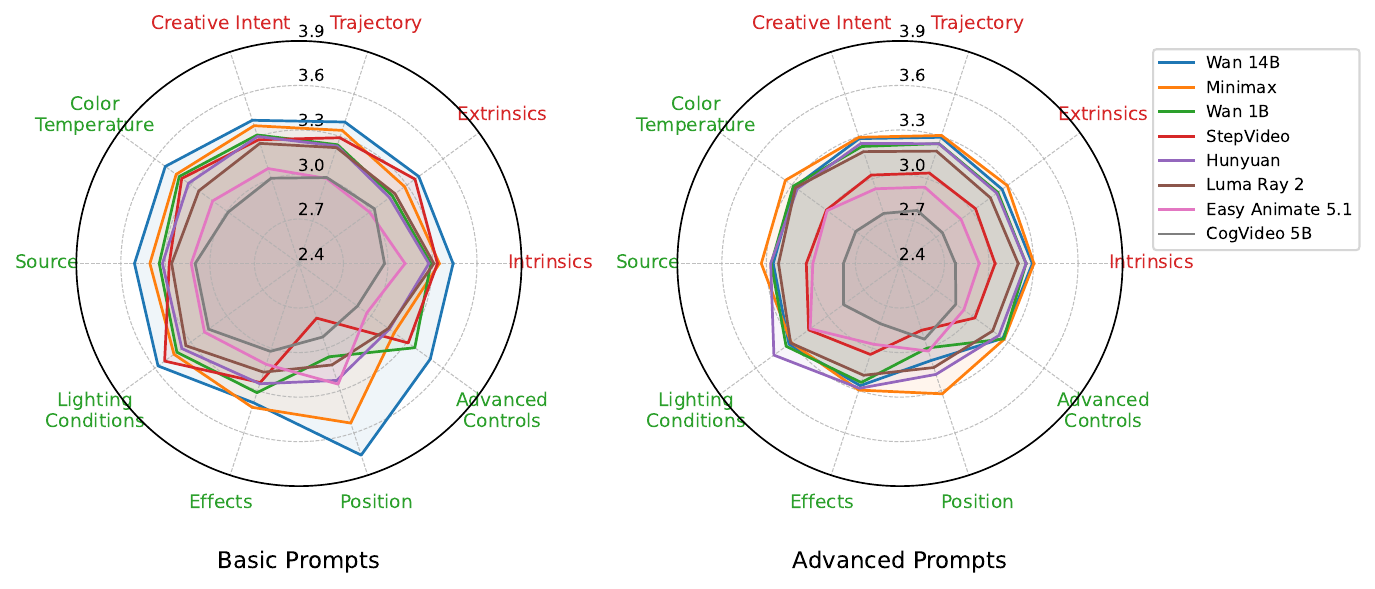}
  \caption{Split results on basic vs.\ advanced prompts for \textbf{Camera} and \textbf{Lighting}. All models show performance drops on advanced prompts, with the largest decline in Lighting Source.}
  \label{fig:easy_hard_camera}
\end{wrapfigure}

\begin{figure}[]
  \centering
  \includegraphics[width=\linewidth]{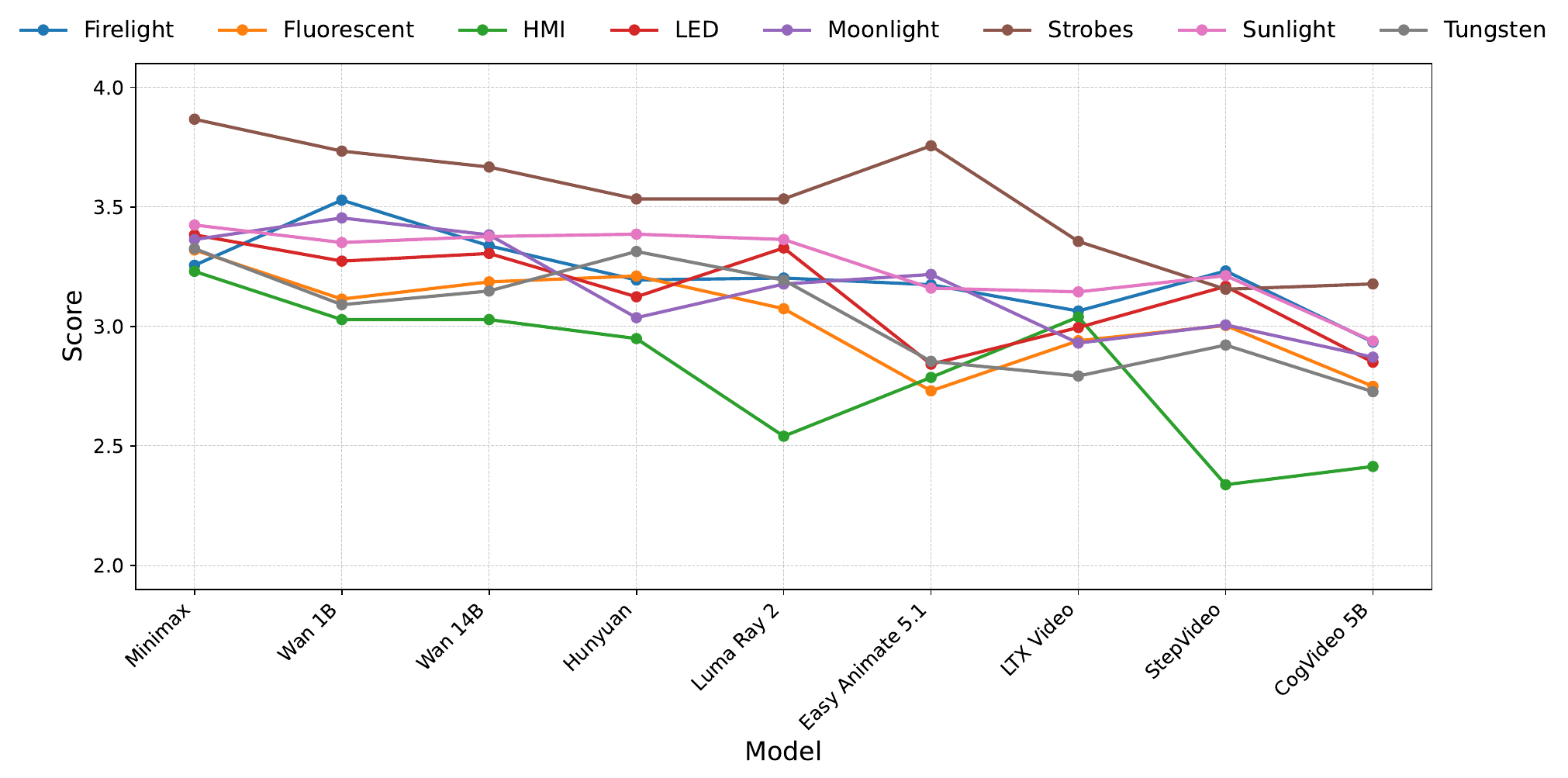}
  \caption {\textbf{Model performances across Lighting Source}. Strobes and Sunlight emerge strongest, whereas HMI and Fluorescent are points of weaknesses.}
  \label{fig:lighting_source_values}
\end{figure}
\begin{figure}[]
  \centering
  \includegraphics[width=\linewidth]{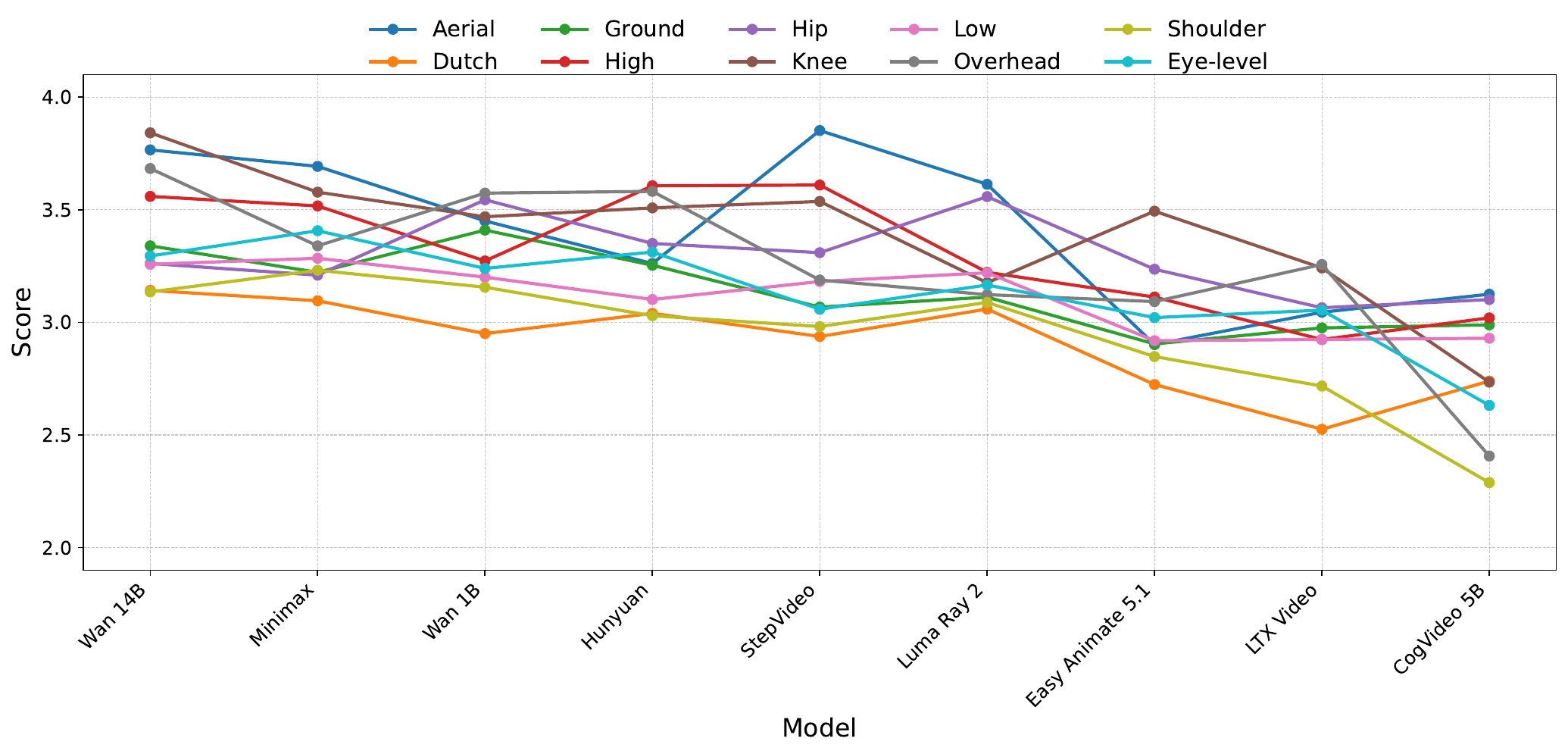}
  \caption {\textbf{Model performances across Camera Angles}. The Dutch angle poses a common challenge to all current video generative models}
  \label{fig:camera_angle_values}
\end{figure}
\begin{figure}[]
  \centering
  \includegraphics[width=\linewidth]{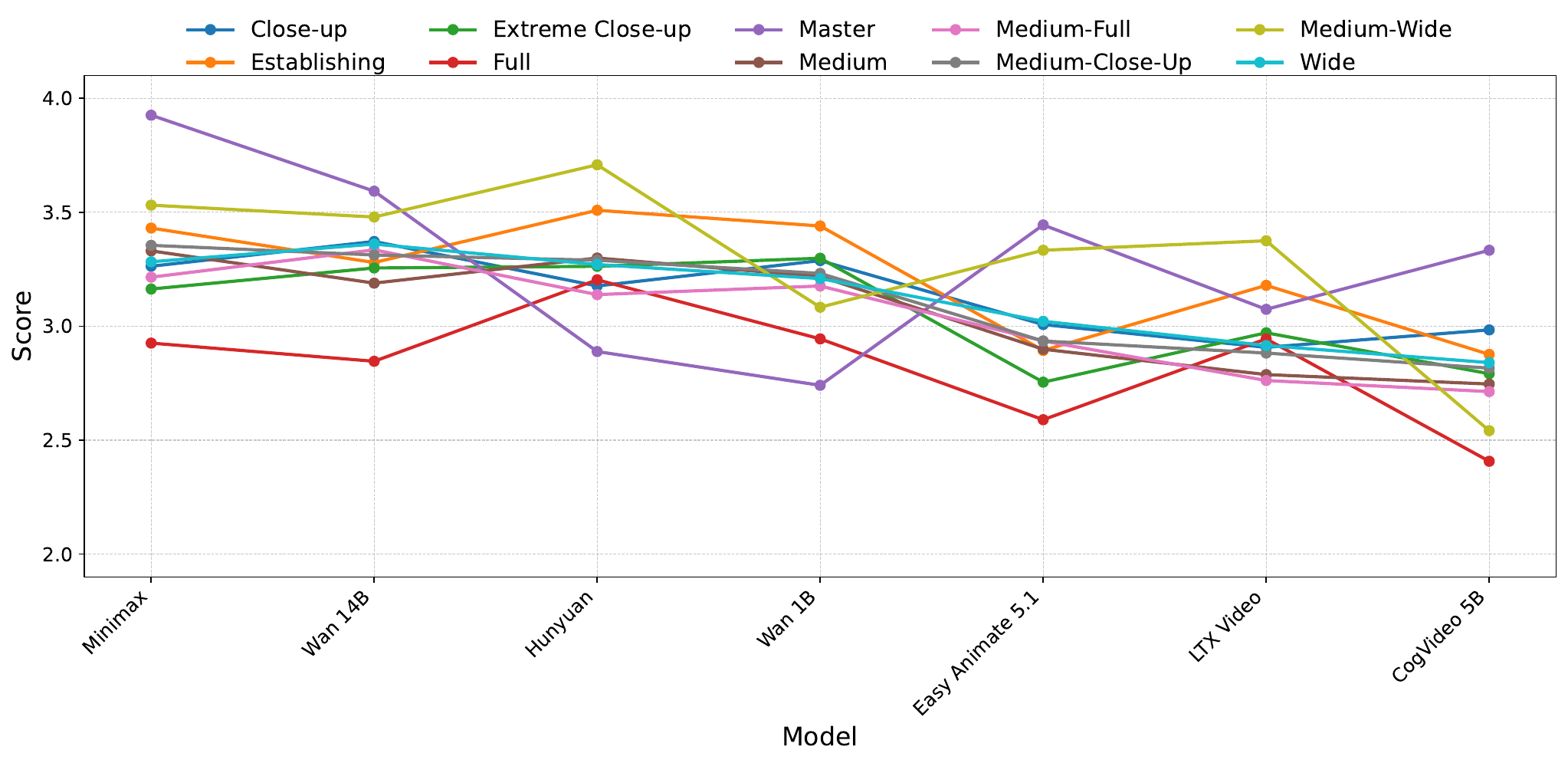}
  \caption {\textbf{Model performances across Camera Shot Size}. Models perform well on Master and Establishing shots and struggle at medium-wide and extreme-close-up shots.}
  \label{fig:camera_shot_size_values}
\end{figure}

 \textbf{Cinematographer.} We evaluate this role by creating prompts that inject control nodes from the \textit{Camera} and \textit{Lighting} taxonomy. Within \textit{Camera}, \textit{Extrinsics} and \textit{Trajectory} have the lowest average performance and the narrowest inter-model spread. For \textit{Lighting}, the primary bottleneck is \textit{Lighting Position}. We further present results, split by prompt difficulty in Figure \ref{fig:easy_hard_camera}. Across all models, performance degrades on advanced prompts, indicating that under conditions resembling professional workflows where cinematographers have a large amount of control, current models struggle. The biggest performance drops occur in \textit{Lighting Source},\textit{ Color Temperature}, and \textit{Creative Intent}. \textit{Lighting Position} and \textit{Advanced Controls} show the smallest performance drops, but remain the weakest categories overall, highlighting persistent challenges regardless of prompt complexity. Hunyuan and WAN-1B exhibit consistent performance across complexity levels. We also probe performance at the \underline{value} level. On Lighting Source (Figure \ref{fig:lighting_source_values}), Sunlight, Strobes, and Firelight are handled more reliably, while HMI, Fluorescent, and Tungsten lighting show lower performance. As shown in Figure \ref{fig:camera_angle_values}, Aerial and Knee level camera angles are depicted better, while the Dutch and Shoulder-level angles show lower performance. On Shot Sizes (Figure~\ref{fig:camera_shot_size_values}), Medium-Wide and Master shots have stronger performance in comparison to Full and Extreme Close-Up shots.

\begin{figure}[t]
  \centering
  \begin{subfigure}[c]{0.48\columnwidth}
    \centering
    \includegraphics[width=\linewidth]
    {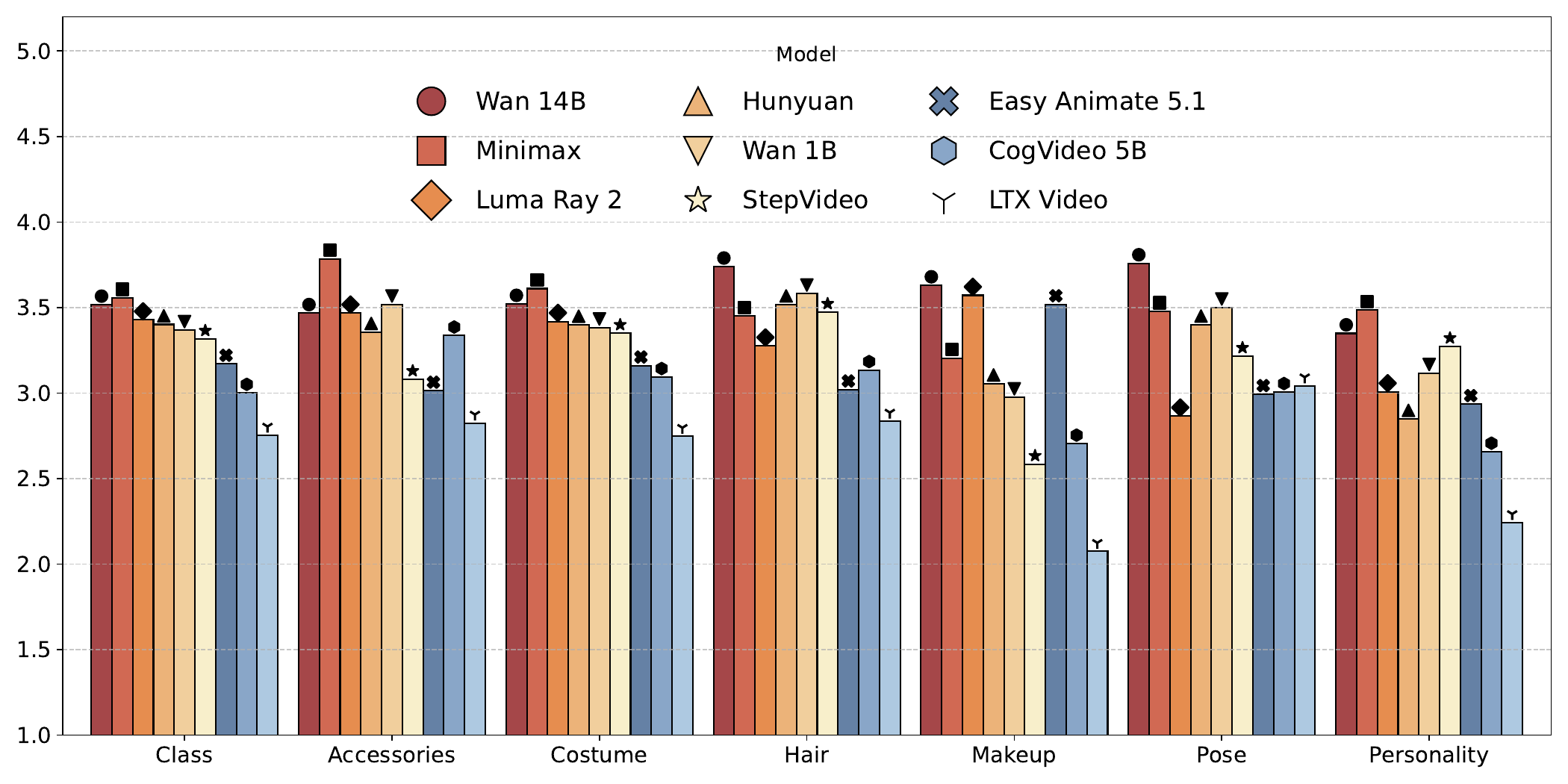}
    \caption{Fine-grained results on \textbf{Subjects}. Models perform well on hair and accessories, but struggle with personality and makeup.}
    \label{fig:subjects_zoom_in}
  \end{subfigure}
  \hfill
  \begin{subfigure}[c]{0.488\columnwidth}
    \centering
    \includegraphics[width=\linewidth]
    {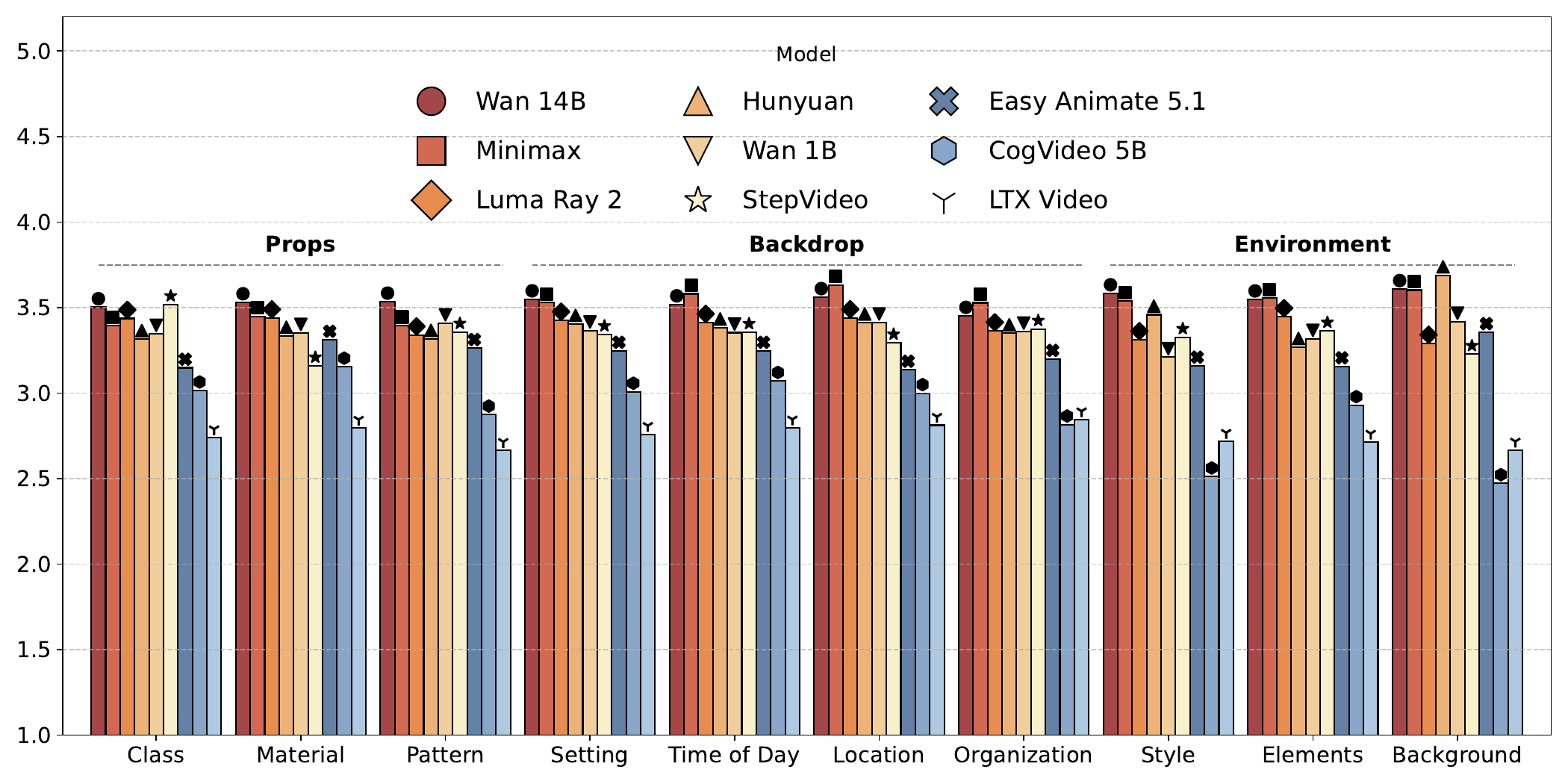}
    \caption{Fine-grained results on \textbf{Set Design}. Models perform better at  Backdrop in comparison to Environment, and struggle most with styling the shot appropriately.}
    \label{fig:set_design_zoom_in}
  \end{subfigure}
  \caption{Fine-grained results on \textbf{Setup} across Subjects and Set Design.}
  \label{fig:twopanel}
\end{figure}

\textbf{Production Designer.} Models perform strongest on the \textit{Setup} pillar; within \textit{Setup}, models achieve comparable performance on \textit{Subject} and \textit{Scene} generation, but show a drop in \textit{Text Generation}. In \textit{Subjects} (Figure \ref{fig:subjects_zoom_in}), model performances largely vary, with the highest scores for \textit{Hair} and \textit{Accessories}, and the weakest in \textit{Personality} and \textit{Make-up}.  In \textit{Set Design} (Figure \ref{fig:set_design_zoom_in}), performance trends follow the order: \textit{Backdrop}  \textgreater \textit{Props} \textgreater \textit{Environment}. Within \textit{Props}, models perform well in \textit{Material} but struggle at generating intricate \textit{Patterns}. For \textit{Environment}, current limitations lie in adhering to a coherent \textit{Style} and \textit{Backgrounds}, whereas better performance is seen in organizing \textit{Space} within a frame.

\begin{figure}[]
  \centering
  \includegraphics[width=0.9\linewidth]{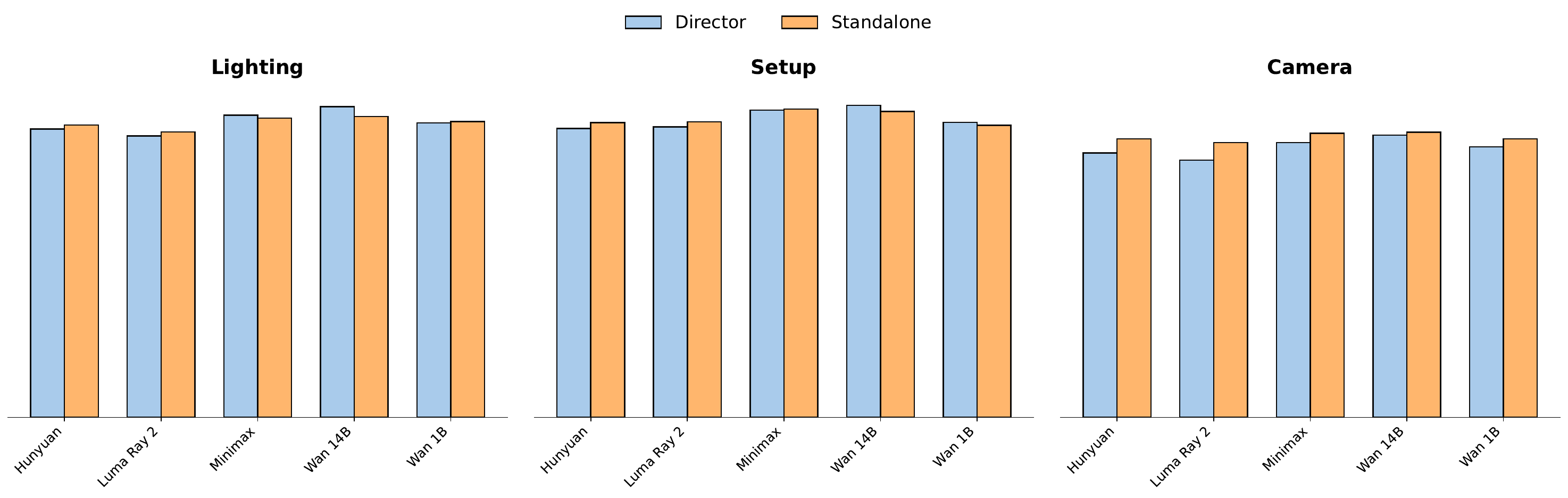}
  \caption {\textbf{Director} results: Joint specification of all controls, mirroring real-world shot creation leads to a performance drop on all models, compared to evaluation in a standalone manner.}
  \label{fig:director}
\end{figure}

\textbf{Director.} Prompts targeting this role differ from prior categories as they evaluate models across all taxonomies simultaneously. Model performance declines on average when all controls are defined jointly (Figure \ref{fig:director}). The largest performance drop is observed in \textit{Camera}, followed by \textit{Setup} and \textit{Lighting}. Wan 14B is the only model to show improved performance on \textit{Director} prompts for Lighting and Setup, compared to its \textit{Standalone} results.

Our evaluation identifies a three-tier hierarchy among current T2V models: Minimax and WAN-14B at the top, followed by Luma Ray 2, Hunyuan, and WAN-1B, with the remaining models forming the third tier. While overall performance varies, most models struggle with the fine-grained elements critical to professional video generation. For example, atomic events are handled reasonably well, but models falter on concurrent and causal events, which demand deeper temporal reasoning. Similarly, high-level cues like lighting conditions are better captured than nuanced aspects like precise light positioning. In summary, even top performing models exhibit substantial room for improvement across all dimensions of our taxonomy. No model achieves consistently strong performance across all aspects of shot composition, underscoring the challenge of aligning generative video models with professional standards. Additional results and analysis are presented in Appendix \ref{sec:additional_results}.

\section{Scalable Evaluation of Professional Videos}
In the previous section, we evaluated video generative models using expert annotations across 76 control nodes defined by our taxonomy. While human evaluation remains the gold standard, it is costly and difficult to scale, and defining reliable automatic metrics for each control node is non-trivial. Recent advances in vision-language models (VLMs) \cite{Achiam2023GPT4TR, Liu2023VisualIT, Bai2025Qwen25VLTR} offer a scalable alternative, showing strong performance in video understanding tasks. In this section, we leverage these models to perform automatic evaluation of professional video generation.

\textbf{Zero-shot VLM Evaluations.} \label{sec:zeroshot_vlm} 
The rise of multimodal VLMs has enabled progress on vision-language tasks, including video understanding, making them natural candidates for evaluating professional video generation.  We use expert annotations as ground truth and measure VLM alignment by prompting models with a video, its associated prompt, and a specific question tied to a taxonomy node, asking for a 1–5 rating similar to our user study. We explicitly instruct the VLM to ignore factors unrelated to the specified category when evaluating the video. 
We determine VLM preferences by independently scoring each video and selecting the higher-scoring one. This design mitigates hallucination and order-sensitivity issues commonly observed when prompting with both videos simultaneously. 
\begin{wrapfigure}[14]{r}{0.48\textwidth} 
  \vspace{-0.7\baselineskip}              
  \centering
  \includegraphics[width=\linewidth]{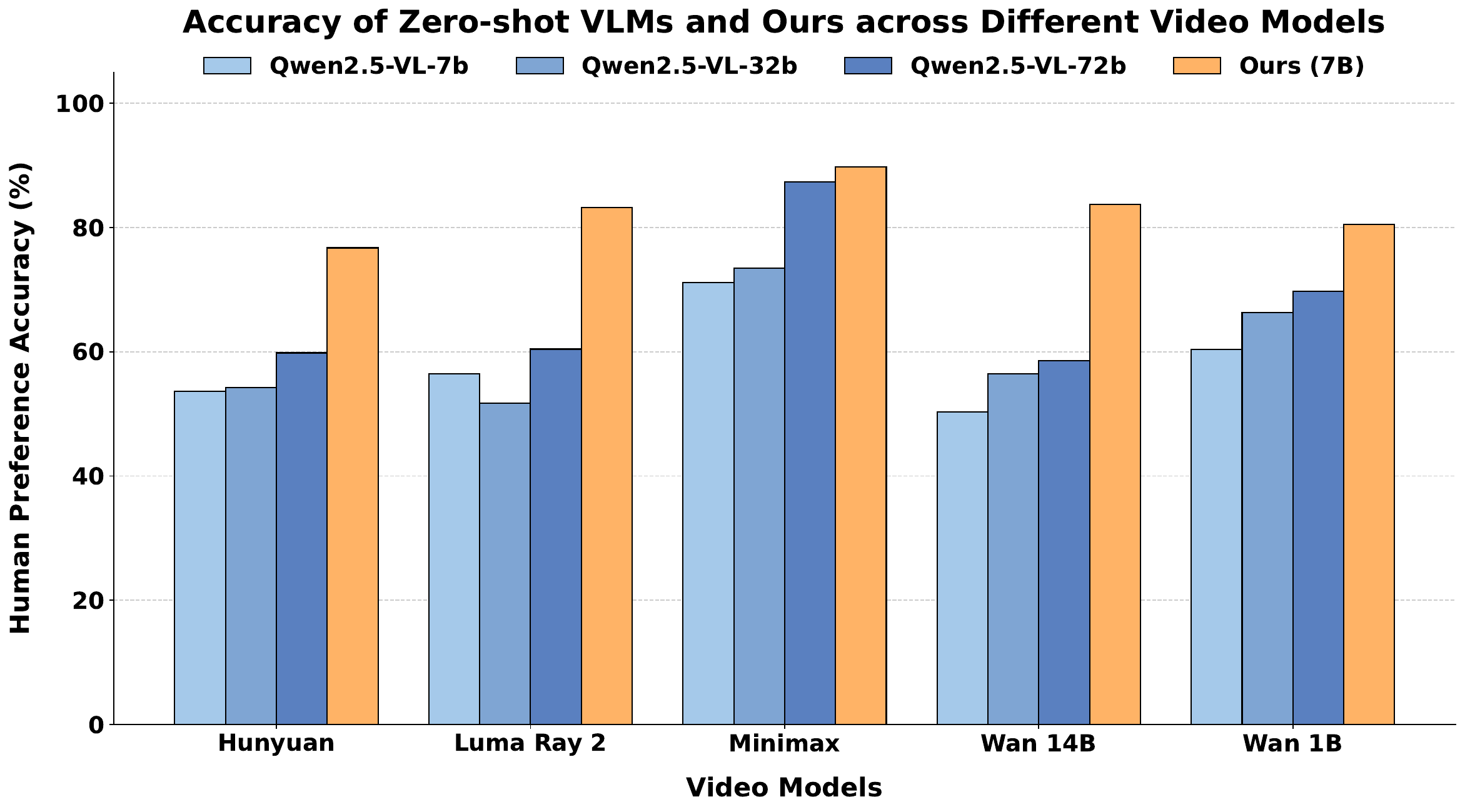}
  \captionsetup{width=0.95\linewidth}     
  \caption{Our trained VLM shows consistent alignment with human annotations across video generation models, outperforming baselines, most notably on WAN-14B.}
  \label{fig:vlm_results}
  \vspace{-0.5\baselineskip}              
\end{wrapfigure}
We use Qwen2.5-VL-Instruct models due to their strong video understanding capabilities. To study the effect of model scale, we evaluate 3 sizes: 7, 32 and 72B. Figure~\ref{fig:vlm_results} shows that increasing model size does not yield significant improvements in alignment with human judgment. Our results reveal overall poor agreement, consistent with prior findings \cite{he2024videoscore, xu2024visionreward} that highlight the need for fine-tuning on in-distribution, human-labeled data.

\textbf{Aligning Human and VLM ratings.} We adopt Qwen-2.5-VL-7B \cite{qwen2.5-VL} as the base model for fine-tuning; our training and validation dataset consist of 44,062 and  12,763 samples, respectively. We aggregate annotator scores (3/video pair) into binary preferences, excluding ties. Videos are preprocessed at 2 fps at their native resolution. Each sample consists of a prompt, two videos, and a binary label. The trained model acts as a classifier; we modify the model architecture to output a scalar score using a linear projection over the final layer’s last token. The model takes as input a single video, its prompt, and the evaluation question as input. For training, we use the Bradley–Terry objective \cite{bradley1952rank} due to its sample efficiency over regression \cite{liu2025improving}. The model is trained for 1 epoch with a batch size of 8 and learning rate 6e-5. Similar to zero-shot evaluation, we compute pairwise preference accuracy against the average of the annotators as the target metric. Our fine-tuned model achieves an overall accuracy of 72.36\%, outperforming all zero-shot VLM baselines. This represents an absolute improvement of $\sim$20\%, over the baseline 7B model. Our model (Figure~\ref{fig:vlm_results}) shows consistent performance across videos generated by different models, highlighting its ability to generalize across different video qualities. Additional VLM results are presented in Appendix \ref{sec:vlm_additional}.

\section{Conclusion and Future Work}

Stable Cinemetrics probes at the intersection of professional video generation and generative video models, grounding prompts, evaluations, and analysis in our structured taxonomies. Our findings reveal where current state-of-the-art models perform well, and where substantial improvements are needed. Our prompt suite offers a strong testbed for future video generative models and can be easily extended as models improve, owing to the flexibility of our taxonomy. To support future evaluations, we introduce a trained VLM aligned with expert annotations, enabling scalable annotation as new models are released. We envision several extensions of our work; while our current focus is on evaluation, the taxonomy can also support analyzing video datasets for cinematic diversity or serve as a structure for video captioning. While today’s text-to-video models are not yet usable in a fully zero-shot capacity, our findings identify the main challenging pillars for professional filmmaking, illuminating the need for potential solutions like fine-tuning and customization that can bring these models closer to real production use. We hope SCINE encourages deeper exploration at the intersection of filmmaking and video generative models, fostering closer collaboration between artists and models.
\section{Acknowledgement}

We thank Robert Legato, Hanno Basse and Heather Ferreira for their valuable input on our work. We are also grateful to the team at MovieLabs for their feedback on our taxonomies. A special thanks to Cedric Wagrez for his invaluable assistance with the human annotations!
\section{Limitations}

Although our taxonomy was developed in consultation with domain experts, it is limited by the scope of our collaborator network. Filmmaking terminology and interpretive nuance vary across regions and cultures, greater expert diversity would enable broader incorporation of global cinematic controls into the taxonomy. Some taxonomy nodes (e.g., Color Temperature, ISO) were abstracted for evaluation, as we found it difficult for annotators to consistently perceive fine-grained values (such as 2000K or ISO 800). Prompt generation is based on LLMs, whose proprietary nature and potential biases can influence the language and structure of the prompts. Our zero-shot VLM evaluations were bounded by compute and data resources, limiting the scale and scope of the experiments.

\bibliography{neurips_2025_bib}

\newpage
\appendix

\section{Appendix}

\newcommand{\apptoc}[2]{
  \item \hyperref[#1]{#2}\dotfill\pageref{#1}
}
\begin{list}{}{\leftmargin 0pt \itemindent 0pt}
  \apptoc{sec:taxonomy}{Taxonomy Details}
  \apptoc{sec:additional_results}{Additional Results and Analysis}
  \apptoc{sec:prompts_gen}{Details on Prompt Generation}
  \apptoc{sec:distribution_prompts}{Distribution of Taxonomy Categories in SCINE Prompts}
  \apptoc{sec:annotation_details}{Annotation Details}
  \apptoc{sec:statistical_tests}{Statistical Tests}
  \apptoc{sec:vlm_additional}{Additional VLM Results}
  \apptoc{sec:new_model_results}{Additional Results on Recent Models}
\end{list}

\subsection{Taxonomy Details}\label{sec:taxonomy}

We provide additional details on control nodes and their associated values in the taxonomies. Some nodes accept open-ended values, for example, a range of stand-alone actions. To simplify evaluation, we abstract certain values that can be fine-grained in future works. For instance, we group \texttt{Aperture} into wide/medium/narrow, though exact f-stop values can also be studied in the future. Similarly, \texttt{color palette} is treated as a discrete value in our current work, but can be decomposed into hue, brightness, and saturation. Table \ref{tab:camera-features} - \ref{tab:event-features} details the control nodes and their values of the Camera, Lighting, Setup and Events Taxonomies, respectively.

\begin{table}[h]
\centering
\caption{\textbf{Camera Taxonomy} Control Nodes and Values}
\label{tab:camera-features}
\resizebox{\linewidth}{!}{
\begin{tabular}{|c|p{8cm}|p{6cm}|}
\hline
\textbf{Name} & \textbf{Description} & \textbf{Potential Values} \\
\hline
Lens Size & Defines the focal length and field of view of the camera lens. & Standard, Fisheye, Wide, Medium, Long Lens, Telephoto \\
\hline
Depth of Field & Controls the range of focus in the image, affecting subject isolation. & Deep, Shallow, Soft, Rack, Split Diopter, Tilt Shift \\
\hline
Aperture & The camera lens opening that controls the amount of light propagated through the camera. & Wide, Medium, Narrow \\
\hline
Shutter Speed & The duration for which the camera sensor is exposed to light. & Slow, Medium, Fast \\
\hline
ISO & Sensitivity of the camera sensor to light. & Low, Medium, High \\
\hline
Angle & Defines the camera’s viewpoint in relation to the subject. & Low, High, Aerial, Overhead, Dutch, Eye-level, Shoulder, Hip, Knee, Ground, Continuous Values \\
\hline
Static & A fixed camera position without any movement. & - \\
\hline
2D & Camera movements restricted to horizontal or vertical axes. & Pan left, Pan right, Tilt up, Tilt down, Zoom in, Zoom out \\
\hline
3D & Camera movements that incorporate spatial depth and multi-axis motion. & Push In, Pull Out, Dolly Zoom, Camera Roll, Tracking, Trucking, Arc, Crane \\
\hline
Gear & Specifies the support systems and stabilization equipment used to facilitate camera movement. & Handheld, Tripod, Pedestal, Cranes, Overhead Rigs, Dolly, Stabilizer, Snorricam, Vehicle Mount, Drones, Motion Control, Steadicam \\
\hline
Shot Size & Determines how much of the subject and surroundings are visible in the frame. & Establishing, Master, Wide, Full, Medium–Full, Medium, Medium–Close-up, Close-up, Extreme Close-up \\
\hline
Framing & Placements and composition of subjects within the frame. & Single, Two Shot, Crowd, OTS, PoV, Insert \\
\hline
\end{tabular}
}
\end{table}

\begin{table}[h]
  \centering
  \caption{\textbf{Lighting Taxonomy} Control Nodes and Values}
  \label{tab:lighting-features}
  \resizebox{\linewidth}{!}{
  \begin{tabular}{|l|p{8cm}|p{6cm}|}
    \hline
    \textbf{Name} & \textbf{Description} & \textbf{Potential Values} \\
    \hline
    Natural Light              & Natural sources of light, such as sunlight, moonlight, or firelight. & Sunlight, Moonlight, Firelight \\
    \hline
    Artificial/Practicals Light & Man-made light sources that illuminate the scene. & LED, HMI, Tungsten, Fluorescent, HID \\
    \hline
    Color Temperature          & Defines the hue of the light, typically measured in Kelvin, affecting the scene’s mood. & Warm, Cool, Cold \\
    \hline
    Lighting Conditions        & Describes various lighting scenarios or ambient conditions present in a scene. & Candlelight, Golden Hour, White Fluorescent, Clear Daylight, Overcast \\
    \hline
    Soft Shadows               & Subtle and diffused shadows resulting from indirect or scattered light. & Diffused Light, High Key Lighting, Reflectors \\
    \hline
    Hard Shadows               & Sharp, well-defined shadows generated by a direct light source. & Direct Light, Low Key Lighting \\
    \hline
    Reflection                 & The effect of light bouncing off surfaces to create a reflective appearance. & - \\
    \hline
    Lighting Position          & Specifies the placement or direction of the light source relative to the subject. & Back Light, Fill Light, Top Light, Side Light, Key Light \\
    \hline
    Motion                     & Dynamic changes or movement in the lighting effect. & Flickering, Pulsing \\
    \hline
    Color Gels                 & Colored filters applied to lights to modify or enhance the color of the illumination. & -  \\
    \hline
  \end{tabular}
 }
\end{table}

\begin{table}[!h]
  \centering
  \caption{\textbf{Setup Taxonomy} Control Nodes and Values}
  \label{tab:setup-features}
  \resizebox{\linewidth}{!}{
  \begin{tabular}{|l|p{8cm}|p{6cm}|}
    \hline
    \textbf{Name} & \textbf{Description} & \textbf{Potential Values} \\
    \hline
    Contrast & Determines the difference between light and dark areas to enhance visual impact. & Low, High \\
    \hline
    Blur & Introduces softness to parts of the image to guide focus or create mood. & Gaussian, Radial, Motion \\
    \hline
    Noise & Adds random variations in brightness or color, mimicking film grain or digital sensor noise. & Gaussian, Salt and Pepper, Poisson \\
    \hline
    Film Grain & Emulates the granular texture of traditional film photography for a classic look. & - \\
    \hline
    Color Palette & Defines the overall range and harmony of colors in the scene, influencing its mood. & \textit{Open Set} \\
    \hline
    Lines & Directional elements that guide the viewer’s gaze within the shot. & Horizontal, Vertical, Diagonal \\
    \hline
    Regular Shapes & Structured, geometric forms such as squares, circles, and triangles that add order to the design. & Square, Circle, Triangle \\
    \hline
    Natural Shapes  & Unstructured shapes that naturally emerge in the scene, without any geometric constraints. & Water-like, Cloud-like \\
    \hline
    Frame Balance & Refers to the distribution of visual weight across the composition, ensuring a harmonious layout. & Rule of Thirds, Symmetry, Right Heavy, Left Heavy \\
    \hline
    Positional Accuracy & The absolute position of an object or a subject in a scene. & \textit{Open Set} \\
    \hline
    Relative Positioning & The relative positioning of an object in relationship to other objects in the scene. & \textit{Open Set} \\
    \hline
    Depth & Controls the perception of distance between elements, enhancing the three-dimensional feel of the scene. & Deep, Flat, Limited, Ambiguous \\
    \hline
    Setting & Defines if the scene is happening indoors or outdoors. & INT/EXT \\
    \hline
    Time of Day & The time of day the scene is set in. & DAY, NIGHT, MORNING, EVENING, DAWN, DUSK, LATE NIGHT, MIDDAY, SUNRISE, SUNSET, AFTERNOON \\
    \hline
    Location & The specific place or setting of the scene. & \textit{Open Set} \\
    \hline
    Negative Space & Defines if there a lot of empty space. & - \\
    \hline
    Positive & Defines how the space is occupied in the environment. & Clean, Cluttered \\
    \hline
    Mood & The emotional atmosphere or feeling created by the environment. & \textit{Open Set} \\
    \hline
    Scale & The relative size or extent of the environment. & \textit{Open Set} \\
    \hline
    Style & The artistic or visual style of the backdrop. & \textit{Open Set} \\
    \hline
    Background & The part of the scene that is behind the main subject and does not need to be exactly described. & \textit{Open Set} \\
    \hline
    Elements & The natural or artificial components of the backdrop. & Rain, Snow, Fog, Wind, Thunder, Smoke, Dust, Ash, Fire \\
    \hline
    Prop Description & A general description of the prop. & \textit{Open Set} \\
    \hline
    Prop Class & The category or type of the prop. & \textit{Open Set} \\
    \hline
    Prop Material & The substance(s) the prop is made of. & Wood, Glass, Gold, Paper, Plastic \\
    \hline
    Prop Pattern & The design on the prop. & Grid, Checker, Stripes, Zigzag, Dots, Bricks, Metal, Hexagons \\
    \hline
    Prop Utility & The purpose or function of the prop, whether it just exists in the scene or will it be used by the subject. & Decorative, Functional \\
    \hline
    Subject Class & The category or type of the subjects. & \textit{Open Set} \\
    \hline
    Subject Accessories & Items worn or carried by the subjects that enhance their appearance or functionality. & \textit{Open Set} \\
    \hline
    Subject Costume & The clothing worn by the subjects, especially for a performance or to create a specific character. & \textit{Open Set} \\
    \hline
    Subject Hair & The style and appearance of the subjects' hair. & \textit{Open Set} \\
    \hline
    Subject Makeup & Cosmetics applied to the subjects' face or body to enhance or alter their appearance. & \textit{Open Set} \\
    \hline
    Subject Pose & The position or stance of the subjects, especially for a photograph or portrait. & \textit{Open Set} \\
    \hline
    Subject Silhouette & The outline or shape of the subjects against a light background. & \textit{Open Set} \\
    \hline
    Subject Proportions & The relative size and scale of the subjects' body parts or features. & \textit{Open Set} \\
    \hline
    Text Generation & The process of creating written content to be displayed on the video. & \textit{Open Set} \\
    \hline
  \end{tabular}
  }
\end{table}

\begin{table}[h]
  \centering
  \caption{\textbf{Events Taxonomy} Control Nodes and Values}
  \label{tab:event-features}
  \resizebox{\linewidth}{!}{
  \begin{tabular}{|l|p{8cm}|p{6cm}|}
    \hline
    \textbf{Name} & \textbf{Description} & \textbf{Potential Values} \\
    \hline
    Standalone Actions & If the action is stand-alone & \textit{Open Set} \\
    \hline
    Interactive Actions & If the action involves subject-subject or object-subject interaction & \textit{Open Set} \\
    \hline
    Temporal (Actions) & How actions unfold across time. & Atomic, Concurrent, Sequential, Causal, Overlapping, Cyclic, Reverse \\
    \hline
    Foreground (Actions) & Describes if the action is taking place in the foreground & Local, Global, Focal\\
    \hline
    Background (Actions) & Describes if the is taking place in the background & -  \\
    \hline
    Uncertainty & The probabilistic nature of the action outcome & Probabilistic, Deterministic, Mixed \\
    \hline
    Implicit Emotions & Emotions that are suggested or implied rather than directly stated. & \textit{Open Set} \\
    \hline
    Explicit Emotions & Emotions that are clearly and directly shown or stated within the scene. & \textit{Open Set} \\
    \hline
    Temporal (Emotions) & How emotions evolve across time. & Atomic, Concurrent, Sequential, Overlapping, Causal \\
    \hline
    Foreground (Emotions) & Describes if the emotion is taking place in the foreground & Local, Global, Focal \\
    \hline
    Background (Emotions) & Describes if the emotion is taking place in the background & - \\
    \hline
    Type of Dialogue Delivery & How the dialogue is delivered & Dash, Ellipsis, Monologue \\
    \hline
    Foreground (Emotions) & Describes if the dialogue is being spoken in the foreground &  Local, Global, Focal \\
    \hline
    Background (Emotions) & Describes if the the dialogue is being spoken  the background & - \\
    \hline
    Change in Environment & Change of environment or occurrences within a shot & \textit{Open Set} \\
    \hline
    Story Structure & Key narrative elements that shape the scene’s progression. & Turning Point, Climax, Foreshadowing, Conflict \\
    \hline
    Pace & How fast the events are happening in a shot & Slow, Fast \\
    \hline
    Regularity & How regularly the events are happening in a shot & Regular, Irregular \\
    \hline
  \end{tabular}
  }
\end{table}

\newpage

\subsection{Additional Results and Analysis}\label{sec:additional_results}

\subsubsection{Events} \label{sec:events_appendix_results}

\begin{figure}[]
  \centering
  \includegraphics[width=\linewidth]{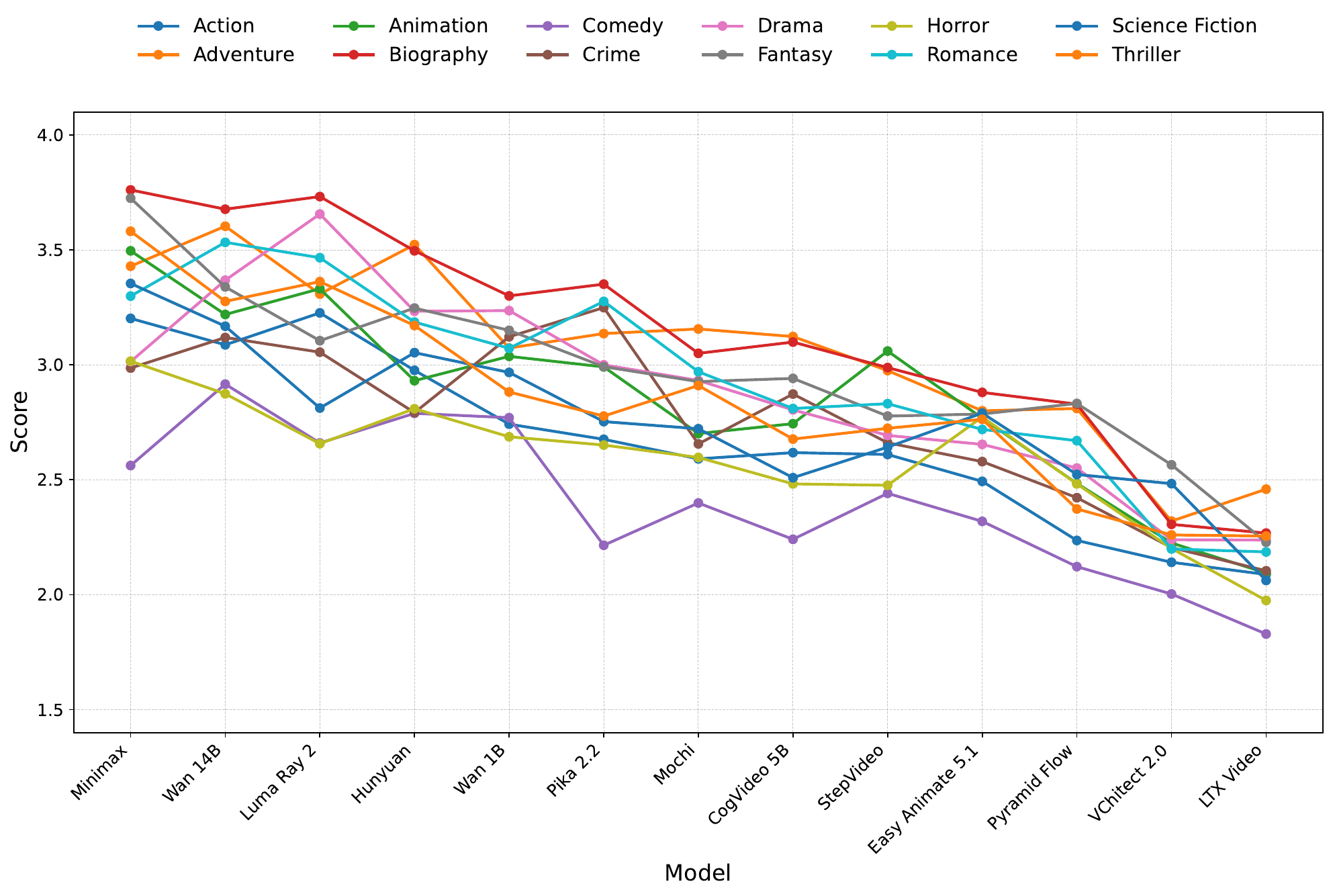}
  \caption {\textbf{Model performance on Events across genres}. Across 13 models and 12 genres, portrayal of Events in Biography and Adventure are the strongest, while Comedy and Horror are the weakest.}
  \label{fig:models_genres}
\end{figure}

Figure \ref{fig:models_genres} shows Events performance across 13 models and 12 genres. Biography and Adventure are strongest whereas Comedy and Horror are the weakest. Minimax leads in 6/12 genres, Luma Ray 2 tops Action and Drama, and WAN-14B is the most consistent, with the lowest standard deviation. 

\begin{figure}[]
  \centering
  \includegraphics[width=\linewidth]{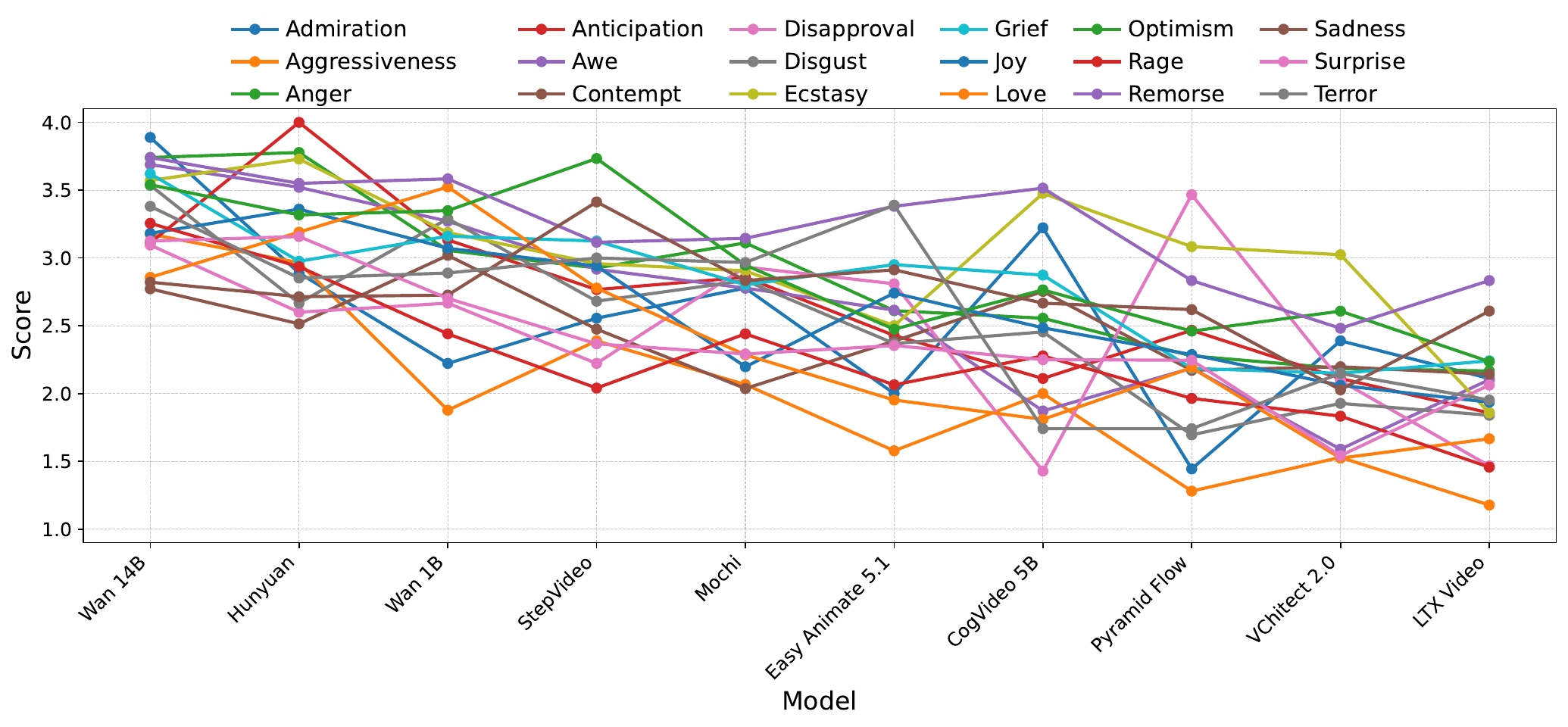}
  \caption {\textbf{Model performance on Emotions}. Among 10 models and 19 emotions, Remorse is best portrayed, while Aggressiveness is the weakest.}
  \label{fig:models_emotions}
\end{figure}

\begin{figure}[!htbp]
  \centering
  \includegraphics[width=0.85\linewidth]{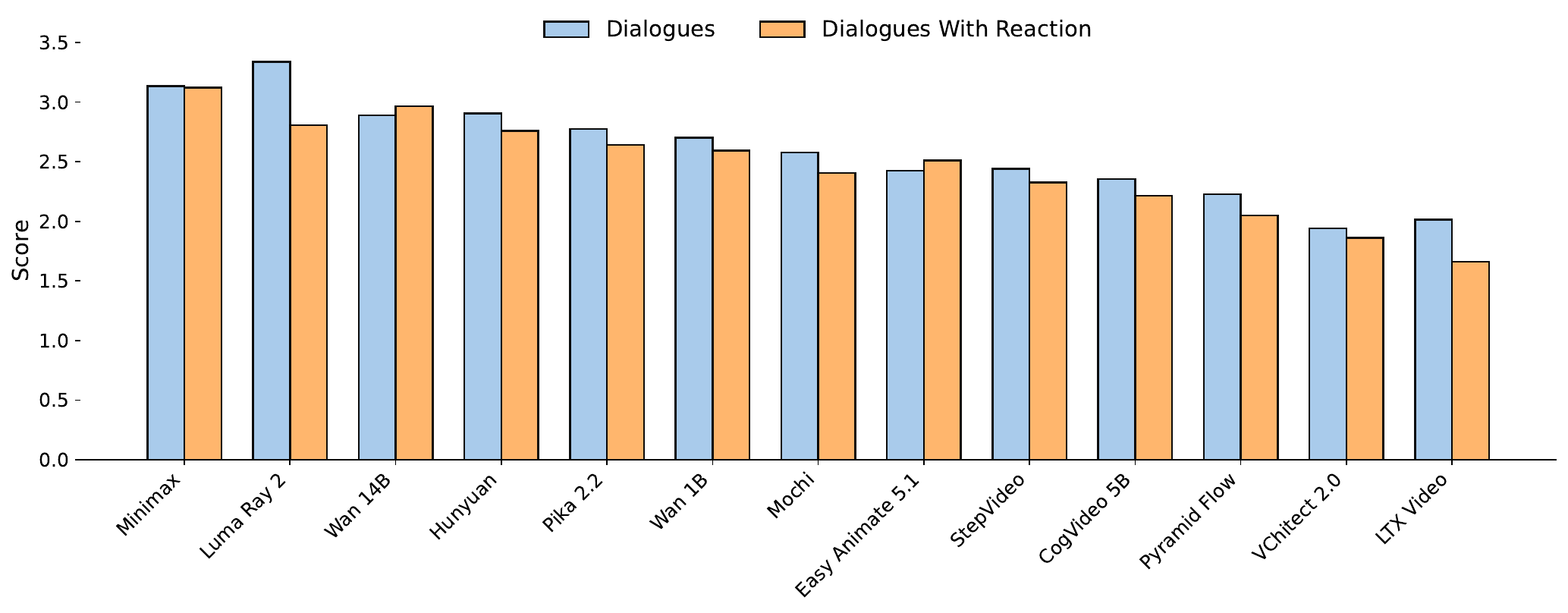}
  \caption {\textbf{Model performance on Dialogues}. Compared to Actions and Emotions, models struggle at Dialogues. Within Dialogues, performance drop is seen during multi-turn conversations.}
  \label{fig:models_dialogues}
\end{figure}

Figure \ref{fig:models_emotions} shows performance of 10 open-source models across 19 emotion classes. Models perform best on remorse and ecstasy, but fare poorly on aggressiveness and rage. As shown in Figure \ref{fig:models_dialogues}, dialogue performance is weaker in comparison to Emotions and Actions. Models particularly struggle with multi-turn dialogues or when non-verbal reactions follow. Since T2V models do not generate audio, we evaluate whether the correct character delivers the line and/or with appropriate visual expression. Most models fail to localize the speaker, often attributing a single dialogue to multiple characters.

\subsubsection{Setup}

\begin{figure}[!htbp]
  \centering
  \includegraphics[width=0.85\linewidth]{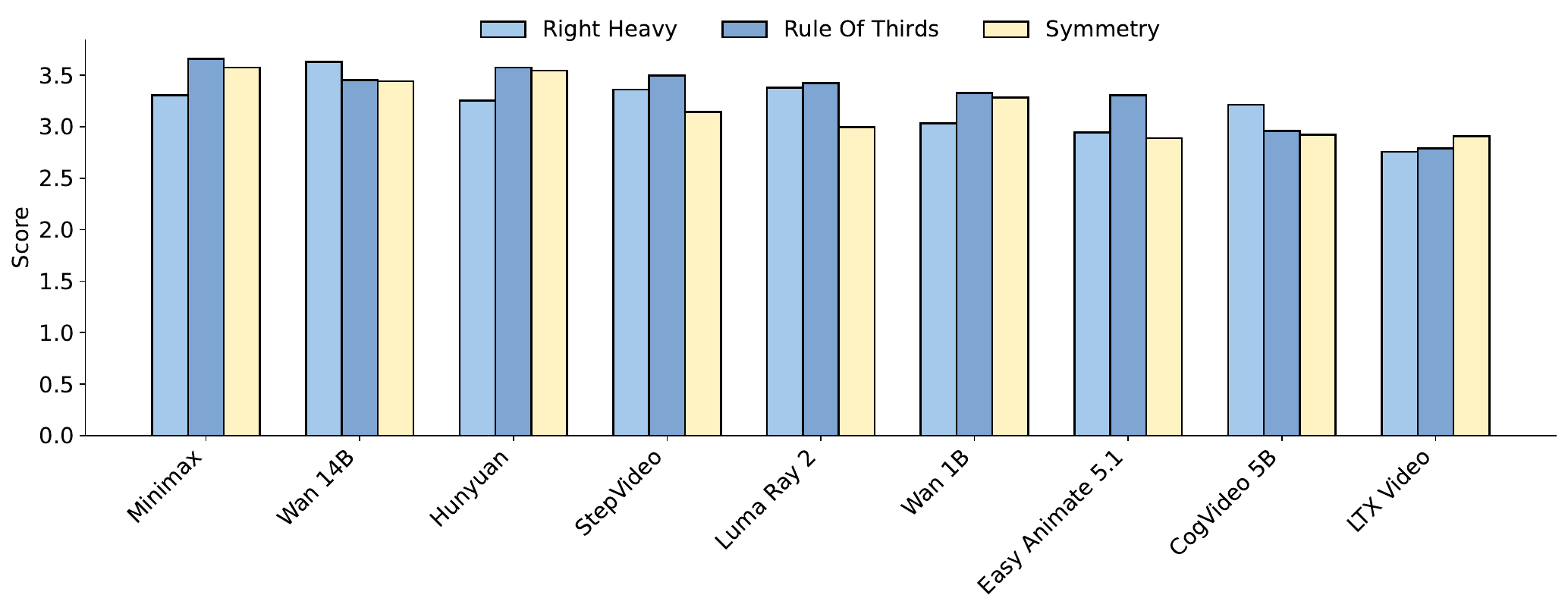}
  \caption {Between different frame compositions, models are better at Rule of Thirds but struggle at maintaining Symmetry.}
  \label{fig:setup_balance_values}
\end{figure}

\begin{figure}[!htbp]
  \centering
  \includegraphics[width=\linewidth]{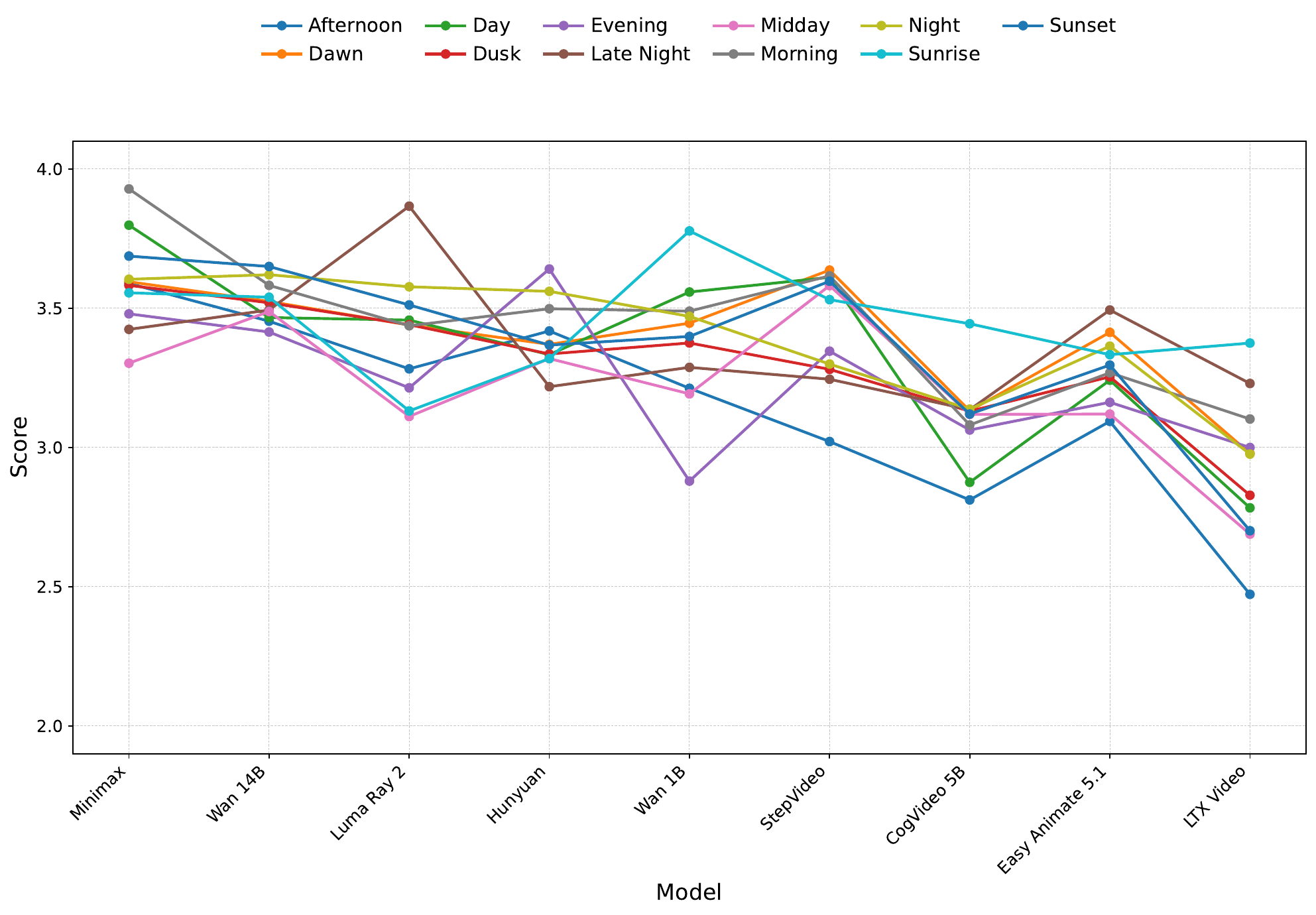}
  \caption {Across Time of Day setups, Sunrise shots are handled better, while Afternoon remains more challenging for models.}
  \label{fig:setup_tod_values}
\end{figure}

For the Setup taxonomy, we also analyze performance at the value level. In Balance (Figure \ref{fig:setup_balance_values}), models handle rule of thirds framing more effectively but struggle with symmetrical compositions. For Time of Day (Figure \ref{fig:setup_tod_values}), among 11 categories, Sunrise and Morning are portrayed well, while Afternoon remains challenging.

\subsection{Details on Prompt Generation}\label{sec:prompts_gen}
Table \ref{tab:cat_q_example} presents an example of a prompt and the corresponding categories and generated questions.

\begin{table}[]
  \centering
  \caption{A working example of a prompt with its corresponding categories and questions. Each question targets a single control node from the taxonomy, enabling human annotators to perform fine-grained, independent evaluations per node.}
  \label{tab:cat_q_example}
  \resizebox{\linewidth}{!}{
  \begin{tabular}{%
      >{\raggedright\arraybackslash}p{6cm}  
      >{\raggedright\arraybackslash}p{5cm}  
      >{\raggedright\arraybackslash}p{6cm}  
    }
    \toprule
    \textbf{Prompt} & \textbf{Final Category} & \textbf{Question} \\
    \midrule
    \multirow{6}{=}{In a stark white laboratory illuminated by cool LEDs casting clinical precision, a scientist carefully drops a single blue chemical into a beaker, the camera framing an intimate close-up as soft depth of field blurs the sterile environment behind. A back light carves a subtle halo around the glassware moments before the liquid erupts into bright green, intensified by a strategic neon-tinted color gel that makes the reaction glow like bottled lightning.}
      & Camera → Creative Intent → Shot Size
      & Does the generated video clearly exhibit a well-executed close-up shot that captures the subject with the intended intimacy and detail? \\
    & Camera → Intrinsics → Depth of Field
      & Does the video effectively showcase a soft depth of field that isolates the subject while smoothly blurring the background? \\
    & Lighting → Sources → Artificial/Practicals Light
      & Is the effect of artificial LED source clearly visible and does it emulate the clinical, cool lighting effect as described in the scene? \\
    & Lighting → Color Temperature
      & Does the video convey a cool color temperature in its lighting setup that reinforces the clinical precision suggested in the prompt? \\
    & Lighting → Lighting Position
      & Is a back lighting effect evident in the video, such that it effectively carves a halo or outline around the subject as described? \\
    & Lighting → Advanced Controls → Color Gels
      & Does the video incorporate a neon-tinted color gel effect that intensifies the lighting during the chemical reaction as detailed in the prompt? \\
    \bottomrule
  \end{tabular}
  }
\end{table}

\tcbset{
  promptbox/.style={
    breakable,                     
    enhanced,                       
    colback=gray!10,                
    colframe=blue!60!black,         
    arc=3pt,                        
    boxrule=0.6pt,                  
    left=4pt, right=4pt, top=4pt, bottom=4pt, 
    fontupper=\ttfamily\small,     
    attach boxed title to top left={yshift=-2pt, xshift=5pt},
    boxed title style={
      colback=blue!60!black,
      colframe=blue!60!black,
      arc=2pt,
      boxrule=0.6pt,
      left=3pt, right=3pt, top=1pt, bottom=1pt,
      fonttitle=\color{white}\ttfamily\footnotesize
    },
    title=Prompt,                   
    before skip=\medskipamount,
    after skip=\medskipamount,
  }
}
\newtcolorbox{promptbox}[1][]{
  promptbox,
  title=Prompt,
  #1                              
}

Below, we show an example of the instruction given to an LLM to upsample a SCINE-Script into SCINE-Visuals by incorporating control nodes from the Camera and Lighting taxonomies.
\begin{promptbox}[title=SCINE Scripts - Cinematographer]

\textbf{System Prompt} : \\
You are a world-class cinematographer known for your visionary storytelling, mastery of light, and camera. You have decades of experience working on award-winning films across genres, collaborating with top directors and production teams. 
Your insights blend technical expertise with artistic sensibility. When describing scenes or advising on visual storytelling, you use cinematic terminology with clarity and inspiration. 
Think like Roger Deakins, Emmanuel Lubezki, and Greig Fraser—your visual choices always elevate the emotional tone and narrative arc of a project.

\textbf{User Prompt} : \\\\
\textbf{GOAL}\\
You will be given a prompt and 2 taxonomies that define camera and lighting controls commonly used by cinematic professionals. Your objective is to enrich the given prompt by sampling relevant nodes from both the taxonomies.
As a cinematographer, your role is to "shoot" this scene using the best possible cinematic expression, utilizing the camera and lighting control options provided in the taxonomy.
\\\\
\textbf{PROMPT}: \{prompt\}  
\\\\\
\textbf{MOST IMPORTANT INFORMATION}\\
1. Only Use Nodes from the Provided Taxonomies :
    - You must never introduce nodes that are not present in the given taxonomies. 
    - While the values within each node can be flexible—allowing for creativity and imagination grounded in your professional experience. For example, the node "Color Gel" is defined, but has no values. It is upto you to define these values.
    - The structure must strictly adhere to the nodes defined in the taxonomy. Think expansively within the bounds of each node, but never go beyond them.\\
2. Preserve the Original Prompt Content : 
    - Do NOT remove or add any of the original content from the input prompt. 
    - Your only task is to enrich the prompt by layering in camera and lighting related information. The core semantics and narrative of the prompt must remain entirely intact.\\
3. Do NOT include the path through which you sample the nodes in the prompt. That is, do NOT add the paths from the taxonomy using '->'.
\\\\
\textbf{GUIDELINES}
1. Input Prompt – The input prompt describes a single continuous event, intended to occur within one uninterrupted shot. Therefore, do not include any cuts or multiple camera setups. Assume this is a one-shot sequence. \\ 
2. Each node in the taxonomies contains:
    - Description: A definition of what the node represents.\\
    - Example: An example of how the node may appear in a prompt. \\
    - Values: A non-exhaustive list of possible values for the node. Some notation:
        a. OPEN SET – Indicates the node supports a wide range of possible values. \\
        b. [] – Indicates the node may have multiple values, which are not predefined and should be selected based on your reasoning and cinematic knowledge. \\
3. Enriched Prompt – Your enriched version will serve as input to a text-to-video model. It must be fluent, natural, and interpretable by the model, while incorporating cinematic elements effectively.
\\\\
\textbf{ CAMERA TAXONOMY}
The Camera Taxonomy defines elements related to the camera’s intrinsics, extrinsics, and its cinematic use : 
\{camera\_taxonomy\}
\\\\
\textbf{ LIGHTING TAXONOMY}
The Lighting taxonomy broadly defines all elements of lighting, including source, position of lighting, along with its effects such as shadows and reflections, along with color temperature, lighting motion such as flickering etc : 
\{lighting\_taxonomy\}

When incorporating lighting into your enriched prompt, remember that a cinematographer can shape the look and feel of a shot by selectively illuminating different depth planes of the scene. 
Lighting can be applied to the foreground, mid-ground, background, and the subject itself—either individually or in combination. Your choices should support the emotional tone, visual focus, and narrative intent of the shot.
\end{promptbox}

Below, we show an example of the instruction given to an LLM to categorize and generate evaluation questions for an input prompt using the Camera taxonomy.

\begin{promptbox}[title=Camera Categorization and Question Generation]
\textbf{GOAL} \\ 
You are an expert prompt evaluator. Your task is to analyze a video generation prompt and categorize it based on a predefined taxonomy.
\\\\
\textbf{PROMPT}: \{prompt\}  
\\
Available Categories (with Examples)\\
The category presented to you is that of Camera. The Camera taxonomy broadly defines everything related to the camera - the intrinsics, the extrinsics and the cinematic use of camera.
\{camera\_taxonomy\}  
\\\\
\textbf{Notes about the Taxonomy} \\

Each node in the taxonomy contains : \\
1. Description : Definition of what that node represents. \\
2. Example : An example of the presence of a node in the form of a prompt. \\
3. Values : A non-exhaustive list of values of these nodes. Values are a list of values that this node can have. Some nomenclature : \\
    a. OPEN SET indicates that this node contains a large number of values. \\
    b. [] indicates that this node may have multiple values, but are not defined explicitly and it is upto your reasoning and knowledge.\\\\

\textbf{Examples of Categorization} \\

1. Static Medium-Close-Up of David's face showing quiet devastation. Quick Push In as tears well up in his eyes. Shot with a medium ISO to capture the dim apartment lighting. \\

Static - Camera -> Trajectory -> Camera Movement -> Static\\
Push In - Camera -> Trajectory -> Camera Movement -> 3D\\
Medium-Close up - Camera -> Creative Intent -> Shot Size\\
Medium ISO - Camera -> Intrinsics -> Exposure -> ISO\\\\

2. Wide shot of a bustling city street at night. The neon lights of the shops and restaurants cast a colorful glow on the wet pavement. People walk by, their faces illuminated by the bright signs. The camera pans up to reveal the towering skyscrapers that loom overhead, their windows reflecting the city lights.\\

Wide shot - Camera -> Creative Intent -> Shot Size\\
Pans Up - Camera -> Trajectory -> Camera Movement -> 2D
\\\\

\textbf{TASK} \\
Analyze the given prompt and return the following structured output in a valid JSON format: \\

Words: Extract important keywords or key phrases from the prompt using the following guidance:\\
    - Identify named entities related to a camera in professional use as you would in NER (Named Entity Recognition).\\
    - Extract noun phrases or descriptive terms that relate to a camera.\\
    - Prefer multi-word expressions where meaningful related to a camera.\\
    - Avoid generic or uninformative words like “a”, “video”, “the”, etc.\\

Categories: For each word or phrase, assign the most appropriate category from the taxonomy. A dictionary of relevant categories from the taxonomy.\\
    - For each relevant category, assign a score between `0` and `1` representing how strongly the prompt matches the category.\\
    - Provide a reason for each score, referring to the words or phrases extracted and how they relate to the category.\\
    - Generate a question that helps a human evaluator determine whether this category is visually present in the generated video. Use your reasoning to guide the question. The evaluator will use this question to rate the video on a scale from 1 (not at all) to 5 (strongly represented). \\
    - The generated question should evaluate quality, consistency and presence of the node in the video.\\    

\textbf{Important Guidelines}: \\
    - The camera information should be explicitly mentioned in the prompt. Do NOT imply, assume or derive anything. Only consider a word or a phrase a match, if it is explicitly mentioned in the prompt.\\
    - Each prompt can have multiple nodes of the Camera taxonomy. You should capture all of the nodes in the prompt and map it back to the taxonomy.\\
    - You must always traverse from the root node, which is Camera in this case. That is, the 'category' should always start as (Camera -> ..)\\
    - You will never create a node that is not in the taxonomy. These nodes can have multiple values, as previously explained and you are expected to be imaginative about the values. But the nodes, should always come from the given taxonomy.\\
    - Since the taxonomy is of Camera, we do not care about objects, subjects, lighting, events, actions or emotions. Your sole focus should be about camera terms that are present in the prompt in accordance with the taxonomy. You will NOT ask any question related to objects, subjects, lighting, events, actions or emotions.\\
\end{promptbox}


\begin{table}[]
  \centering
  \caption{Lexical Diversity of SCINE Scripts. Compared to existing prompt-based benchmarks, SCINE-Scripts demonstrate higher lexical diversity across multiple metrics.}
  \label{tab:lexical_diversity}
  \resizebox{\linewidth}{!}{
  \begin{tabular}{lccc}
    \toprule
    \textbf{Benchmark} & TTR $\uparrow$ & Distinct Bi-Grams $\uparrow$ & Jaccard Distance $\uparrow$ \\
    \midrule
    VBench  \citepapp{Huang_2024_CVPR}      & 0.1489 & 0.4605 & 0.9384 \\
    MovieGenBench \citepapp{polyak2025moviegencastmedia}   & 0.1660 & 0.5311 & 0.9285 \\
    EvalCrafter   \citepapp{Liu_2024_CVPR}    & \textbf{0.2270} & \underline{0.6038} & \underline{0.9413} \\
    T2V-CompBench  \citepapp{sun2025t2vcompbenchcomprehensivebenchmarkcompositional}  & 0.1435 & 0.4781 & 0.9350 \\
    SCINE Scripts   & \underline{0.1760} & \textbf{0.6177} &\textbf{ 0.9445} \\
    \bottomrule
  \end{tabular}
  }
\end{table}

Table \ref{tab:lexical_diversity} compares SCINE-Scripts with existing prompt-based video generation benchmarks. We compute token level metrics: Type-Token Ratio (TTR), Distinct Bi-Grams, and average pairwise Jaccard Distance, and find that SCINE-Scripts exhibits strong lexical diversity.

\subsection{Distribution of Taxonomy Categories in SCINE Prompts} \label{sec:distribution_prompts} 
\begin{figure}[]
  \centering
   \begin{subfigure}[b]{0.48\textwidth}
    \includegraphics[width=\textwidth]{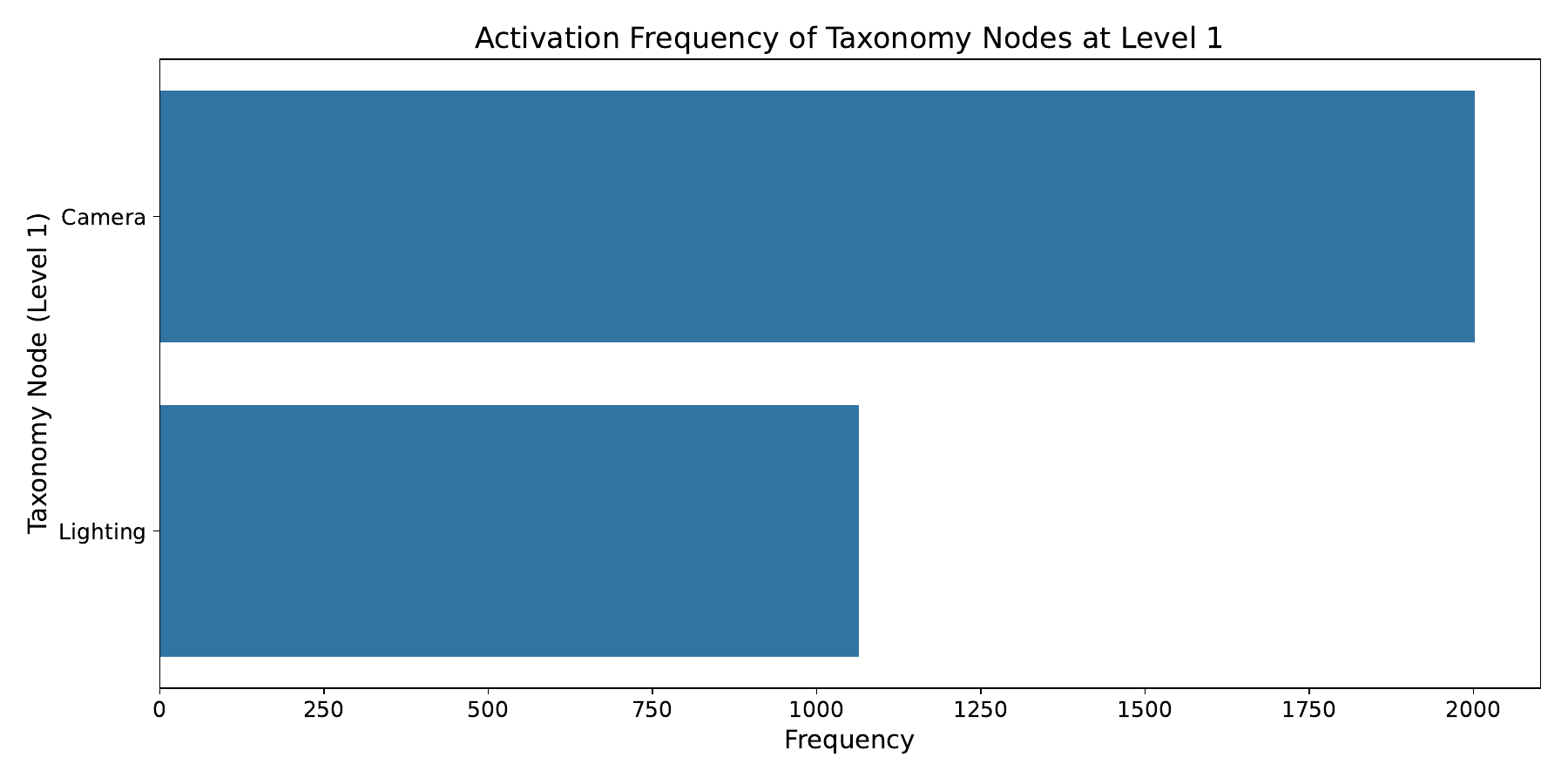}
    \caption{Level 1 Activations}
    \label{fig:cine_sub1}
  \end{subfigure}\hfill
  \begin{subfigure}[b]{0.48\textwidth}
    \includegraphics[width=\textwidth]{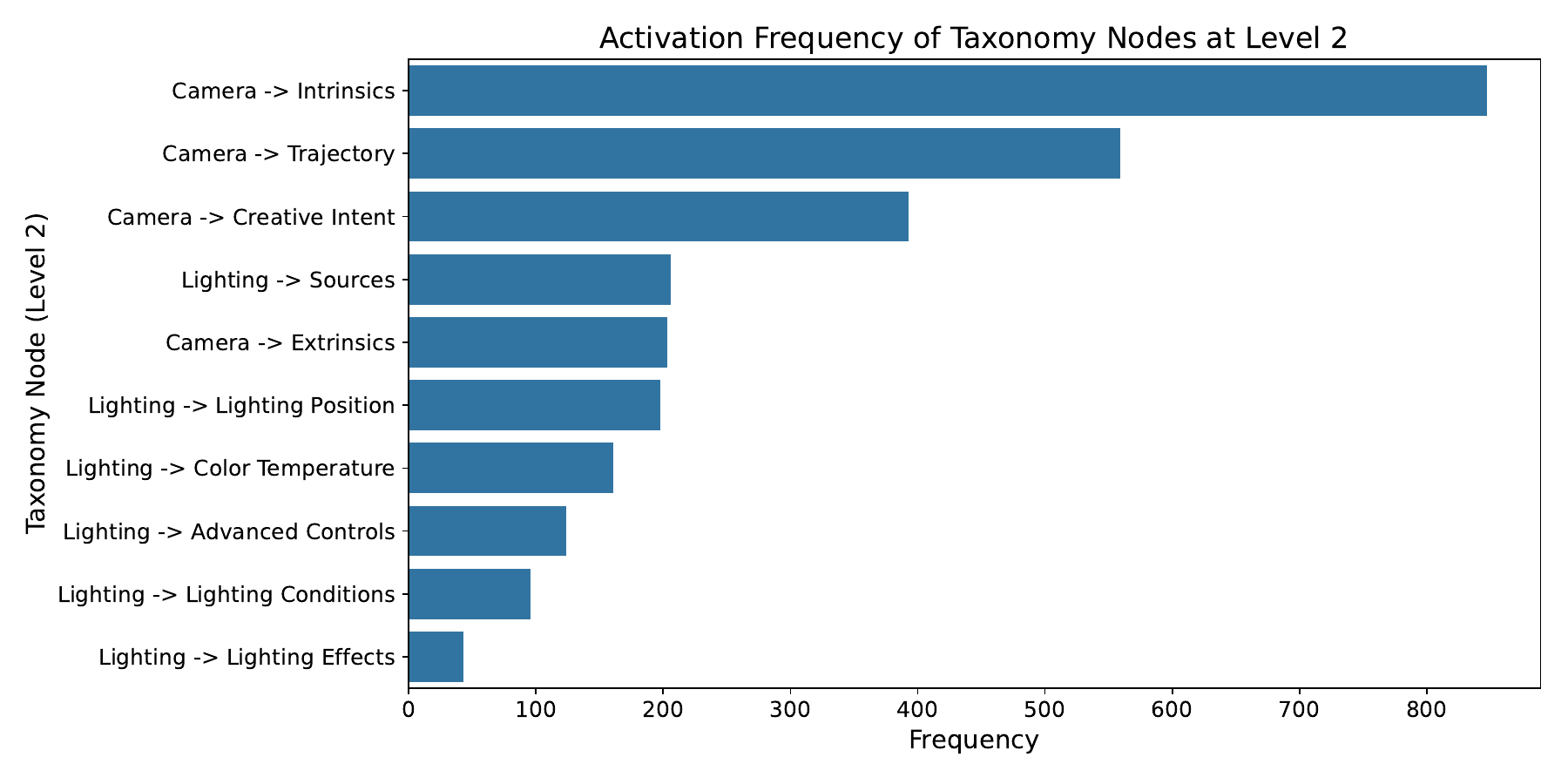}
    \caption{Level 2 Activations}
    \label{fig:cine_sub2}
  \end{subfigure}

[1em] 

  \begin{subfigure}[b]{0.48\textwidth}
    \includegraphics[width=\textwidth]{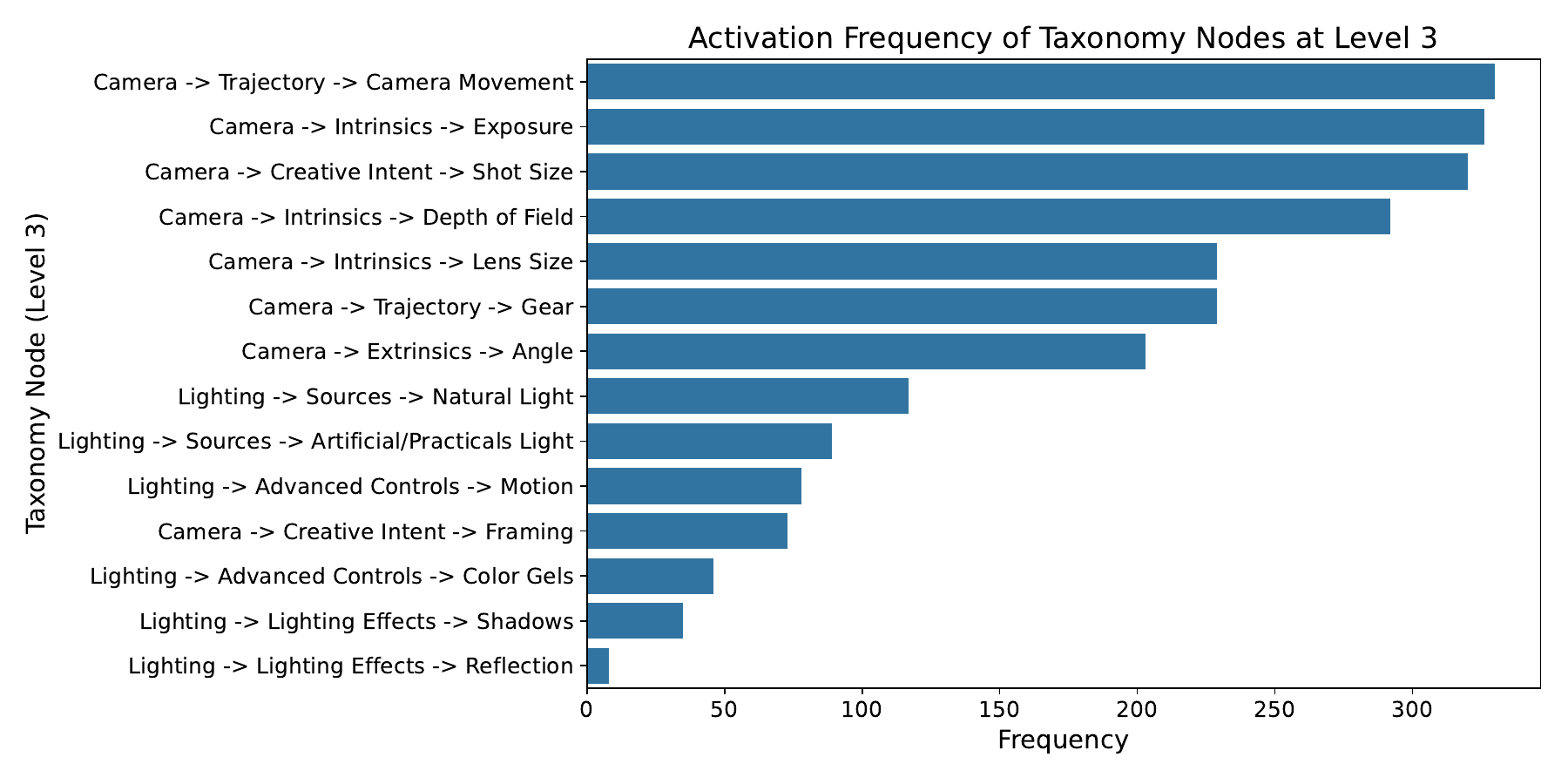}
    \caption{Level 3 Activations}
    \label{fig:cine_sub3}
  \end{subfigure}\hfill
  \begin{subfigure}[b]{0.48\textwidth}
    \includegraphics[width=\textwidth]{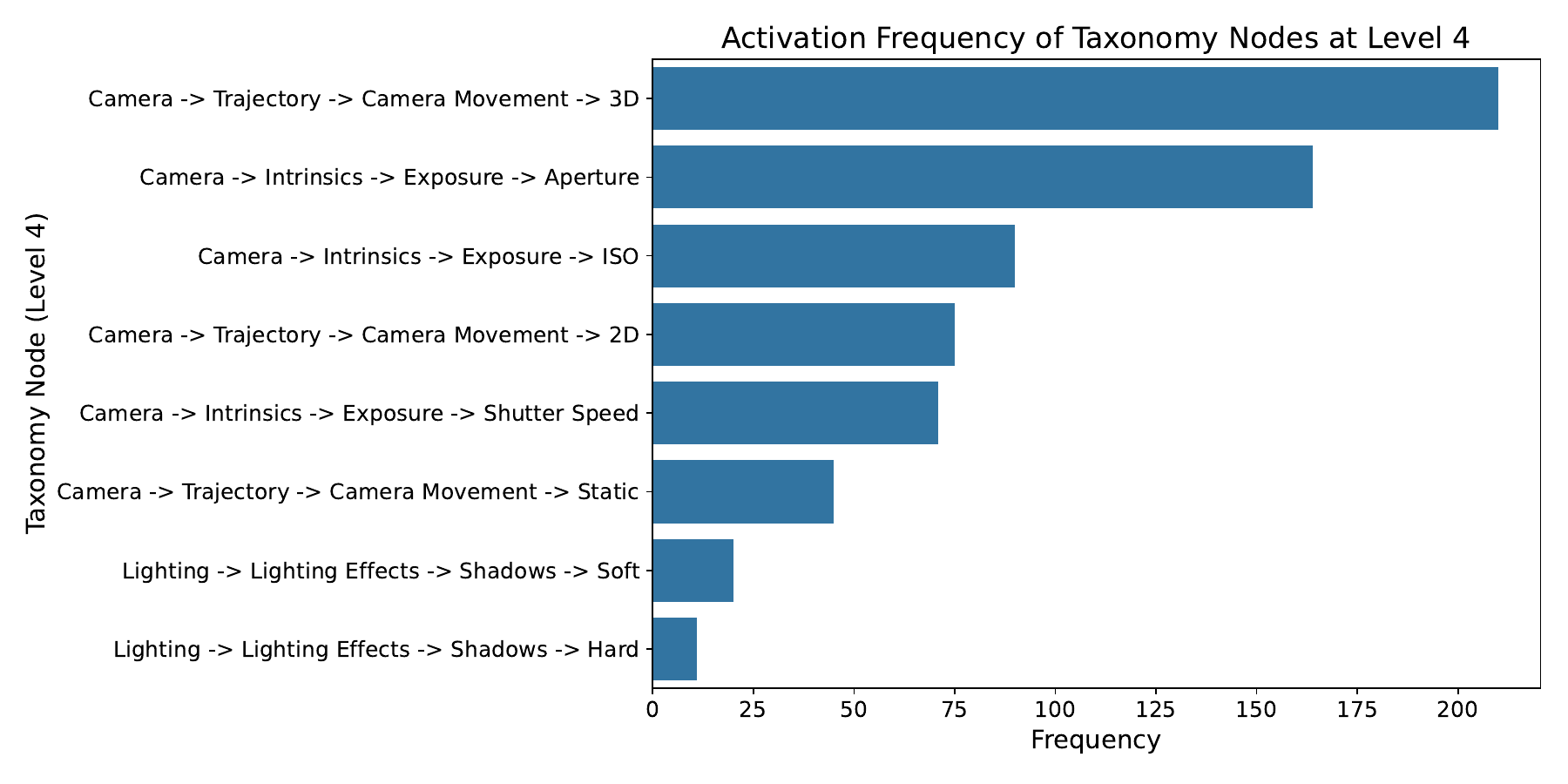}
    \caption{Level 4 Activations}
    \label{fig:cine_sub4}
  \end{subfigure}

  \caption{Node activations in Camera and Lighting taxonomies for the Cinematographer role in SCINE Visuals.}
  \label{fig:cine_activations}
\end{figure}

\begin{figure}[]
  \centering
   \begin{subfigure}[b]{0.48\textwidth}
    \includegraphics[width=\textwidth]{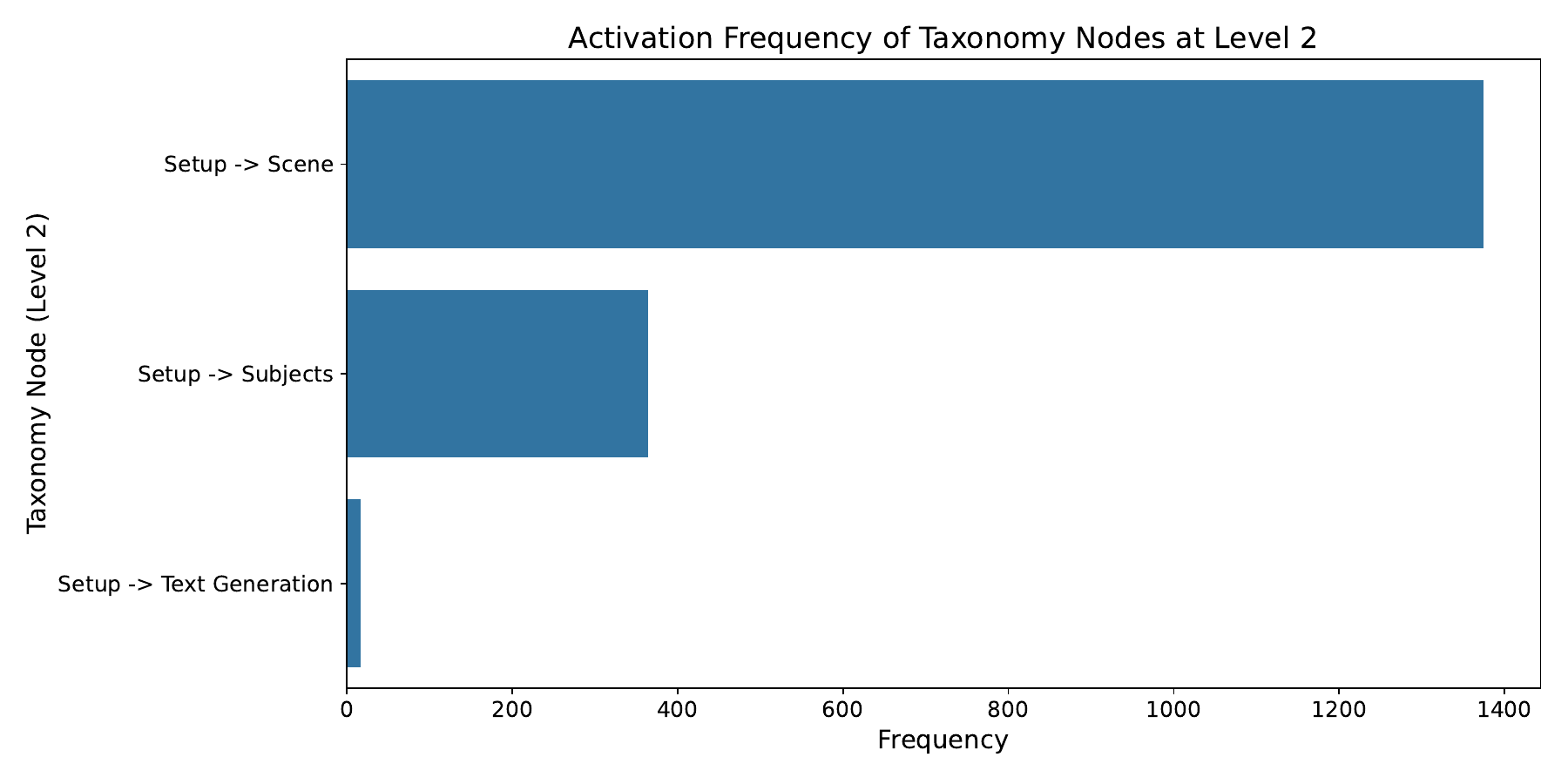}
    \caption{Level 2 Activations}
    \label{fig:pd_sub1}
  \end{subfigure}\hfill
  \begin{subfigure}[b]{0.48\textwidth}
    \includegraphics[width=\textwidth]{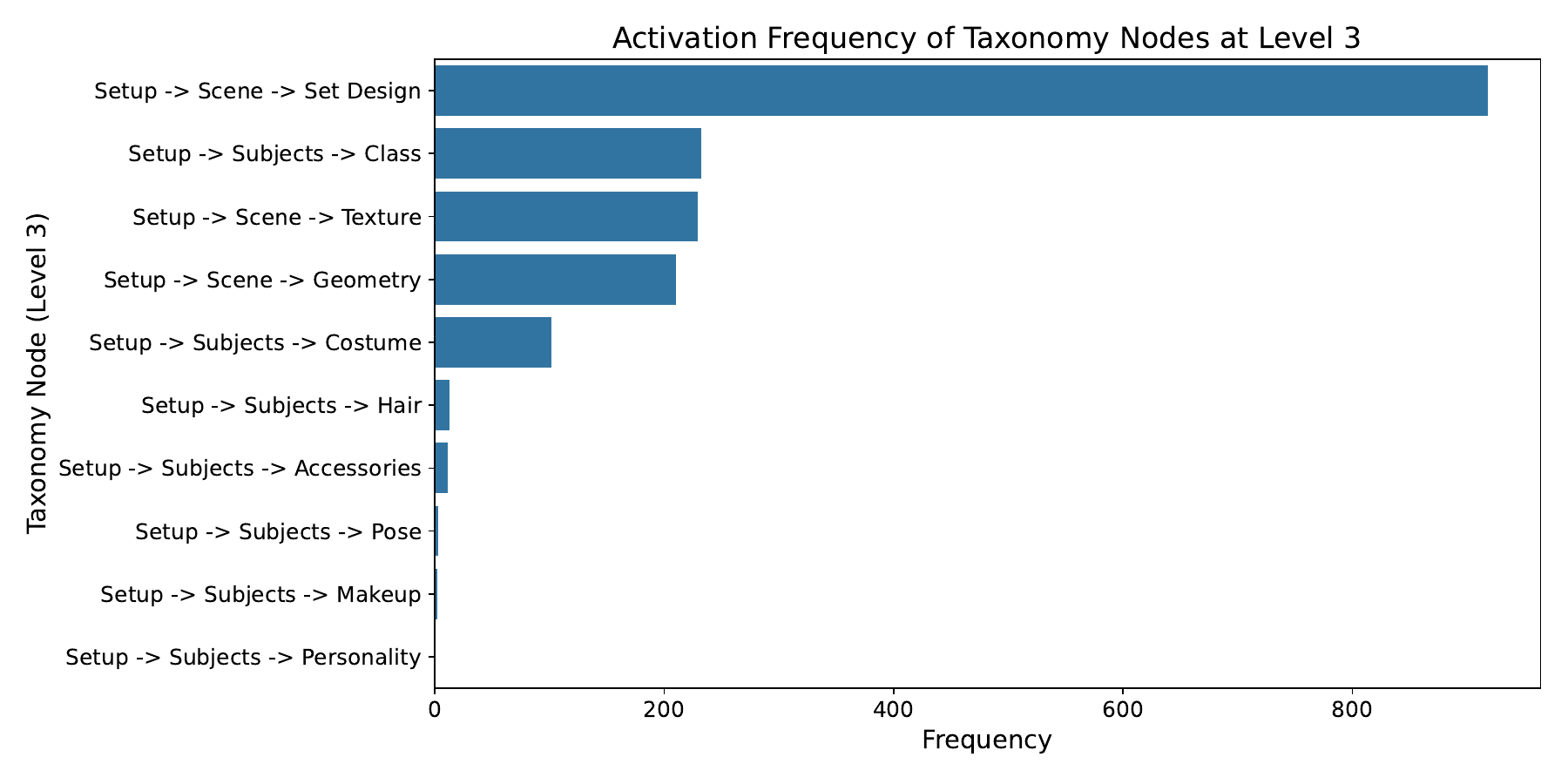}
    \caption{Level 3 Activations}
    \label{fig:pd_sub2}
  \end{subfigure}

  [1em] 

  \begin{subfigure}[b]{0.48\textwidth}
    \includegraphics[width=\textwidth]{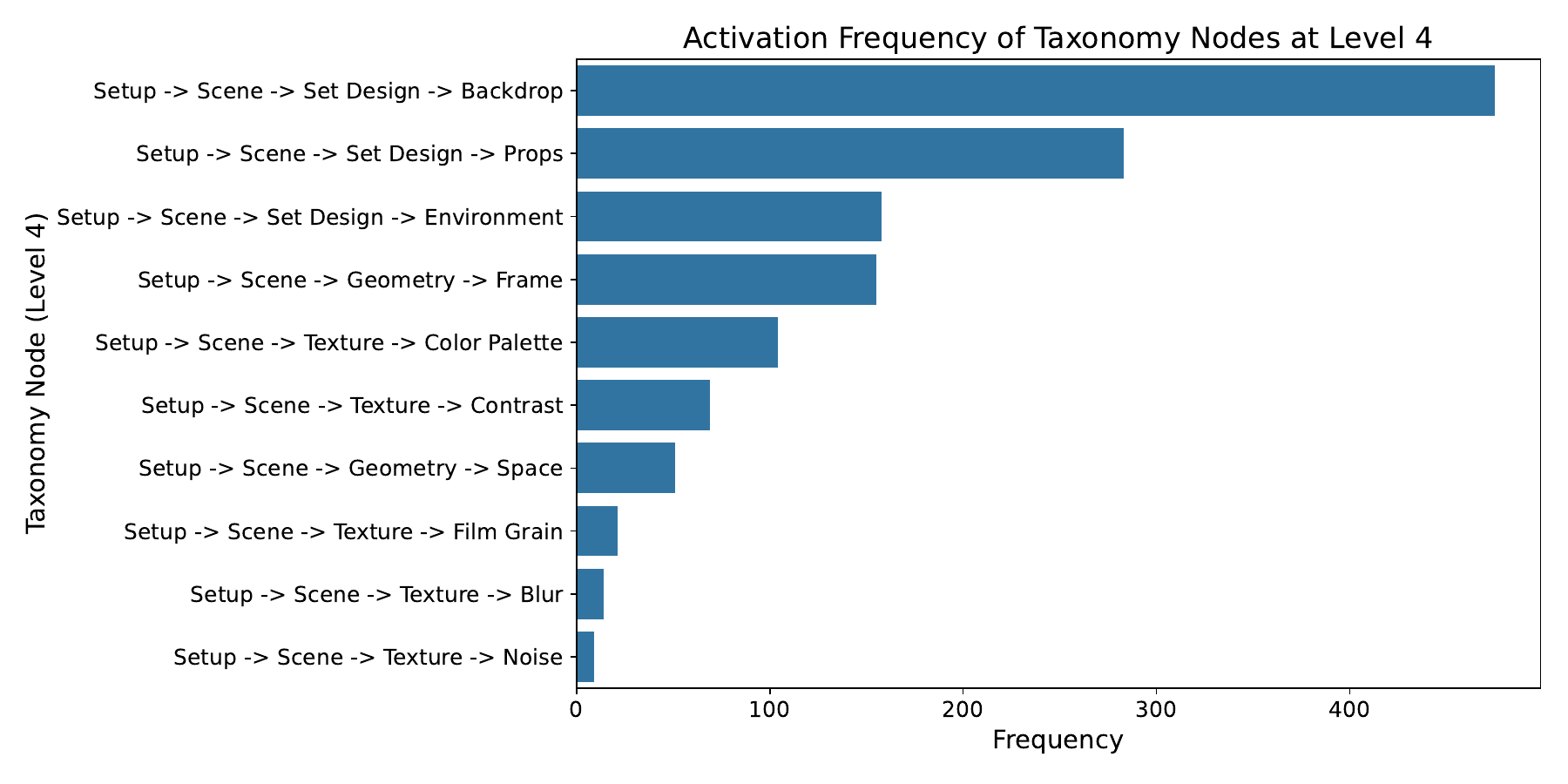}
    \caption{Level 4 Activations}
    \label{fig:pd_sub3}
  \end{subfigure}\hfill
  \begin{subfigure}[b]{0.48\textwidth}
    \includegraphics[width=\textwidth]{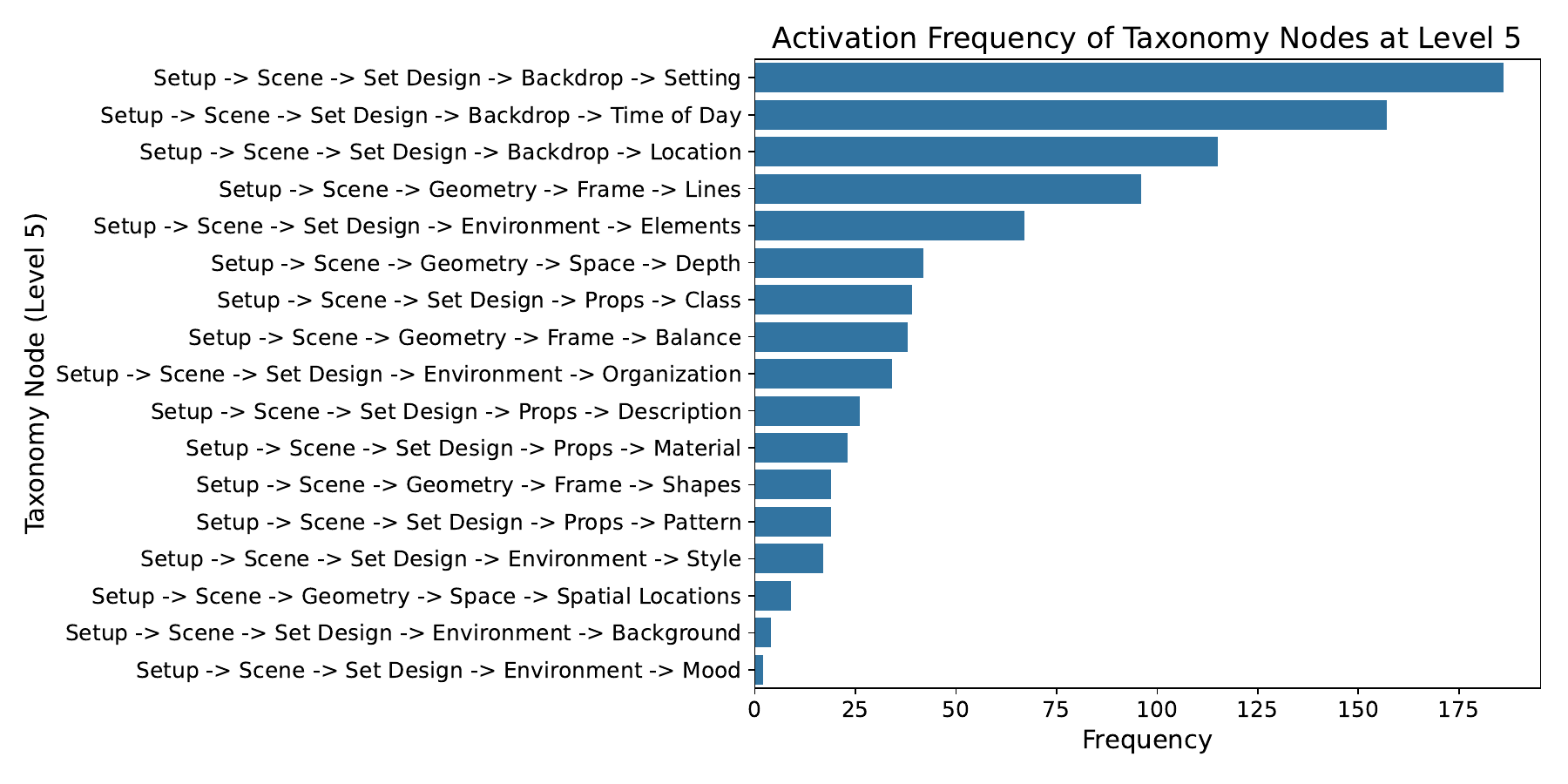}
    \caption{Level 5 Activations}
    \label{fig:pd_sub4}
  \end{subfigure}

  \caption{Node activations in Setup taxonomy for the Production Designer role in SCINE Visuals.}
  \label{fig:pd_activations}
\end{figure}

\begin{figure}[]
  \centering
   \begin{subfigure}[b]{0.48\textwidth}
    \includegraphics[width=\textwidth]{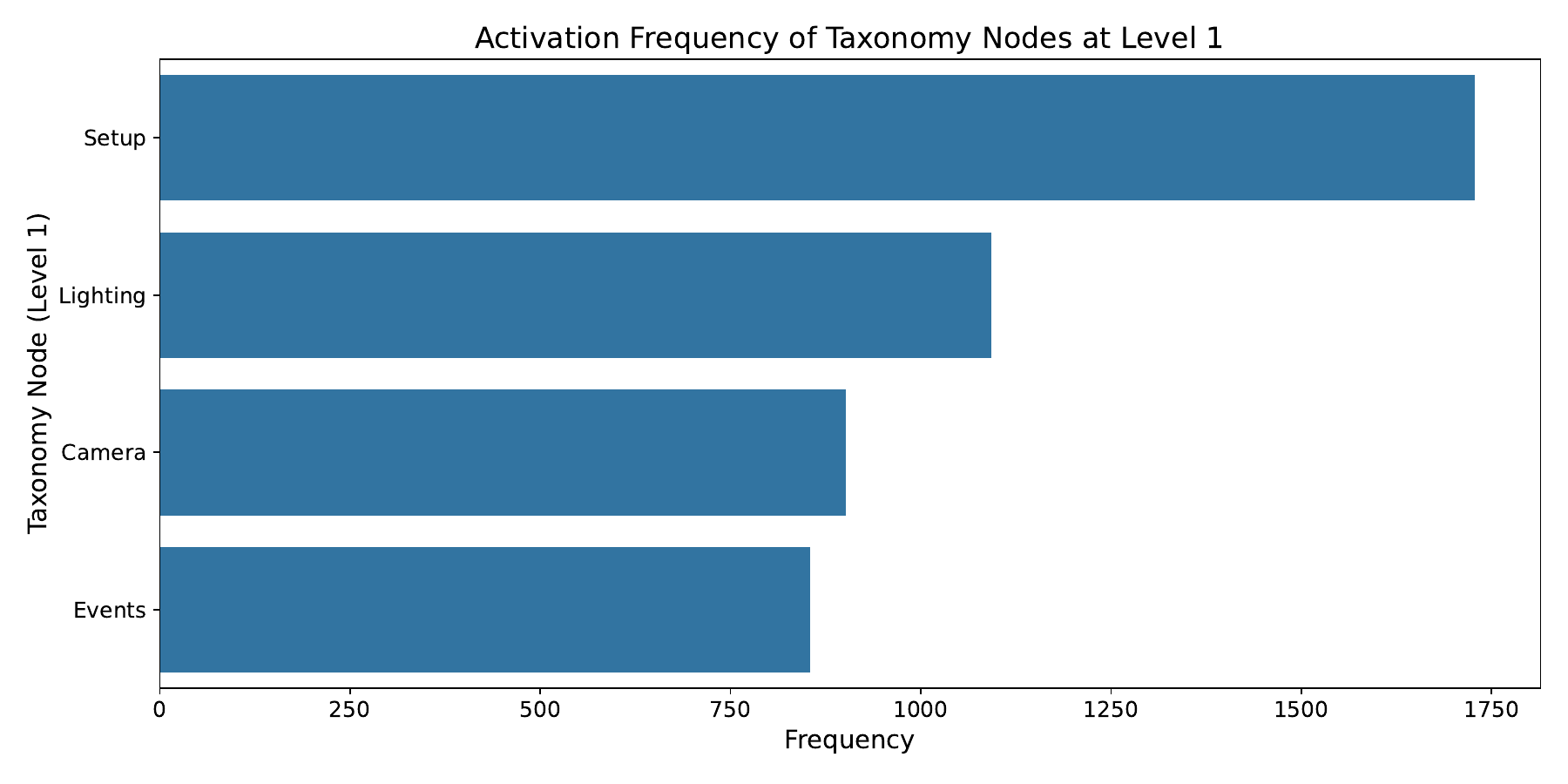}
    \caption{Level 1 Activations}
    \label{fig:director_sub1}
  \end{subfigure}\hfill
  \begin{subfigure}[b]{0.48\textwidth}
    \includegraphics[width=\textwidth]{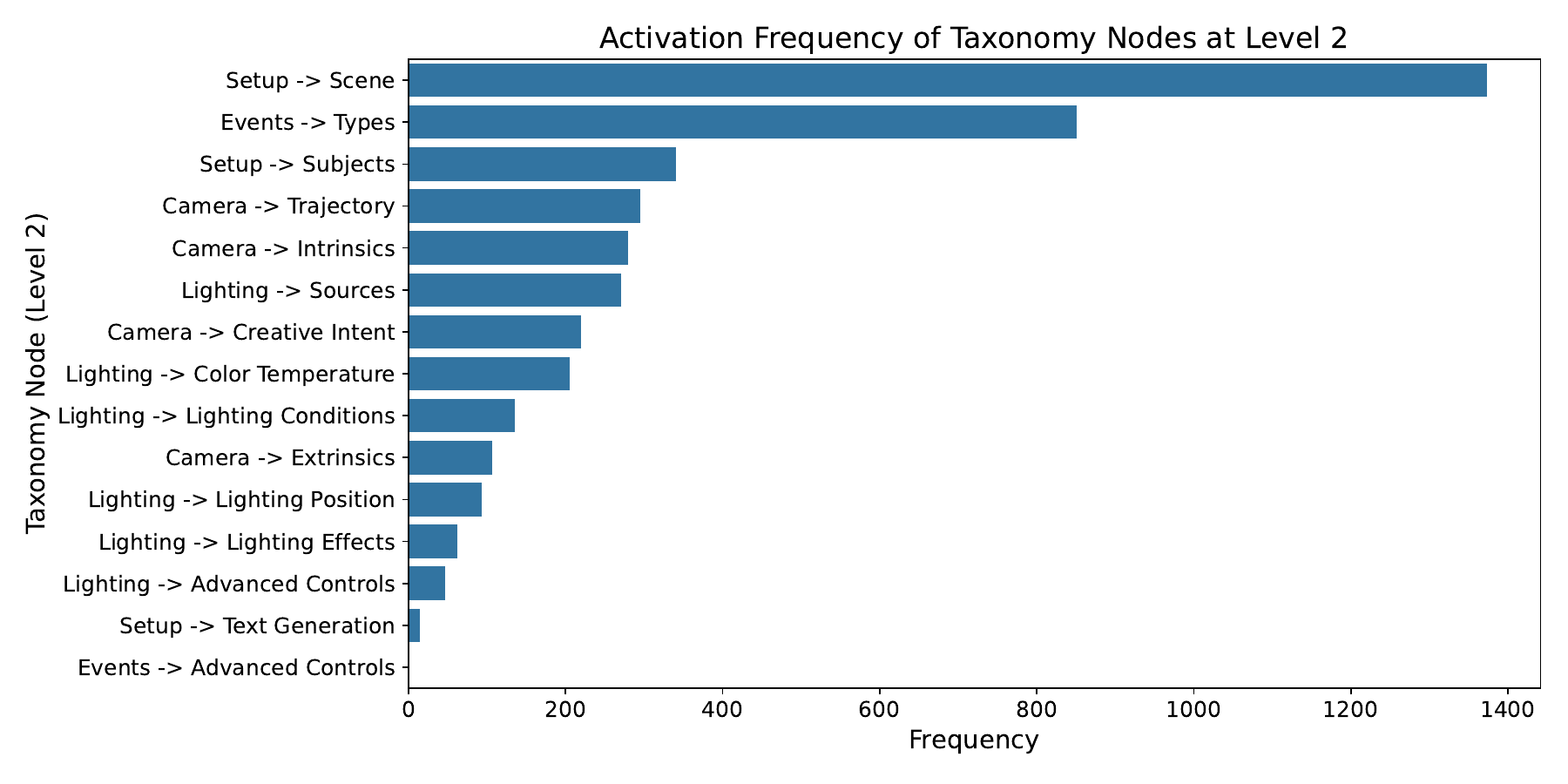}
    \caption{Level 2 Activations}
    \label{fig:director_sub2}
  \end{subfigure}

  [1em] 

  \begin{subfigure}[b]{0.48\textwidth}
    \includegraphics[width=\textwidth]{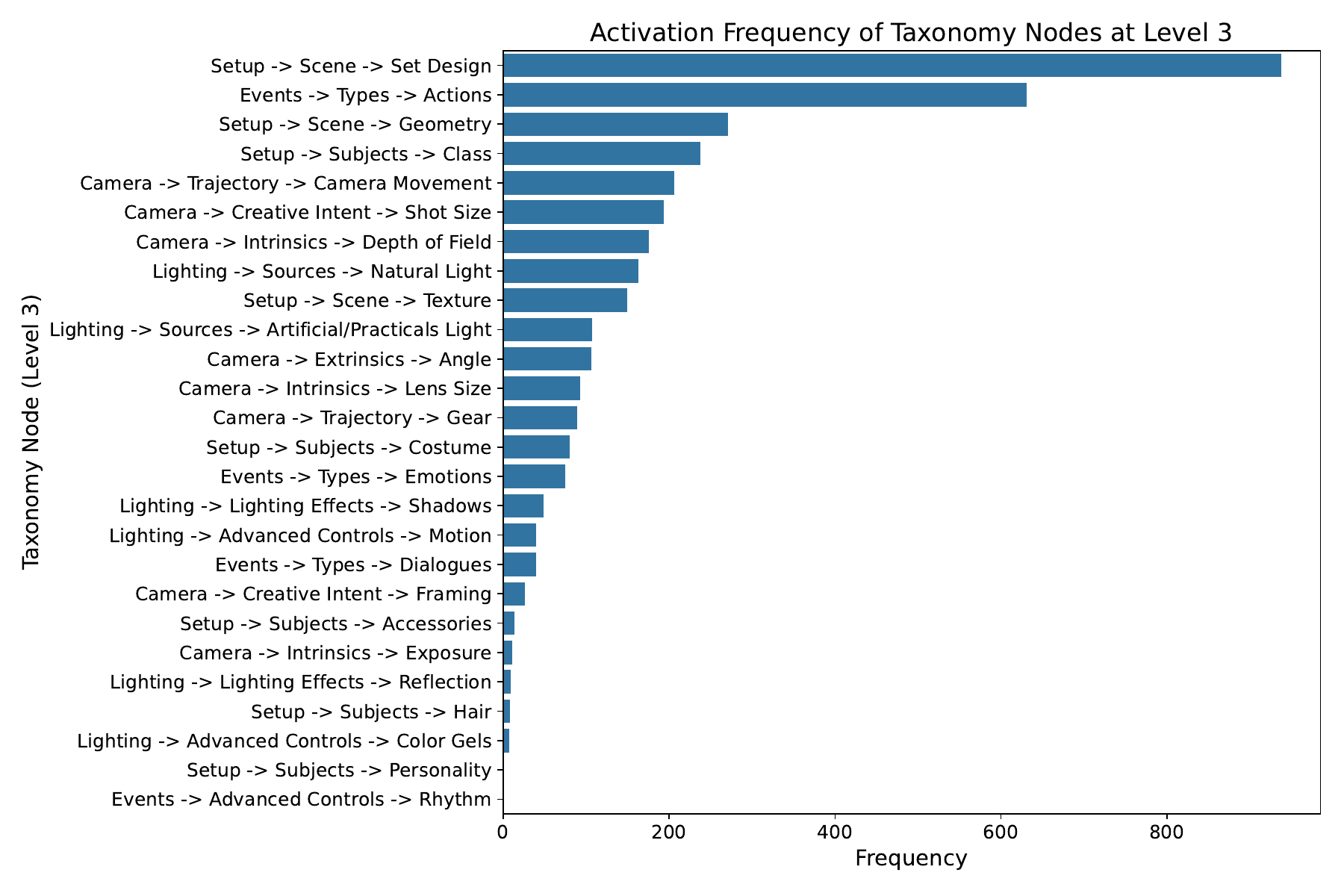}
    \caption{Level 3 Activations}
    \label{fig:director_sub3}
  \end{subfigure}\hfill
  \begin{subfigure}[b]{0.48\textwidth}
    \includegraphics[width=\textwidth]{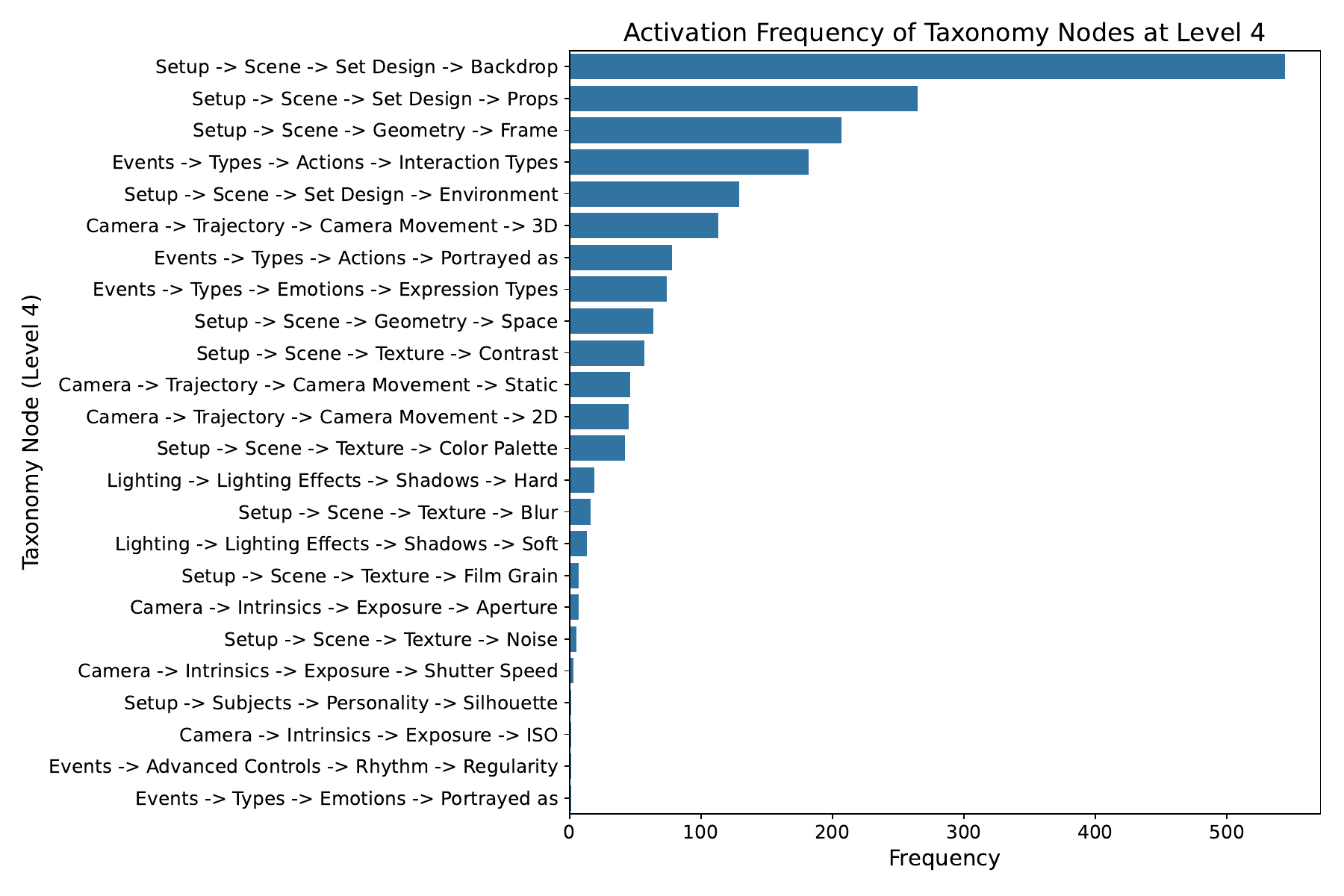}
    \caption{Level 4 Activations}
    \label{fig:director_sub4}
  \end{subfigure}

  \caption{Node activations in All taxonomies for the Director role in SCINE Visuals.}
  \label{fig:director_activations}
\end{figure}

Figures \ref{fig:cine_activations}, \ref{fig:pd_activations}, and \ref{fig:director_activations} show the distribution of activated nodes in SCINE Visuals, aggregated at the node level, across the roles of Cinematographer, Production Designer, and Director, respectively. As shown, our prompts cover a broad distribution of nodes across all the taxonomies.

\subsection{Annotation Details} \label{sec:annotation_details}

\begin{figure}[]
  \centering
  \includegraphics[width=\linewidth]{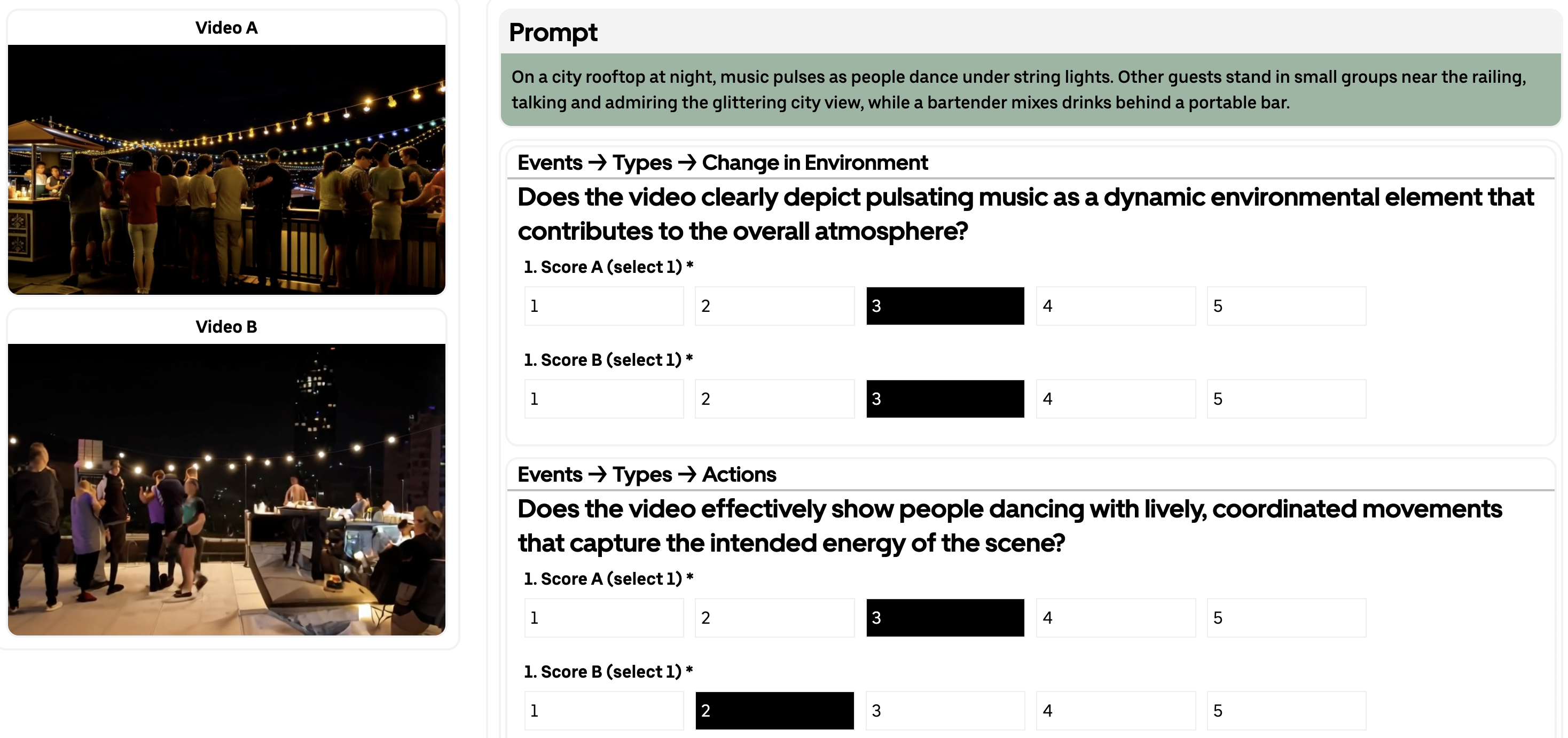}
  \caption {User Interface used by annotators to perform evaluations.}
  \label{fig:uber_ui}
\end{figure}

\begin{figure}[]
  \centering
  \includegraphics[width=\linewidth]{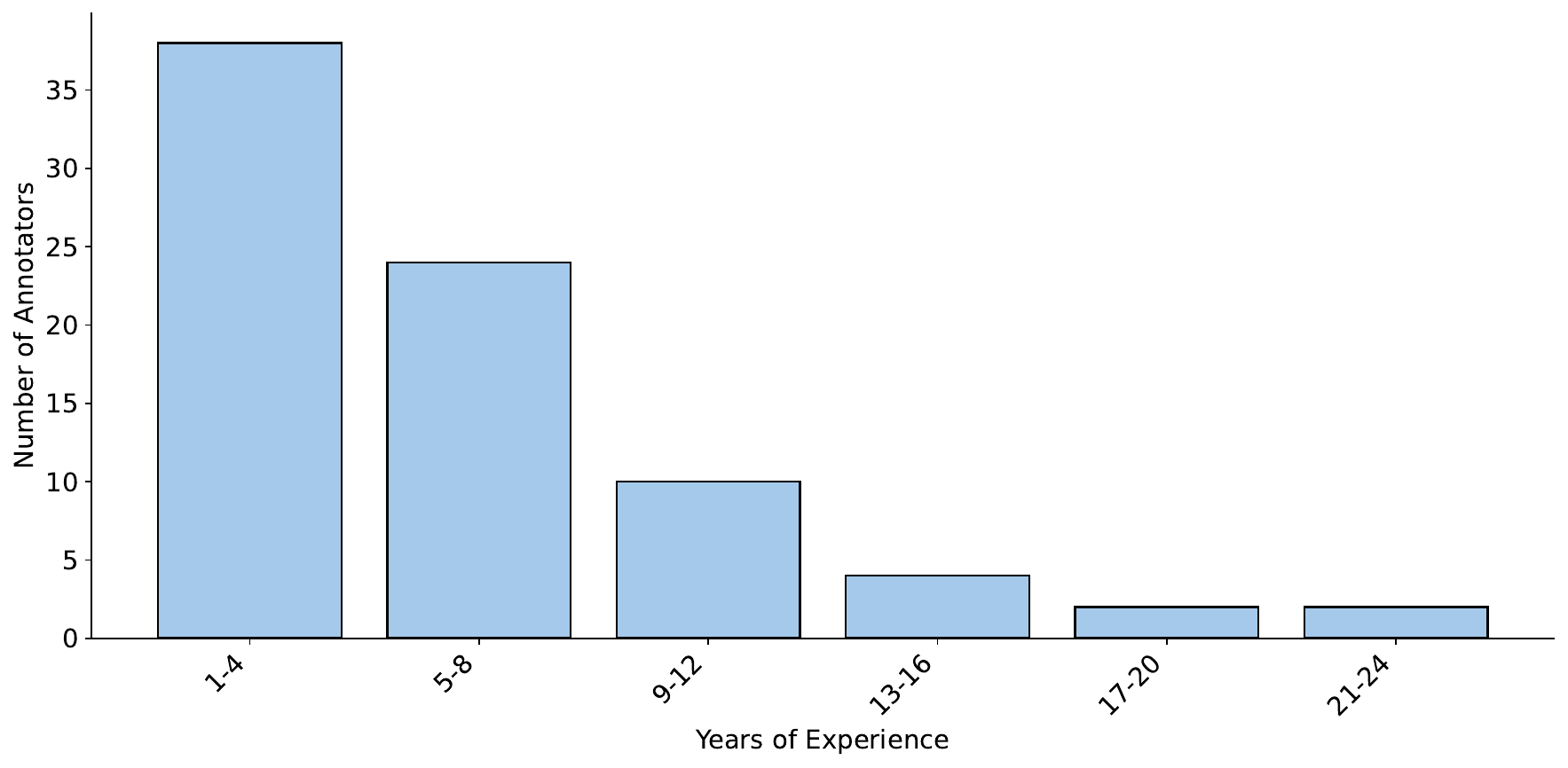}
  \caption {Distribution of the years of film production experience amongst human annotators in our evaluation setup.}
  \label{fig:yoe_histogram}
\end{figure}

Figure \ref{fig:uber_ui} shows the annotation interface used by human annotators during evaluation.  We also present the distribution of annotators' years of experience in film production in Figure \ref{fig:yoe_histogram}. While annotations for cinematic controls can be subjective, especially given the large number of control nodes, we try our best to mitigate this by providing clear rating guidelines to annotators for each control node. Table \ref{tab:rating-scales} presents a minimal example of the rating guidelines shared with the annotators.

\begin{table}[h]
  \centering
  \caption{Examples of rating guidelines provided to human annotators for different control nodes, across all taxonomies.}
  \label{tab:rating-scales}
  \resizebox{\linewidth}{!}{
  \begin{tabular}{@{}l r p{0.7\textwidth}@{}}
    \toprule
    \textbf{Dimension} & \textbf{Score} & \textbf{What to Look For} \\
    \midrule
    \multirow{5}{*}{Low Angle} 
      & 1 & Camera is at or above eye level, not low angle at all. \\
      & 2 & Slight upward tilt, but still feels neutral. \\
      & 3 & Below subject, mild upward view, light impact. \\
      & 4 & Clear low angle, subject looks larger or imposing. \\
      & 5 & Strong low angle, subject dominates, towering presence. \\
    \midrule
    \multirow{5}{*}{Overlapping Actions} 
      & 1 & No action or one action is present. \\
      & 2 & Actions are isolated or unrelated. \\
      & 3 & Timing is off, they start or end awkwardly. \\
      & 4 & Some overlap, but hard to follow. \\
      & 5 & Fluid overlap, actions feel natural and dynamic together. \\
    \midrule
    \multirow{5}{*}{Back Light Position} 
      & 1 & Light clearly comes from front or side, no rim light or background separation. \\
      & 2 & Some edge lighting, but not consistent or strong, subject may still blend into background. \\
      & 3 & Back light is {partially visible}, outline is hinted but not clear on full subject. \\
      & 4 & Back light is {clearly present}, rim light separates subject from background. \\
      & 5 & Strong back light effect, glowing edges around hair or shoulders. Subject clearly pops against the background. Perfect match. \\
    \midrule
    \multirow{5}{*}{Symmetrical Frame Balance} 
      & 1 & Composition is clearly asymmetrical. \\
      & 2 & Some repeating elements, but no visual mirror. \\
      & 3 & Partial symmetry or {mirrored clutter} that’s not clean. \\
      & 4 & Almost perfect symmetry, small inconsistencies exist. \\
      & 5 & Clear and precise symmetry, mirrored subjects, reflections, or centered framing. Strong and intentional. \\
    \bottomrule
  \end{tabular}
  }
\end{table}

\clearpage
\subsection{Statistical tests} \label{sec:statistical_tests}

\begin{figure}[!htbp]
  \centering

  \begin{subfigure}[t]{0.48\linewidth}
    \includegraphics[width=\linewidth]{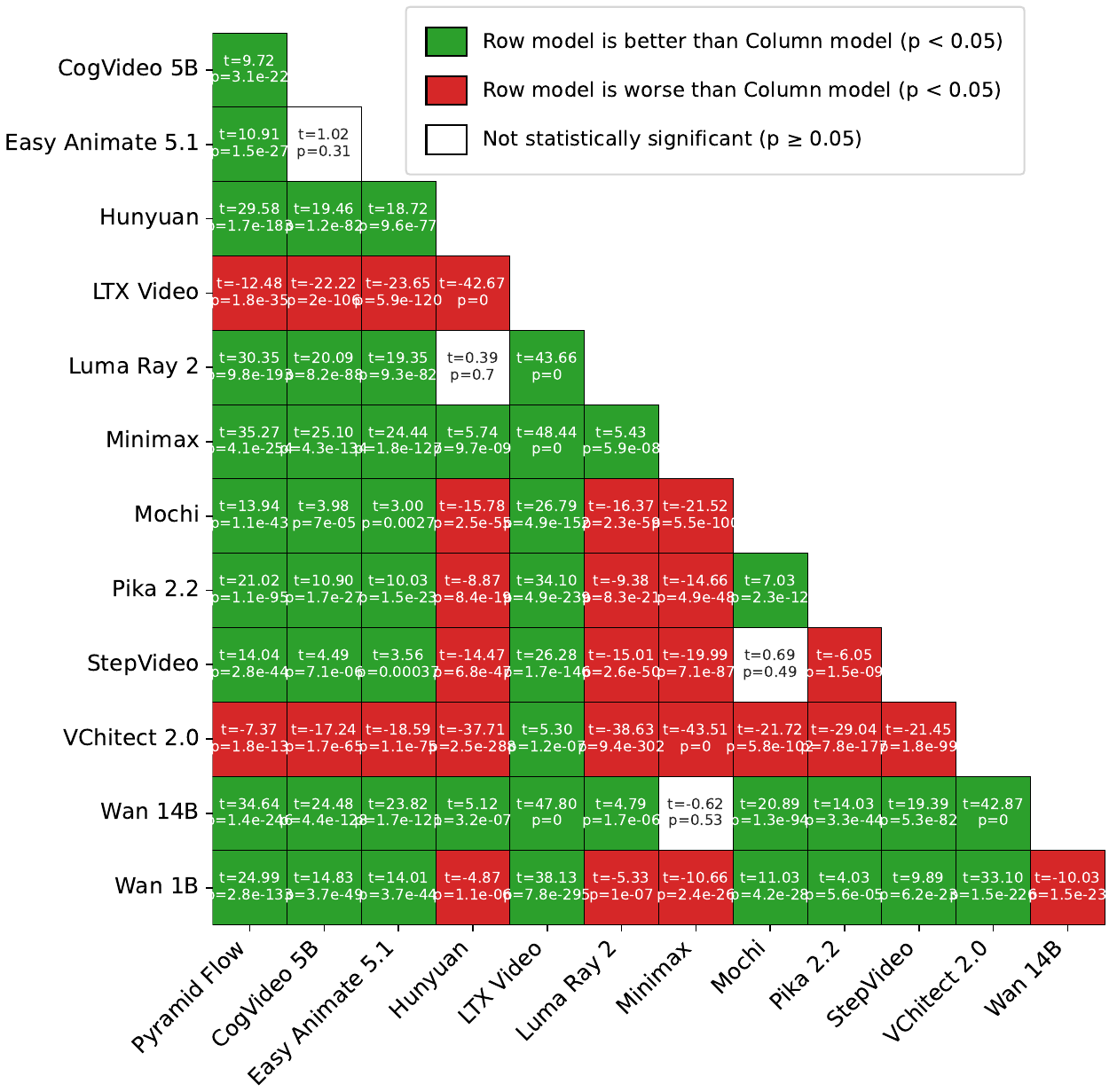}
    \caption{Pairwise t-tests across models on emotions}
    \label{fig:events_emotion}
  \end{subfigure}
  \begin{subfigure}[t]{0.48\linewidth}
    \includegraphics[width=\linewidth]{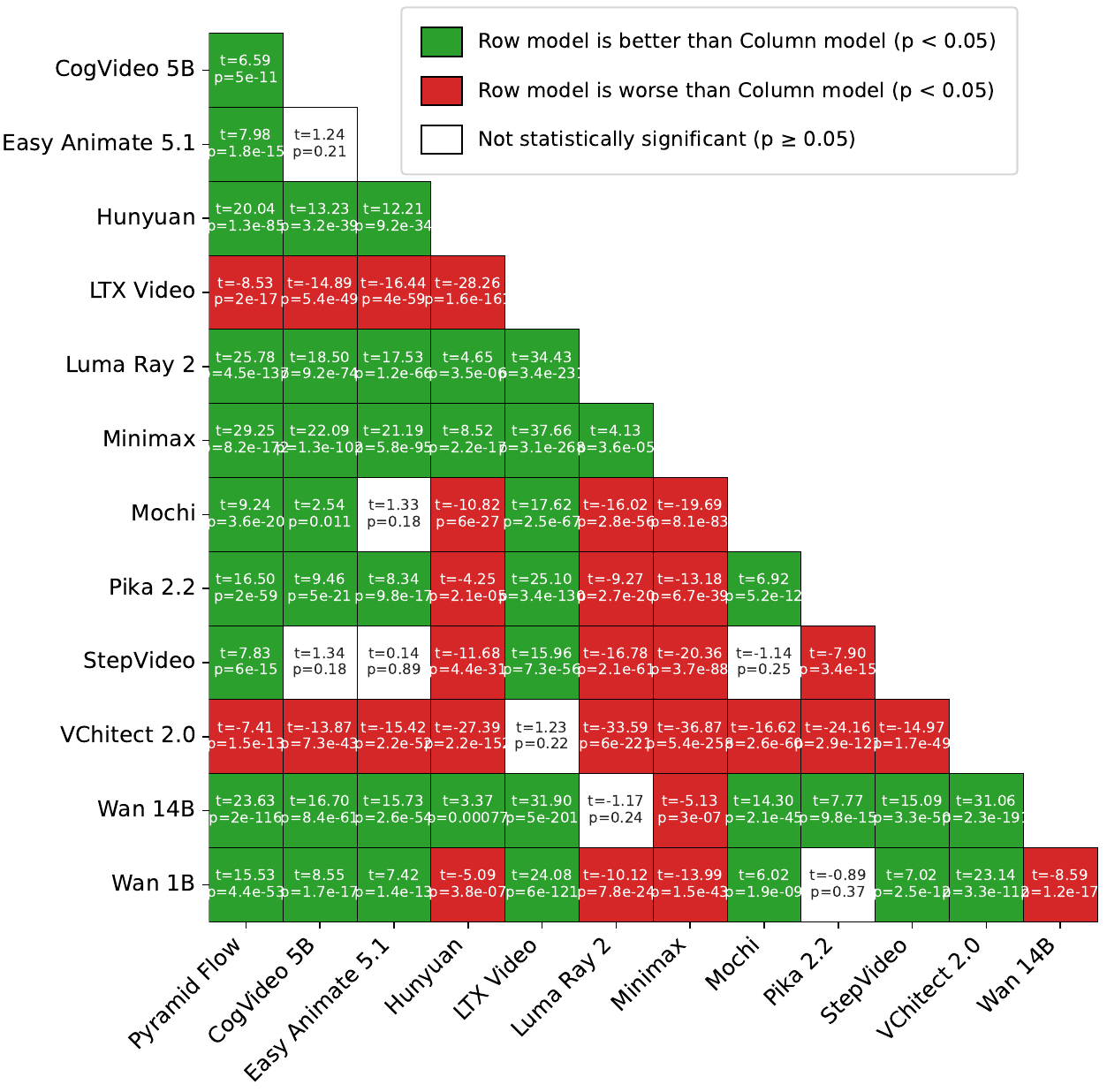}
    \caption{Pairwise t-tests across models on dialogues}
    \label{fig:events_dialogue}
  \end{subfigure}

  \vspace{1em}

  \begin{subfigure}[b]{0.6\textwidth} 
    \centering
   \includegraphics[width=\linewidth]{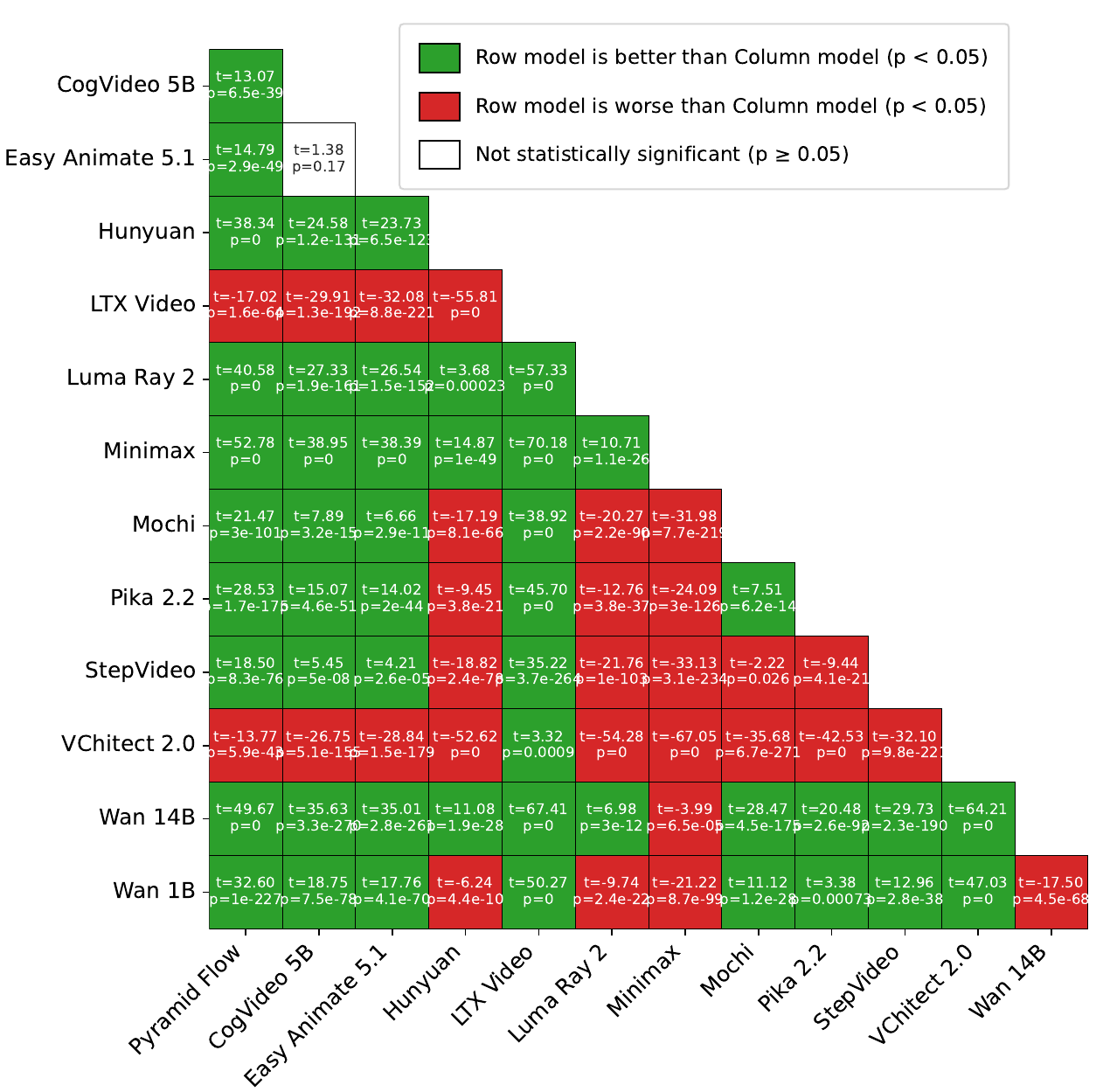}
    \caption{Pairwise t-tests across models on actions}
    \label{fig:events_action}
  \end{subfigure}

  \caption{Statistical comparison matrices for Events: Emotions, Dialogue, and Actions.}
  \label{fig:t_test_events}
\end{figure}

\begin{figure}[!tp]
  \centering

  \begin{subfigure}[b]{0.48\textwidth}
    \centering
     \includegraphics[width=\linewidth]{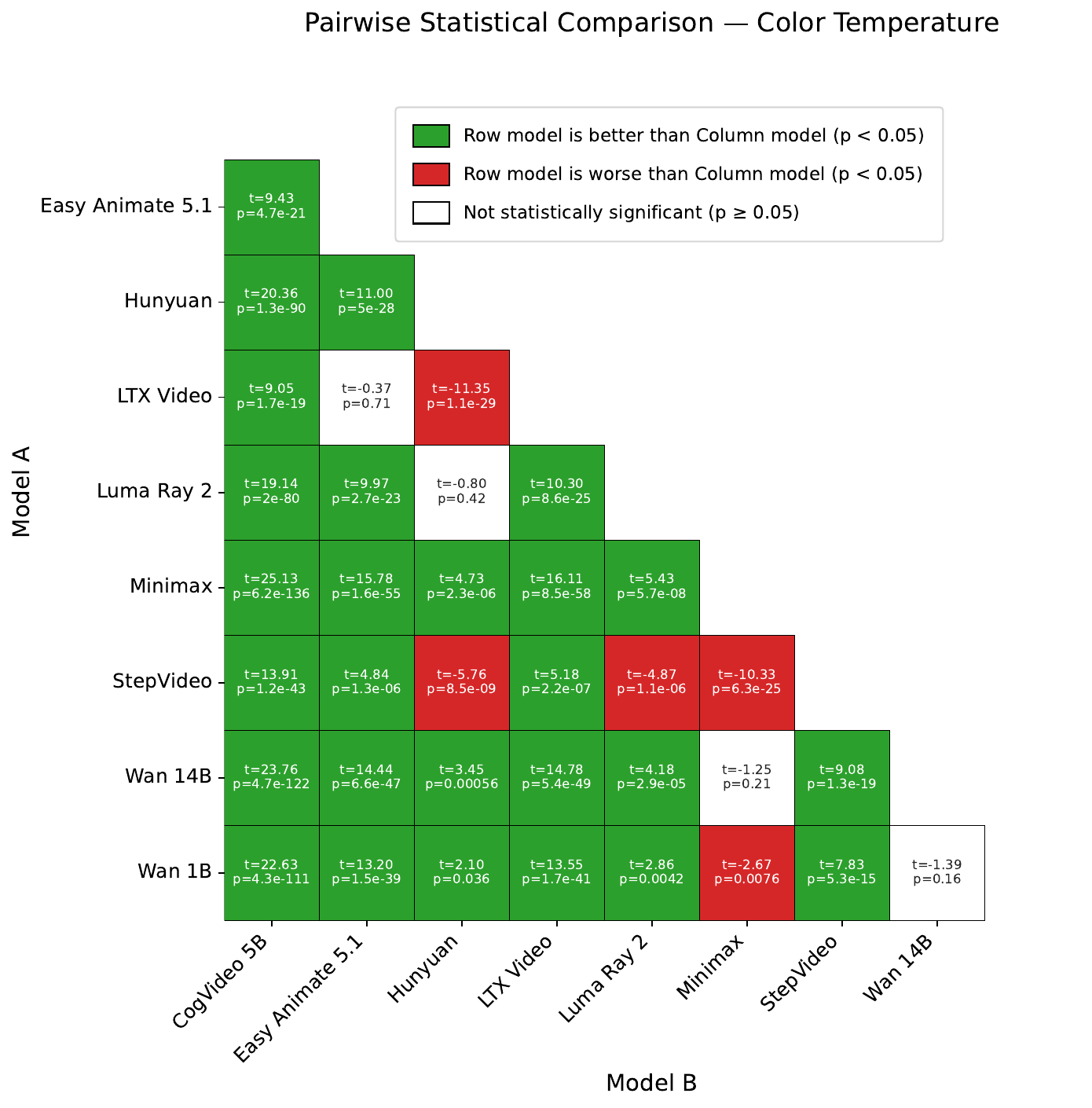}
    \caption{Pairwise t-tests across models on Color Temperature}
    \label{fig:lighting_ct}
  \end{subfigure}
  \hfill
  \begin{subfigure}[b]{0.48\textwidth}
    \centering
   \includegraphics[width=\linewidth]{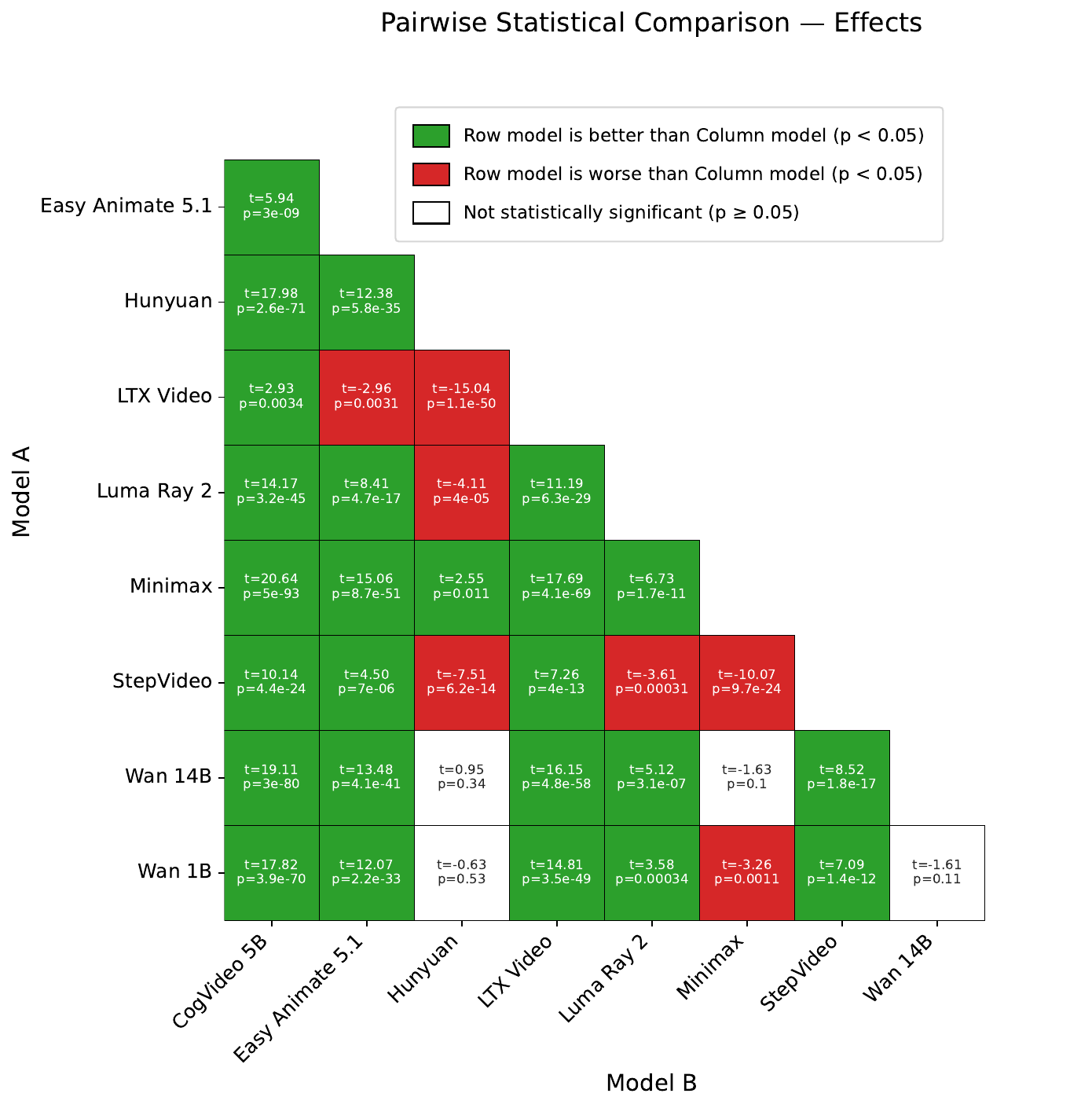}
    \caption{Pairwise t-tests across models on Lighting Effects}
    \label{fig:lighting_effects}
  \end{subfigure}

  \begin{subfigure}[b]{0.48\textwidth}
    \centering
    \includegraphics[width=\linewidth]{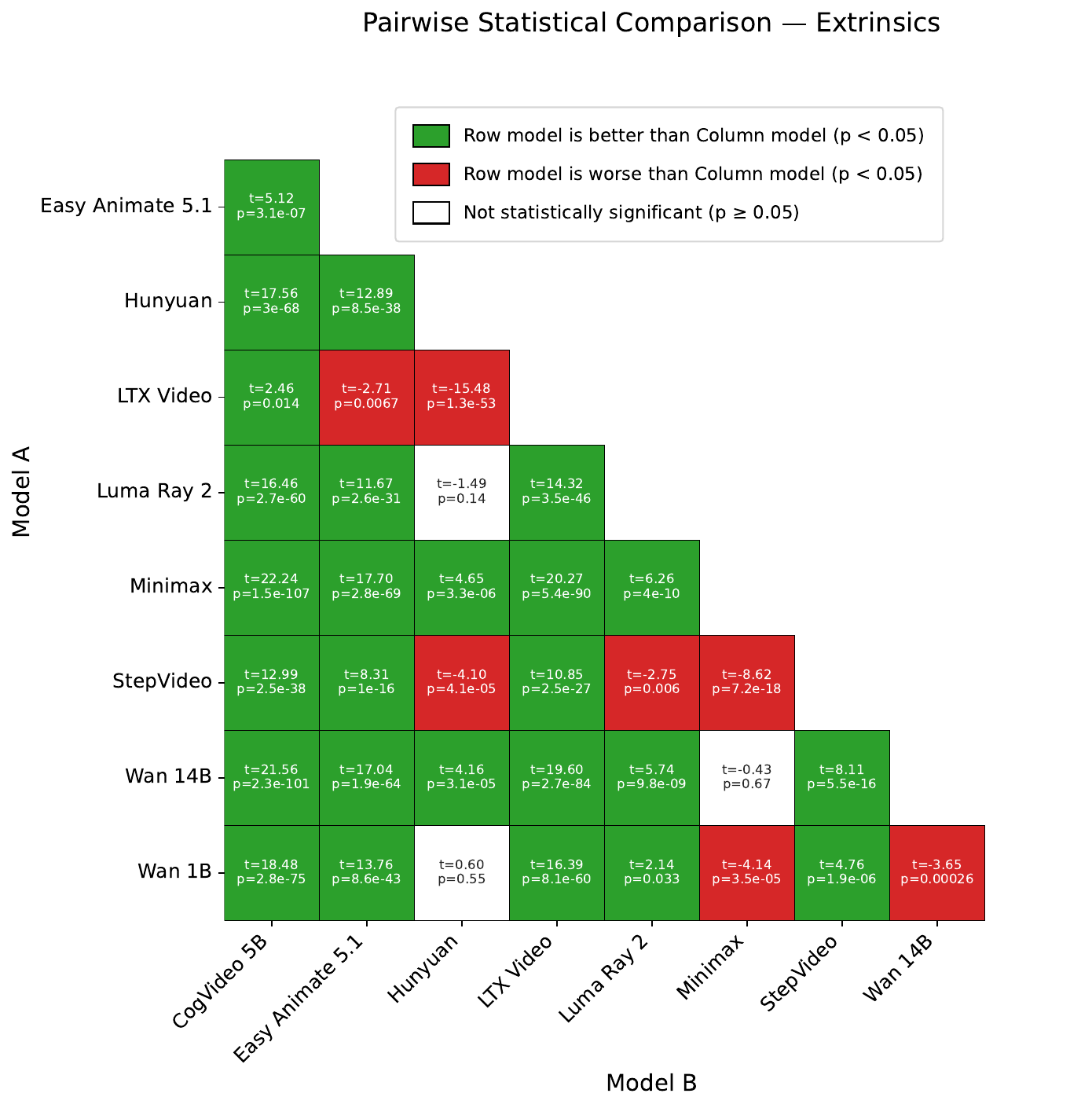}
    \caption{Pairwise t-tests across models on Extrinsics}
    \label{fig:camera_extrinsics}
  \end{subfigure}
  \hfill
  \begin{subfigure}[b]{0.48\textwidth}
    \centering
     \includegraphics[width=\linewidth]{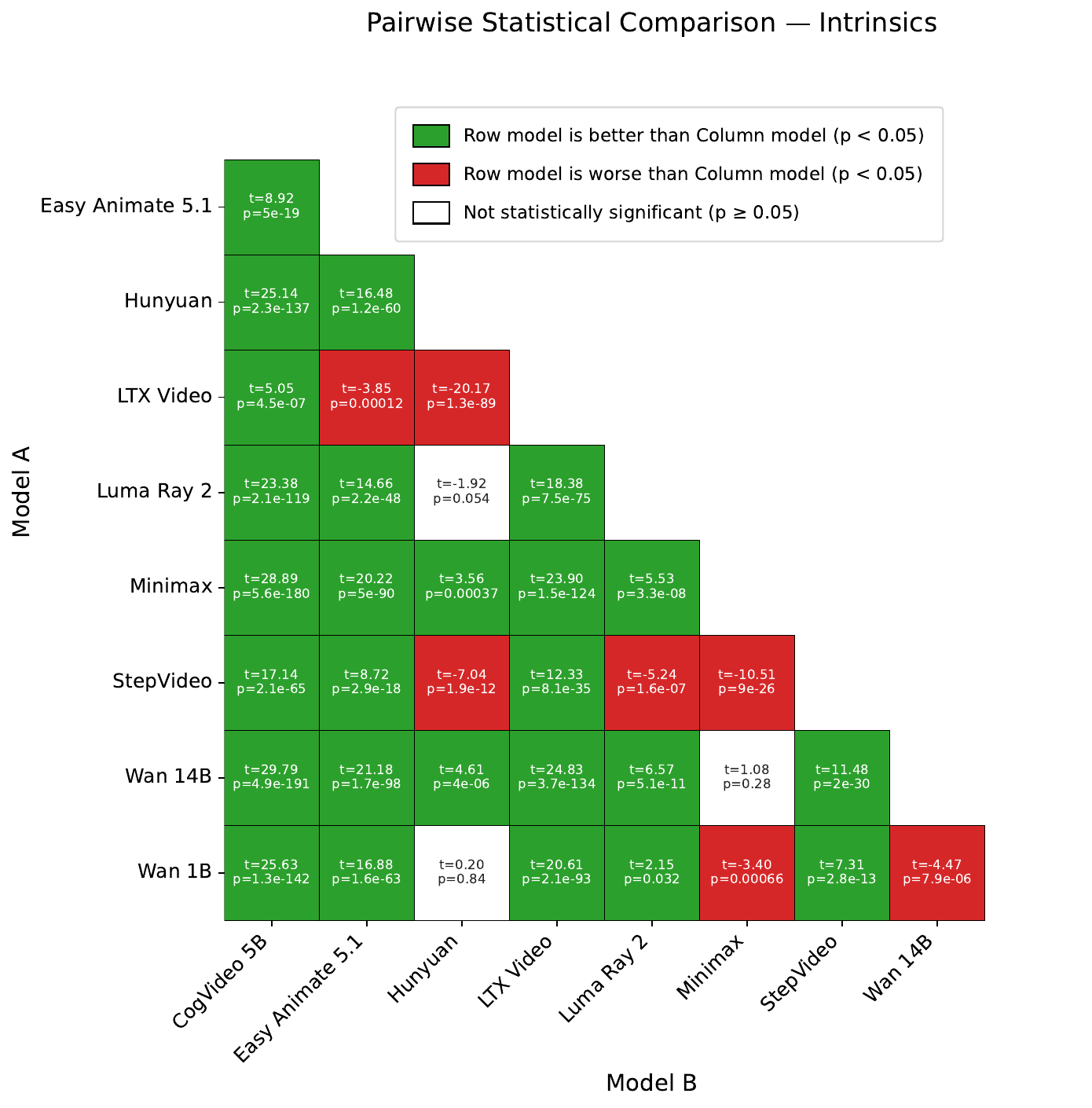}
    \caption{Pairwise t-tests across models on Intrinsics}
    \label{fig:camera_intrinsics}
  \end{subfigure}


  \begin{subfigure}[b]{0.5\textwidth}
    \centering
    \includegraphics[width=\linewidth]{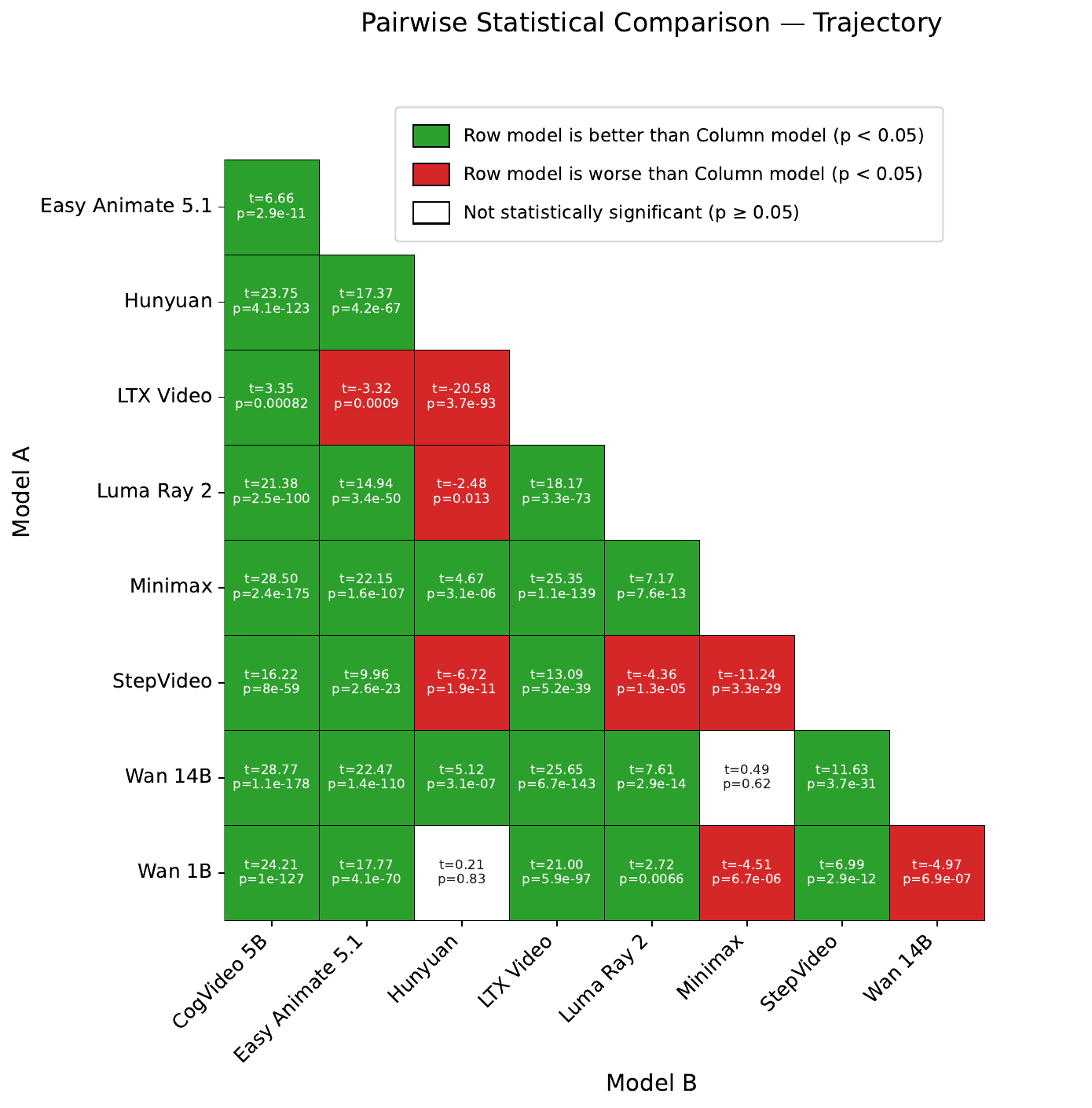}
    \caption{Pairwise t-tests across models on Trajectory}
    \label{fig:camera_trajectory}
  \end{subfigure}

  \caption{Statistical comparison matrices for Camera and Lighting}
  \label{fig:t_test_camera}
\end{figure}

\begin{figure}[!htbp]
  \centering
  \begin{subfigure}[b]{0.48\textwidth}
    \centering
     \includegraphics[width=\linewidth]{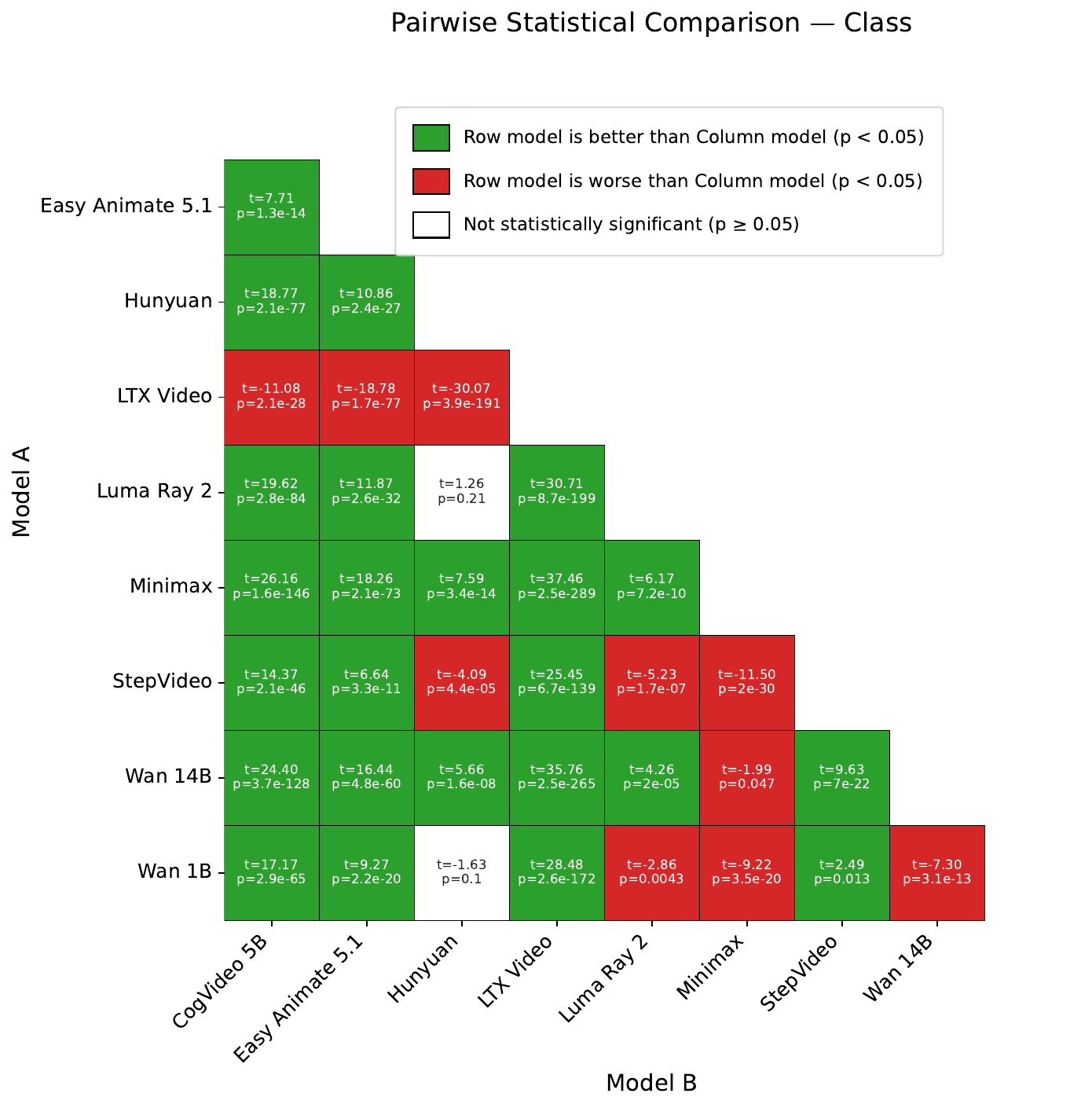}
    \caption{Pairwise t-tests across models on Subject Class}
    \label{fig:setup_subject_class}
  \end{subfigure}
  \hfill
  \begin{subfigure}[b]{0.48\textwidth}
    \centering
    \includegraphics[width=\linewidth]{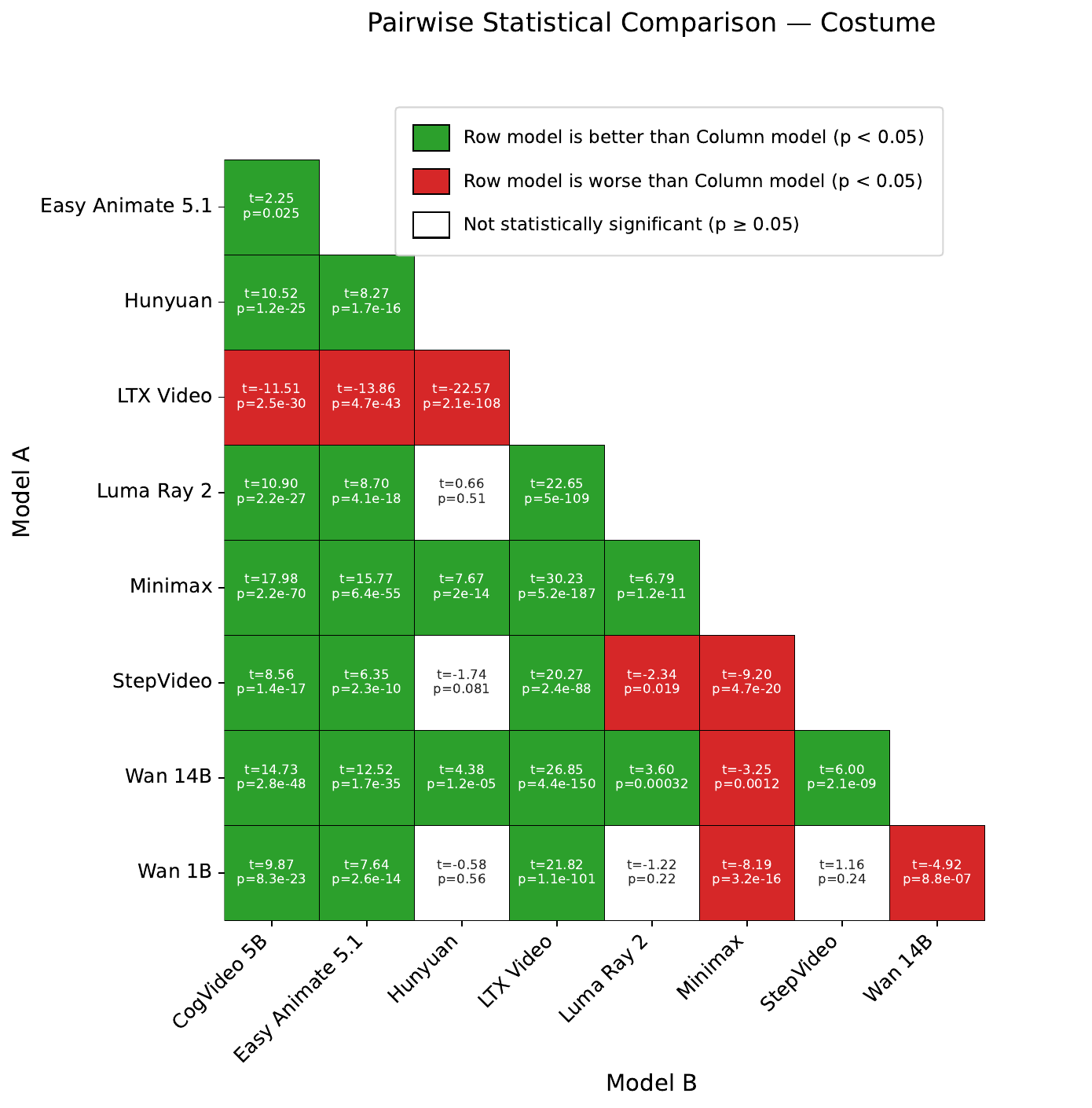}
    \caption{Pairwise t-tests across models on Subjet Costume}
    \label{fig:setup_subject_costume}
  \end{subfigure}

  \vspace{0.8em}

  \begin{subfigure}[b]{0.48\textwidth}
    \centering
    \includegraphics[width=\linewidth]{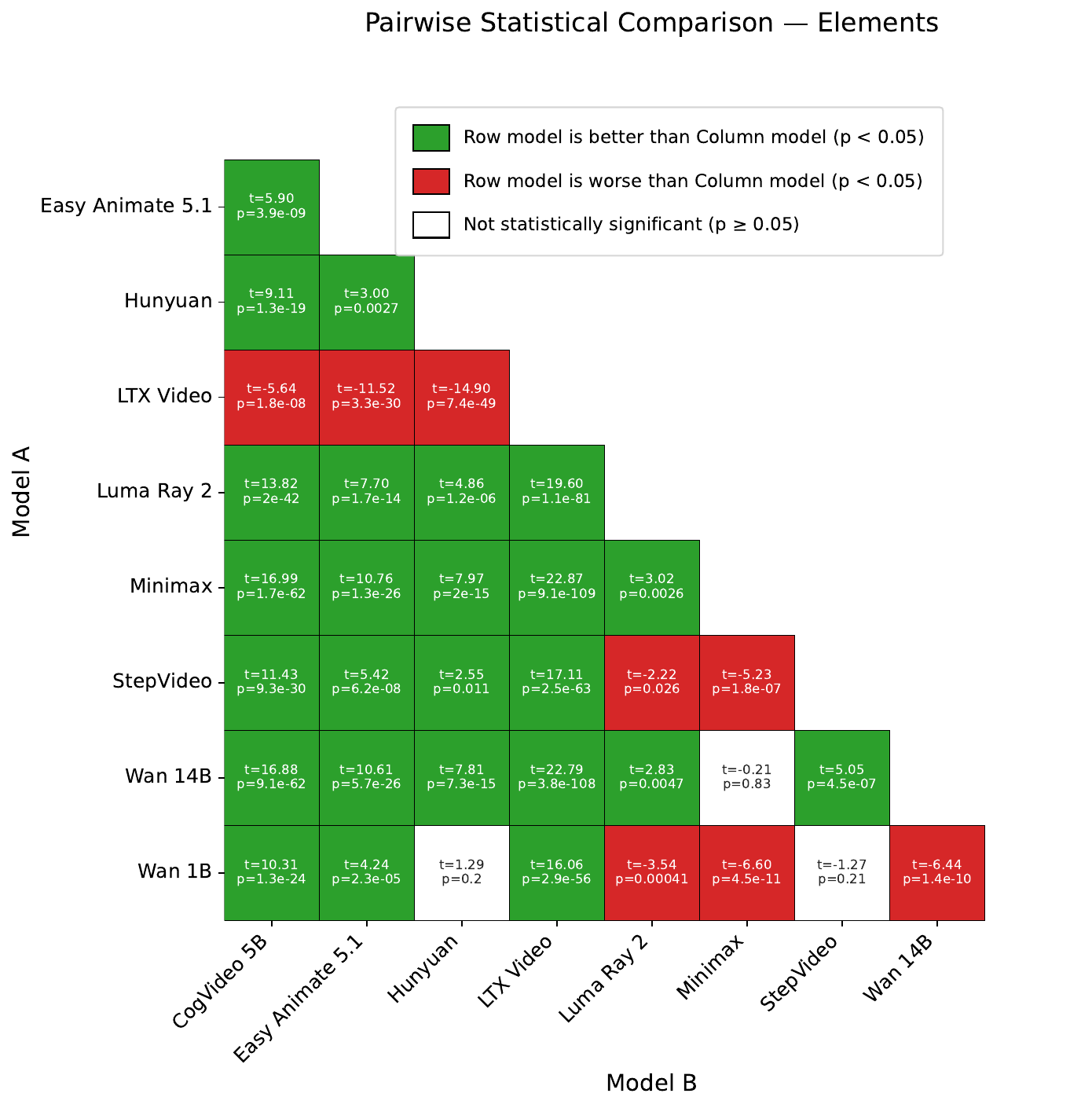}
    \caption{Pairwise t-tests across models on Elements}
    \label{fig:setup_elements}
  \end{subfigure}
  \hfill
  \begin{subfigure}[b]{0.48\textwidth}
    \centering
    \includegraphics[width=\linewidth]{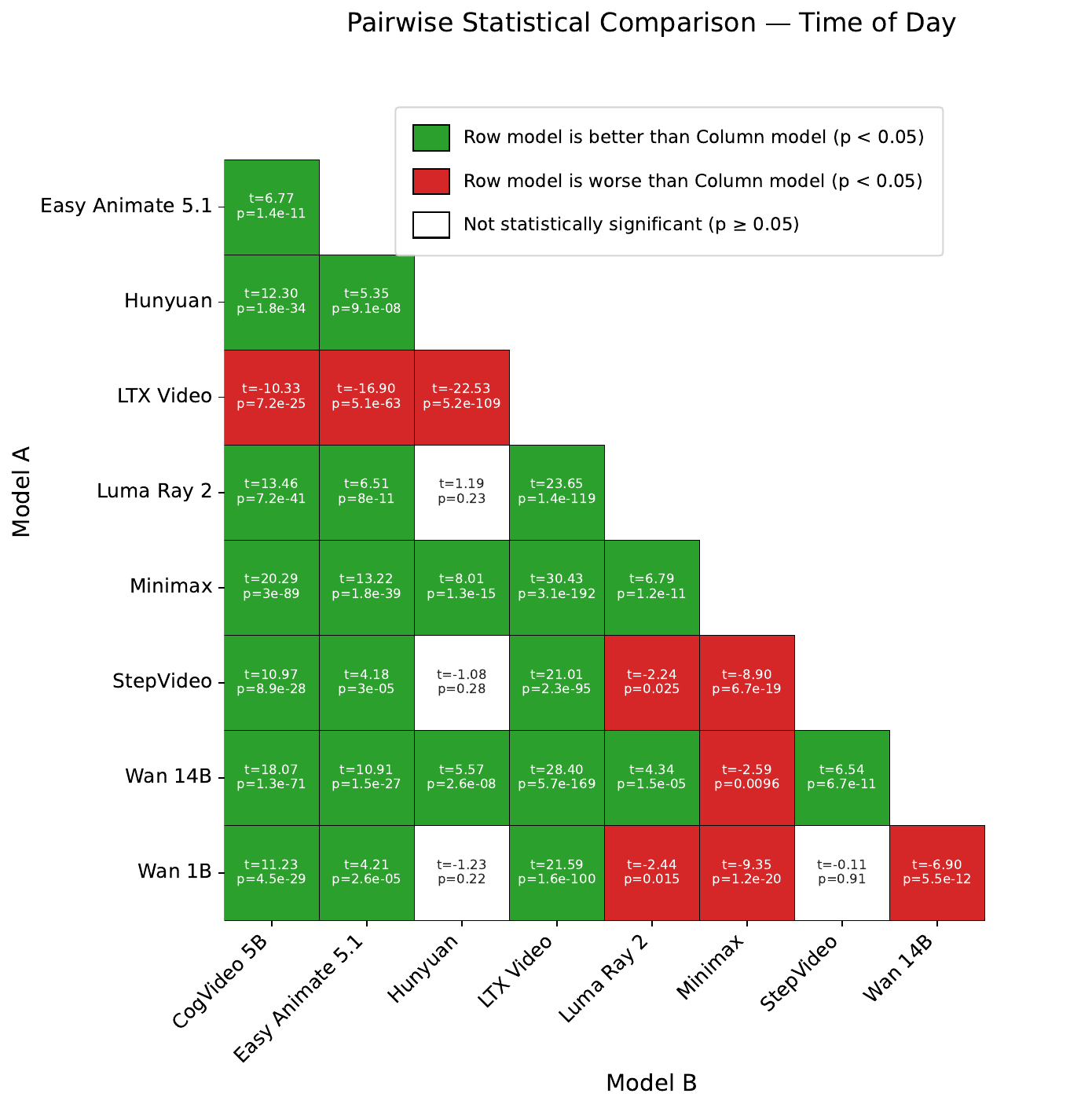}
    \caption{Pairwise t-tests across models on Time of Day}
    \label{fig:setup_time_of_day}
  \end{subfigure}

  \vspace{0.8em}

  \begin{subfigure}[b]{0.5\textwidth}
    \centering
    \includegraphics[width=\linewidth]{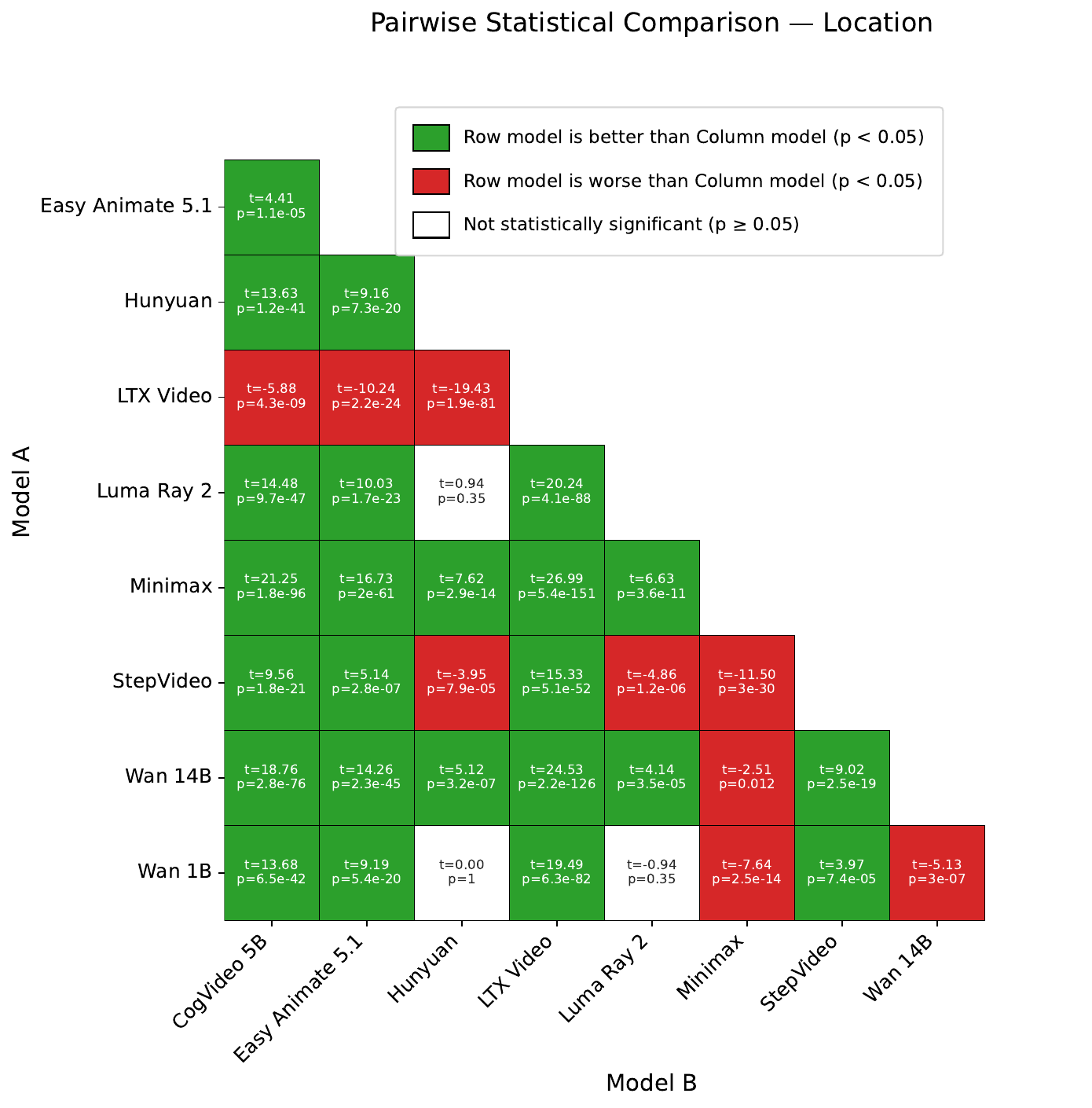}
    \caption{Pairwise t-tests across models on Location}
    \label{fig:setup_location}
  \end{subfigure}

  \caption{Statistical comparison matrices for Setup}
  \label{fig:t_test_setup}
\end{figure}

Pairwise t-tests show that the vast majority of model comparisons in our human evaluation, across all taxonomies are statistically significant at the 5\% level (p \textless 0.05). Figure (\ref{fig:t_test_events} - \ref{fig:t_test_setup}) presents the t-test results of Events, Lighting and Camera, and Setup, respectively.

\pagebreak
\subsection{Additional VLM results} \label{sec:vlm_additional}

\textbf{Node-specific results of VLM evaluator}  Table \ref{tab:vlm_nodes_top} and \ref{tab:vlm_nodes_bottom} list the set of nodes on which our VLM evaluator has the strongest and weakest performance, respectively.

\begin{table}[t]
  \centering
  \caption{Top 10 nodes where the VLM shows the strongest performance. 
  \textit{Score} indicates the accuracy of the VLM’s choice with the most common preference among human annotators.}
  \label{tab:vlm_nodes_top}
  \begin{tabularx}{\linewidth}{>{\raggedright\arraybackslash}X c}
    \toprule
    \textbf{Node} & \textbf{Score} \\
    \midrule
    Setup$\rightarrow$Scene$\rightarrow$Set Design$\rightarrow$Environment$\rightarrow$Mood & 0.88 \\
    Events$\rightarrow$Adv.Controls$\rightarrow$Rhythm$\rightarrow$Pace & 0.82 \\
    Setup$\rightarrow$Scene$\rightarrow$Set Design$\rightarrow$Environment$\rightarrow$Style & 0.82 \\
    Setup$\rightarrow$Scene$\rightarrow$Set Design$\rightarrow$Props$\rightarrow$Utility & 0.80 \\
    Events$\rightarrow$Types$\rightarrow$Emotions$\rightarrow$Exp.Types$\rightarrow$Explicit & 0.79 \\
    Setup$\rightarrow$Scene$\rightarrow$Geometry$\rightarrow$Frame$\rightarrow$Shapes$\rightarrow$Regular & 0.78 \\
    Lighting$\rightarrow$Lighting Effects$\rightarrow$Shadows$\rightarrow$Soft & 0.76 \\
    Setup$\rightarrow$Scene$\rightarrow$Geometry$\rightarrow$Space$\rightarrow$Spatial Loc.$\rightarrow$Rel.Pos. & 0.75 \\
    Events$\rightarrow$Types$\rightarrow$Actions$\rightarrow$Int.Types$\rightarrow$Standalone & 0.75 \\
    Events$\rightarrow$Types$\rightarrow$Emotions$\rightarrow$Exp.Types$\rightarrow$Explicit & 0.75 \\
    \bottomrule
  \end{tabularx}
\end{table}

\begin{table}[t]
  \centering
  \caption{Bottom 10 nodes where the VLM shows the weakest performance. 
  \textit{Score} indicates the accuracy of the VLM’s choice with the most common preference among human annotators.}
  \label{tab:vlm_nodes_bottom}
  \begin{tabularx}{\linewidth}{>{\raggedright\arraybackslash}X c}
    \toprule
    \textbf{Node} & \textbf{Score} \\
    \midrule
    Setup$\rightarrow$Subjects$\rightarrow$Makeup & 0.33 \\
    Setup$\rightarrow$Scene$\rightarrow$Set Design$\rightarrow$Environment$\rightarrow$Background & 0.33 \\
    Events$\rightarrow$Types$\rightarrow$Actions$\rightarrow$Portrayed as$\rightarrow$Contextual$\rightarrow$Background & 0.46 \\
    Setup$\rightarrow$Subjects$\rightarrow$Accessories & 0.53 \\
    Lighting$\rightarrow$Lighting Effects$\rightarrow$Reflection & 0.55 \\
    Setup$\rightarrow$Scene$\rightarrow$Texture$\rightarrow$Color Palette & 0.55 \\
    Camera$\rightarrow$Intrinsics$\rightarrow$Exposure$\rightarrow$Shutter Speed & 0.57 \\
    Lighting$\rightarrow$Adv.Controls$\rightarrow$Color Gels & 0.57 \\
    Setup$\rightarrow$Scene$\rightarrow$Texture$\rightarrow$Blur & 0.57 \\
    Lighting$\rightarrow$Color Temperature & 0.58 \\
    \bottomrule
  \end{tabularx}
\end{table}

\begin{figure}[]
  \centering
\includegraphics[width=0.9\linewidth]{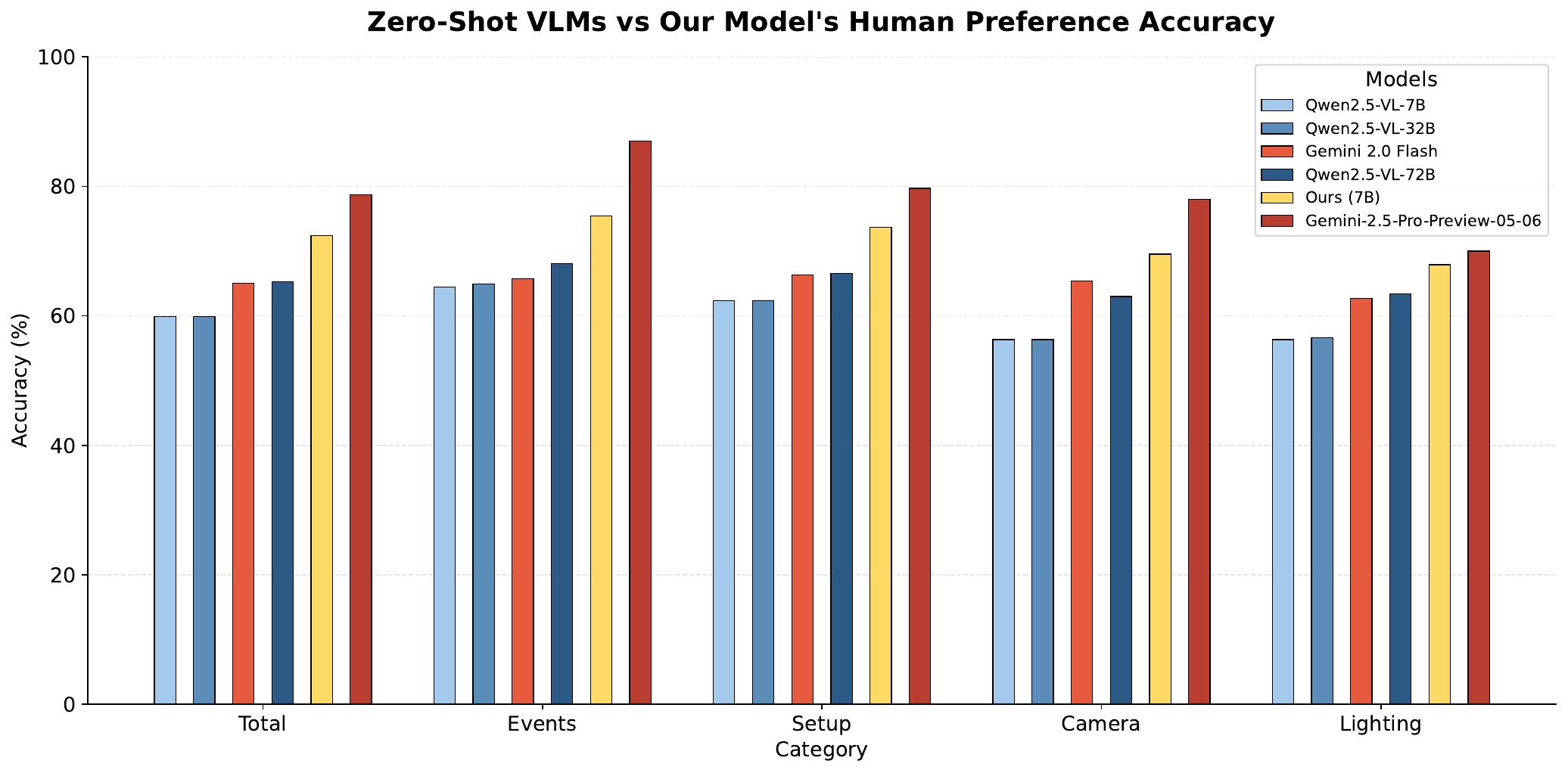}
  \caption {Preference Accuracy of open and closed-sourced VLMs in rating videos generated for Professional Use}
  \label{fig:vlm_acc_app}
\end{figure}

\textbf{Comparison with Closed Source Models} 
We extend our validation to closed-source, flagship SOTA models. Specifically, we evaluate two recent models from the Gemini family with distinct purposes: Gemini-2.0-Flash, optimized for fast inference, and Gemini-2.5-Pro-Preview-05-06, optimized for complex reasoning. We use the same human-aligned preference accuracy metric as with open-source models. Due to the lack of public details on model sizes, we cannot draw conclusions about scaling effects. However, Gemini-2.5-Pro consistently outperforms open-source models, including QwenVL-2.5-72B, across all categories. Notably, as shown in Figure \ref{fig:vlm_acc_app}, our 7B model outperforms Gemini-Flash across all categories and performs competitively with Gemini-2.5-Pro. This highlights the strength and scalability of our approach for professional video evaluation.

\textbf{Reliability in VLMs}  A reliable VLM-as-a-Judge should produce consistent scores when given the same video, prompt, and focus aspect. In this analysis, we evaluate the raw scores generated by VLMs rather than preference rankings, and measure their stability under Best-of-5 sampling. Since VLMs are probabilistic, we evaluate reliability via the standard deviation of scores across runs. We use temperature=0 to sample to make ensure that the highest probability is selected at each sampling step. We exclude our model from this analysis, as its architecture includes a dedicated value head, unlike zero-shot VLMs that produce rewards as text. Our results show that Qwen-2.5VL-3B exhibits a high variance, making it unreliable under repeated sampling. In contrast, the flagship models and the strongest open-source model, QwenVL-2.5-72B, demonstrate high reliability, with consistently low variance (Table \ref{tab:metrics_comparison}).

\begin{table}[htbp]
    \centering
    \caption{Measuring VLM Reliability across best-of-5 sampling}
    \label{tab:metrics_comparison}
    \resizebox{0.7\linewidth}{!}{
    \begin{tabular}{@{}l|c|c@{}}
        \toprule
        \textbf{Model} & \textbf{Standard-Deviation} $\downarrow$ & \textbf{Krippendorff-alpha} $\uparrow$ \\
        \midrule
        Qwen2.5-VL-3B & 2.34 & 0.36 \\
        Qwen2.5-VL-7B & 0.37 & 0.84 \\
        Qwen2.5-VL-32B & 0.47 & 0.65 \\
        Qwen2.5-VL-72B & 0.23 & \textbf{0.95} \\
        \midrule
        Gemini2.5-Flash & 0.20 & 0.90 \\
        {Gemini2.5-Pro} & \textbf{0.14} & \textbf{0.95} \\
        \bottomrule
    \end{tabular}
    }
\end{table}

\subsection{Additional Results on Recent Models} \label{sec:new_model_results}

We also perform small-scale evaluations on recently released models, specifically Veo 3 (Fast) \cite{deepmind2025veo3} and the Wan 2.2 family of models. Results are presented in Table \ref{tab:new_model_category_scores}.

\begin{table}[!t]
  \centering
  \caption{Average scores across taxonomy categories on recently released video generative models.}
  \label{tab:new_model_category_scores}
  \begin{tabular}{lccccc}
    \toprule
    \textbf{Model} & \textbf{Camera} & \textbf{Events} & \textbf{Lighting} & \textbf{Setup} & \textbf{Overall Avg.} \\
    \midrule
    Veo 3 Fast     & \textbf{3.686} & \textbf{3.382} & 3.974 & 4.078 & \textbf{3.780} \\
    Wan 2.2 14B    & 3.584 & 3.060 & \textbf{4.009} & \textbf{4.114} & 3.692 \\
    Wan 2.1 14B    & 3.239 & 2.821 & 3.827 & 3.975 & 3.466 \\
    Wan 2.2 5B     & 3.194 & 2.705 & 3.838 & 3.913 & 3.412 \\
    Wan 2.1 1B     & 3.240 & 2.579 & 3.698 & 3.742 & 3.315 \\
    \bottomrule
  \end{tabular}
\end{table}

\end{document}